\documentclass[journal]{IEEEtran}
\usepackage{makecell}
\usepackage{siunitx}
\usepackage{multirow,bigstrut}
\usepackage[normalem]{ulem}
\useunder{\uline}{\ul}{}
\usepackage{graphicx}
\usepackage{float}
\usepackage{scrextend}
\usepackage{subfig}
\usepackage{makecell}
\usepackage{amsmath,amssymb,mathrsfs,bm}
\usepackage{algorithm}
\usepackage{algpseudocode}
\usepackage{array}
\usepackage{fixltx2e}
\usepackage{stfloats,float}
\usepackage{arydshln} 
\usepackage[nocompress]{cite}
\usepackage[colorlinks]{hyperref}
\usepackage{xurl}
\usepackage{setspace}
\usepackage{tikz,pgfplots}
\usepackage{amsthm}

\newtheorem{lemma}{Lemma} 

\newtheorem{theorem}{Theorem}
\newtheorem{definition}[theorem]{Definition} 

\newtheorem{corollary}[theorem]{Corollary}

\newcommand\tensor[1]{\mathcal{#1}}

\begin{document}
\title{Cross-Frequency Implicit Neural Representation with Self-Evolving Parameters}
\author{Chang Yu,
Yisi Luo,
Kai Ye,
Xile Zhao,
\and Deyu Meng
\thanks{Chang Yu, Yisi Luo, and Deyu Meng are with the School of Mathematics and Statistics, Xi'an Jiaotong University, Xi'an, 710049, China.}
\thanks{Kai Ye is with the School of Automation Science and Engineering, Faculty of Electronic and Information Engineering, Xi'an Jiaotong University, Xi'an, 710049, China.}
\thanks{Xile Zhao is with the School of Mathematical Science, University of Electronic Science and Technology of China, Chengdu, 610000, China.}}
\maketitle
\begin{abstract}
Implicit neural representation (INR) has emerged as a powerful paradigm for visual data representation. However, classical INR methods represent data in the original space mixed with different frequency components, and several feature encoding parameters (e.g., the frequency parameter $\omega$ or the rank $R$) need manual configurations. In this work, we propose a self-evolving cross-frequency INR using the Haar wavelet transform (termed CF-INR), which decouples data into four frequency components and employs INRs in the wavelet space. CF-INR allows the characterization of different frequency components separately, thus enabling higher accuracy for data representation. To more precisely characterize cross-frequency components, we propose a cross-frequency tensor decomposition paradigm for CF-INR with self-evolving parameters, which automatically updates the rank parameter $R$ and the frequency parameter $\omega$ for each frequency component through self-evolving optimization. This self-evolution paradigm eliminates the laborious manual tuning of these parameters, and learns a customized cross-frequency feature encoding configuration for each dataset. We evaluate CF-INR on a variety of visual data representation and recovery tasks, including image regression, inpainting, denoising, and cloud removal. Extensive experiments demonstrate that CF-INR outperforms state-of-the-art methods in each case. 
\end{abstract}
\IEEEpeerreviewmaketitle
\section{Introduction}
Implicit neural representation (INR) \cite{INR1,INR2} has recently emerged as a powerful paradigm for continuously representing signals, offering advantages such as compactness, resolution independence, and differentiability. The use of INR has attracted significant attention across diverse fields, such as low-level vision \cite{low-level, LRTFR}, graphics \cite{graphics}, medical imaging \cite{medical1, medical2}, remote sensing \cite{remote}, PDE solving \cite{PDE}, etc. Unlike traditional grid-based discrete representations, INR learns a continuous function in the form of inputting a spatial coordinate and outputting the corresponding value using deep neural networks, which provide an implicit and continuous representation of data with greater flexibility and differentiability, therefore enabling diverse applications across fields.\par
\begin{figure}[htbp]
\scriptsize
    \setlength{\tabcolsep}{0.9pt}
    \begin{center}
\begin{tabular}{c}
\centering
    \includegraphics[width=0.42\textwidth]{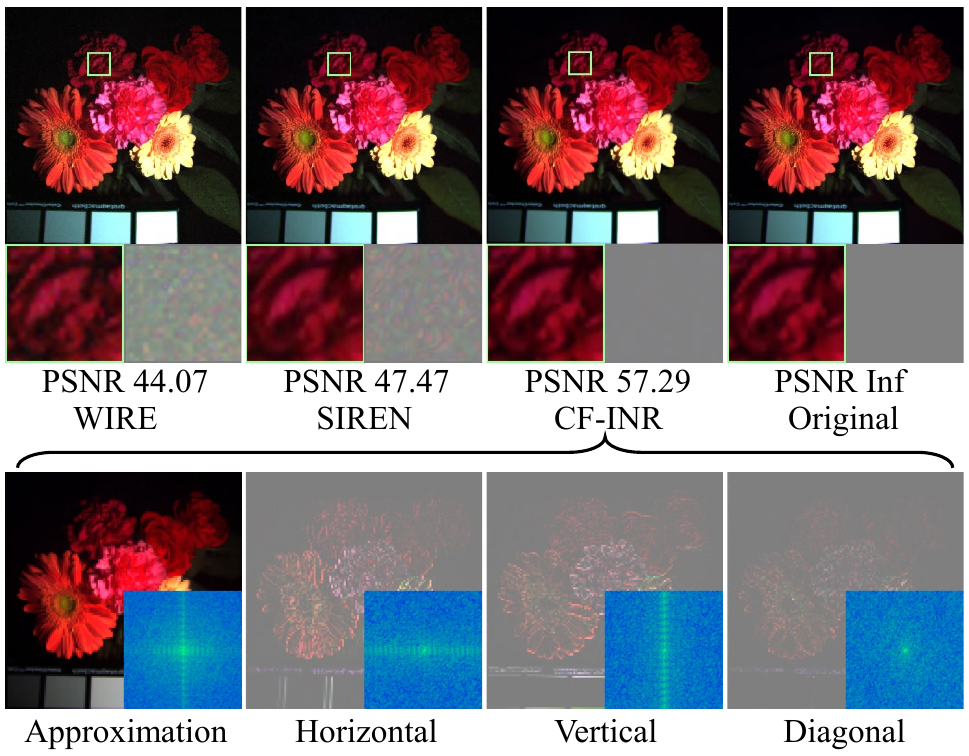}\vspace{-0.37cm} 
    \end{tabular}
    \end{center}
\caption{The frequency decoupling of CF-INR enables more accurate continuous data representation. Top: fitting a multispectral image using different INRs (WIRE \cite{Wire}, SIREN \cite{activation_function1}, and the proposed frequency decoupled CF-INR), along with zoom-in figures and the corresponding error maps. Bottom: the wavelet coefficients learned by CF-INR with the corresponding Fourier spectrum.\label{fig:big_figure}\vspace{-0.5cm}}
\end{figure}
Classical INR methods typically represent data in the original space (e.g., image space), where different frequency components are usually mixed \cite{activation_function1,LRTFR,CRNL}. Specifically, one of the classical choices is to utilize a multilayer perceptron (MLP) to map an arbitrary spatial coordinate to the corresponding value\cite{Fourier_feature,activation_function1}, which implicitly represents the data and encodes the local smoothness (i.e., input coordinates that are closer to each other result in more similar outputs\cite{LRTFR}). It was shown that traditional MLPs with ReLU activation fail to capture high-frequency components under the INR framework due to the spectral bias towards low-frequency components\cite{Fourier_feature}. One common approach to enhance the representation ability of MLPs for learning high-frequency details is to introduce specialized mappings. For instance, the well-known SIREN method \cite{activation_function1} constructs a differentiable MLP for continuous representation by stacking multiple linear layers with the sinusoidal activation function $\sin(\omega\cdot)$, where $\omega$ is a global tunable frequency parameter that benefits high-frequency information capture in natural images and scenes, such as texture and edge details. The rationale of using the sinusoidal activation function for INR is that an MLP with sinusoidal functions (under certain conditions) is equivalent to a kernel regression with a diagonal shift-invariant kernel from the neural tangent kernel perspective\cite{NTK}, while conventional activations such as ReLU lead to a non-diagonal kernel\cite{Fourier_feature}. Hence, the sinusoidal activation function-based MLP \cite{activation_function1} tends to more effectively represent high-frequency information and has become a widely considered baseline for INR methods and applications \cite{ECCV_Wang,ECCV_Li}. Recently, Luo et al. \cite{LRTFR} proposed a low-rank tensor function representation based on SIREN, which parameterizes the INR using the tensor decomposition by pre-defining a suitable rank vector $R$ for the given data.\par
Despite their demonstrated effectiveness, existing INR methods with a global frequency parameter $\omega$ face a fundamental limitation---they often neglect the intrinsic multi-frequency nature of real-world data. Real-world data inherently exhibit heterogeneous frequency characteristics, making the use of a single global frequency parameter suboptimal for accurate representation. Although adjusting $\omega$ could shift the frequency preference of the INR model towards specific signal characteristics, it still neglects the frequency multiplicity inside a signal. The intrinsic limitation necessitates the development of a fundamentally different INR paradigm that can account for the frequency multiplicity of a signal by modeling different frequency components separately. On the other hand, the requirement for manual configuration of the frequency parameter $\omega$ and the tensor rank $R$ \cite{LRTFR} introduces substantial hyperparameter optimization overhead. Current practices rely on heuristic parameter tuning, which demands domain expertise and may fail to guarantee cross-dataset generalization, where static parameter sets often prove inadequate for capturing diverse properties of different datasets.\par
To address these fundamental challenges, we propose a self-evolving cross-frequency INR by using the Haar wavelet transform (termed CF-INR). The Haar wavelet transform effectively decomposes the input signal into four distinct frequency components (i.e., wavelet coefficients). By designing dedicated INRs tailored to represent different wavelet coefficients, we can effectively encode cross-frequency features, enabling independent characterizations of different frequency components in the wavelet space. Consequently, the frequency-decoupled CF-INR encodes the characteristics of various frequencies, thereby effectively capturing both coarse-grained and fine-grained data structures to obtain higher accuracy for signal representation (see Fig. \ref{fig:big_figure} for examples).\par 
Customized with the frequency decoupling, we further propose a self-evolving cross-frequency tensor decomposition paradigm for CF-INR, which automatically updates the rank parameter $R$ and the frequency parameter $\omega$ for each wavelet coefficient that enables precise characterizations of cross-frequency features. In particular, the self-evolving CF-INR enjoys three essential advantages for continuous data representation. First, {\bf the cross-frequency tensor decomposition paradigm} tailored for CF-INR enhances the computational efficiency and effectively captures the intricate intra- and inter-relationships among different wavelet coefficients, thus promoting the data recovery performance through intrinsic knowledge sharing among wavelet coefficients. Second, by tailoring different {\bf cross-frequency parameters ($\omega_1$-$\omega_4$) and cross-frequency ranks ($R_1$-$R_4$)} for distinct wavelet coefficients, CF-INR holds intrinsic superiority for heterogeneous cross-frequency feature encoding that benefits a more meticulous data characterization. Third, the cross-frequency parameters ($\omega_1$-$\omega_4$) and cross-frequency ranks ($R_1$-$R_4$) are {\bf further automatically updated and evolved} toward the desired cross-frequency features under self-evolution, with such update rules being derived from theoretical analysis.\par 
Specifically, we have established two innovative theoretical frameworks that facilitate our understanding of CF-INR---the cross-frequency tensor rank (CF-Rank) and the cross-frequency Laplacian smoothness. In terms of CF-Rank, we show that the low-rankness of different wavelet coefficients varies, and hence introduce the cross-frequency nuclear norm as a theoretical convex surrogate of CF-Rank to automatically update the rank of each wavelet coefficient during optimization. In terms of smoothness, we develop the theoretical Laplacian bound for the four wavelet coefficients represented by CF-INR, which is polynomially related to $\omega_1$-$\omega_4$. Hence, we automatically update the cross-frequency parameters $\omega_1$-$\omega_4$ based on the Laplacian values of wavelet coefficients during optimization. As the learning process proceeds, the CF-Rank ($R_1$-$R_4$) and cross-frequency parameters ($\omega_1$-$\omega_4$) progressively converge toward more precise values that effectively capture the intrinsic cross-frequency features of data. Such self-evolving CF-INR reduces the labor-intensive process of hyperparameter search and automatically identifies suitable frequency configurations through theoretical relationships. 
The main contributions of this work are outlined as follows: 
\begin{itemize}
\item We propose CF-INR, the first INR method that explicitly disentangles the multi-frequency characteristics of data by using the Haar wavelet transform, thus achieving superior accuracy for continuous data representation. 
\item We further suggest the cross-frequency tensor decomposition with spectral coupling and spatial decoupling paradigms to improve CF-INR, which largely enhances computational efficiency and exploits both the intra- and inter-relationships among different wavelet coefficients. 
\item We introduce two innovative theoretical foundations underlying CF-INR---the cross-frequency tensor rank and the cross-frequency Laplacian smoothness, and establish rigorous theoretical analysis to quantify the attributes underlying cross-frequency features (e.g., the cross-frequency low-rankness and smoothness). 
\item Inspired by theoretical analysis, we introduce self-evolution mechanisms that automatically update the cross-frequency ranks $R_1$-$R_4$ and cross-frequency parameters $\omega_1$-$\omega_4$ of CF-INR to precisely capture multi-frequency features without laborious manual tuning.
\item Comprehensive experiments on multi-dimensional data representation and recovery (image regression, inpainting, denoising, and cloud removal) validate the effectiveness and superiority of CF-INR as compared with state-of-the-art methods for continuous data representation. We also integrate CF-INR with instant neural graphics primitives (INGP) \cite{INGP} for neural radiance field (NeRF), and the detailed results are included in supplementary file.
\end{itemize}\par
The rest of this paper is organized as follows. Sec. \ref{sec: Related work} provides a comprehensive review of related work on high-frequency-enhanced INR and wavelet transform for deep learning. Sec. \ref{sec: The Proposed Method} presents the detailed theories and methodologies of the proposed CF-INR. Sec. \ref{sec: Experiments} conducts experiments on various tasks and presents ablation studies to justify the proposed method. Sec. \ref{sec: Conclusion} concludes the paper.
\section{Related work\label{sec: Related work}}
\subsection{High Frequency-Enhanced INR}
INR provides a compact representation of data by mapping spatial coordinates to values, and has been successfully applied to 2D images \cite{activation_function1,Fourier_feature}, 3D shapes \cite{3D_images1,diverse_fields2,shape_generation2,3D_images2,3D_images3}, point clouds \cite{LRTFR,CRNL}, neural rendering \cite{neural_rendering1,neural_rendering3}, and scene representation \cite{diverse_fields3}. Researchers have explored various strategies for enhancing classical INR methods, such as feature encoding\cite{Fourier_feature}, activation function design\cite{activation_function1, Wire, FINER}, normalization techniques\cite{Batch}, and weight parameterization methods\cite{Fourier_Reparameterized}.
To be specific, Tancik et al.\cite{Fourier_feature} proposed the positional encoding method by mapping input coordinates into a higher-dimensional Fourier space, thereby enhancing the ability of INR to capture high-frequency components. Another widely used approach is the SIREN method\cite{activation_function1}, which utilizes sinusoidal activation functions to enhance the representation abilities for high-frequency details. Additionally, Liu et al. \cite{FINER} proposed the variable-periodic activation function for INR, which enables the network to flexibly tune its supported frequency set. The WIRE method\cite{Wire} incorporates the complex Gabor wavelet basis functions as activation functions, which leverage the time-frequency localization property of the wavelet to improve the representation ability of INR. These enhanced activation function-based methods (such as positional encoding/SIREN) rely on predefined sets of frequency parameters (for instance $\omega$ in SIREN), and thus may introduce a mismatch when the frequency characteristics of the target signal are unknown. Compared to these INR methods, we provide an advanced frequency decoupled INR by using the Haar wavelet transform to decouple the signal into four different frequency components, which allows us to use INRs with varying frequency parameters to characterize different wavelet coefficients. Moreover, the proposed CF-INR further benefits from the self-evolving strategy to more efficiently and automatically capture multi-frequency characteristics. Notably, our method is orthogonal to current INR methods and we have attempted to combine them; see Table \ref{tab:fitting}.\par
More recently, Shi et al. \cite{Fourier_Reparameterized} proposed the Fourier reparameterized training, which reparameterizes weights of INR in the Fourier domain to accelerate the convergence for high-frequency components. It was also shown that batch normalization could alter the eigenvalue distribution of the neural tangent kernel of INR and increase its effective rank \cite{Batch},\cite{NTK}, thereby enhancing the network's capacity to learn high-frequency details. 
Compared to these approaches, our CF-INR introduces a novel methodology by decoupling the data into four frequency components using the Haar wavelet transform, which allows us to model different frequency components separately using INR, achieving better representation abilities. The proposed frequency self-evolving strategy further enables more efficient characterization of cross-frequency components within the CF-INR framework. 
\subsection{Wavelet Transform in Deep Learning}
The wavelet transform has been widely used for data processing by decomposing data into multiple frequency components. Recent studies have demonstrated that integrating wavelet transform into deep learning frameworks would be helpful for various vision applications\cite{Finder, gao2024efficient, duan2024local}.
For example, Finder et al.\cite{Finder} leveraged wavelet transform to enhance the receptive field and frequency awareness of CNNs. Yang et al.\cite{SFFNet} proposed SFFNet, combining spatial and frequency features via wavelet-guided transformers. Gao et al.\cite{gao2024efficient} introduced a learnable wavelet transform for deblurring, enabling adaptive restoration of high-frequency details. Similarly, Duan et al.\cite{duan2024local} utilized the wavelet transform in the arbitrary-scale image super-resolution task by introducing the local implicit wavelet transformer.
INR is born with frequency preference \cite{Fourier_feature} due to its spectral bias. Hence, decoupling data into multiple frequency components is naturally suitable for INR to better characterize each component. Different from the above wavelet transform-based deep learning methods, we innovatively introduce the frequency decoupling strategy for INR and propose the self-evolving strategy by automatically learning the cross-frequency configurations (i.e., cross-frequency low-rankness and smoothness) of each wavelet coefficient. Hence, our CF-INR obtains higher accuracy for data representation under the INR framework. Moreover, as a general data representation framework, CF-INR can be more easily applied to different signal processing tasks compared to other end-to-end training-based deep learning methods. 
\section{The Proposed CF-INR\label{sec: The Proposed Method}}
We first introduce necessary preliminaries in Sec. \ref{sec_1}. Then we introduce the proposed Haar wavelet transform-induced CF-INR in Sec. \ref{sec_2}, followed by the efficient cross-frequency tensor decomposition design for CF-INR in Sec. \ref{sec_3}. We establish the theoretical framework of cross-frequency rank and introduce the automatic CF-Rank evolution algorithm in Sec. \ref{sec_4_rank}. Then we establish the theoretical Laplacian bound of CF-INR and propose self-evolution for cross-frequency parameters in Sec. \ref{sec_5_Lap}. Finally, we present the models and algorithms of applying CF-INR for data representation and recovery tasks in Sec. \ref{sec_6}.
\subsection{Notations and Preliminaries\label{sec_1}}
\subsubsection{Tensor Notations}
The scalar, vector, matrix, and tensor are denoted as $x$, $\boldsymbol{x}$, $\boldsymbol{X}$, and $\tensor{X}$, respectively. Given a tensor $\tensor{X}\in \mathbb{R}^{n_1 \times n_2 \times n_3}$, the element at position $(i, j, k)$ in $\tensor{X}$ is denoted as $\tensor{X}(i, j, k)$. The Frobenius norm of a tensor $\tensor{X} \in \mathbb{R}^{n_1 \times n_2 \times n_3}$ is defined as $\|\tensor{X}\|_{F} = \sqrt{\sum_{i, j, k} \tensor{X}(i, j, k)^2}$.
The mode-$i$ ($i=1,2,3$) unfolding operator of a tensor $\tensor{X} \in \mathbb{R}^{n_1 \times n_2 \times n_3}$ results in a matrix denoted by $
{\tt unfold}_{i}(\tensor{X}) = \boldsymbol{X}_{(i)} \in \mathbb{R}^{n_{i} \times \prod_{j \neq i} n_{j}}$. The operator ${\tt fold}_{i}(\cdot)$ is defined as the inverse of ${\tt unfold}_{i}(\cdot)$. The mode-$i$ product between a tensor $\tensor{X}\in \mathbb{R}^{n_1 \times n_2 \times n_3}$ with a matrix $\boldsymbol{A}\in \mathbb{R}^{n\times n_i}$ is defined as $
\tensor{X} \times_{i} \boldsymbol{A} = {\tt fold}_{i}\left(\boldsymbol{A} \boldsymbol{X}_{(i)}\right)$.

\begin{definition}[Tensor Tucker rank \cite{Tucker}]
For a tensor $\tensor{X} \in \mathbb{R}^{n_1 \times n_2 \times n_3}$, its Tucker rank is vector defined as $
\mathrm{rank}_{T}(\tensor{X}) = \left(\mathrm{rank}(\boldsymbol{X}_{(1)}), \mathrm{rank}(\boldsymbol{X}_{(2)}), \mathrm{rank}(\boldsymbol{X}_{(3)})\right)$.
\end{definition}
\begin{lemma}[Tensor Tucker decomposition
\cite{Tucker}]\label{Tucker}
For a tensor $\tensor{X} \in \mathbb{R}^{n_1 \times n_2 \times n_3}$ with Tucker rank $\mathrm{rank}_{T}(\tensor{X}) = (r_1, r_2, r_3)$, there exist a core tensor $\tensor{C} \in \mathbb{R}^{r_1 \times r_2 \times r_3}$ and three factor matrices $\boldsymbol{U} \in \mathbb{R}^{n_1 \times r_1}$, $\boldsymbol{V} \in \mathbb{R}^{n_2 \times r_2}$, and $\boldsymbol{W} \in \mathbb{R}^{n_3 \times r_3}$ such that $\tensor{X}$ can be represented by
$
\tensor{X} = \tensor{C}\times_{3} \boldsymbol{W}  \times_{2} \boldsymbol{V}\times_{1} \boldsymbol{U} .
$
\end{lemma}
\subsubsection{Haar Wavelet Transform}
In this section, we introduce the notion of Haar wavelet transform and its properties.
\begin{definition}[Haar discrete wavelet transform \cite{wavelets}]
Given a matrix $\boldsymbol{A} \in \mathbb{R}^{n_1 \times n_2}$, the 2-D Haar discrete wavelet transform (HWT) is defined as $
\boldsymbol{B} = \boldsymbol{W}_{n_1} \boldsymbol{A} \boldsymbol{W}_{n_2}^{T}\in \mathbb{R}^{n_1 \times n_2},$ where $\boldsymbol{W}_{n_1}\in \mathbb{R}^{n_1 \times n_1}$ and $\boldsymbol{W}_{n_2}\in\mathbb{R}^{n_2 \times n_2}$ are two orthogonal projection matrices given by
\begin{equation}\small\label{eq:haar_wavelet_transform_matrix}
    \begin{aligned}
        \boldsymbol{W}_{n_1}  = \left[ \begin{array}{c}
            \boldsymbol{H}_{n_1/2}\\
            \boldsymbol{G}_{n_1/2}\\
        \end{array} \right],\;\;
\boldsymbol{W}_{n_2}  = \left[ \begin{array}{c}
            \boldsymbol{H}_{n_2/2}\\
            \boldsymbol{G}_{n_2/2}\\
        \end{array} \right],
\end{aligned}
\end{equation}
where
\begin{equation}\small\label{eq:haar_wavelet_transform_matrix}
    \begin{aligned}
    \boldsymbol{H}_{n_1/2}=&\underbrace{\left[ \begin{matrix}
                \frac{\sqrt{2}}{2}&		\frac{\sqrt{2}}{2}&		0&		0&		\cdots&		0&		0\\
                0&		0&		\frac{\sqrt{2}}{2}&		\frac{\sqrt{2}}{2}&		\cdots&		0&		0\\
                \vdots&		&		&		&		\ddots&		&		\vdots\\
                0&		0&		0&		0&		\cdots&		\frac{\sqrt{2}}{2}&		\frac{\sqrt{2}}{2}\\  
            \end{matrix} \right]}_{\frac{n_1}{2}\times n_1},\\
\boldsymbol{G}_{n_1/2}=&\underbrace{\left[ \begin{matrix}
                \frac{\sqrt{2}}{2}&		-\frac{\sqrt{2}}{2}&		0&		0&		\cdots&		0&		0\\
                0&		0&		\frac{\sqrt{2}}{2}&		-\frac{\sqrt{2}}{2}&		\cdots&		0&		0\\
                \vdots&		&		&		&		\ddots&		&		\vdots\\
                0&		0&		0&		0&		\cdots&		\frac{\sqrt{2}}{2}&		-\frac{\sqrt{2}}{2}\\
            \end{matrix} \right]}_{\frac{n_1}{2}\times n_1}.         
    \end{aligned}
\end{equation} 
The matrices $\boldsymbol{H}_{n_2/2}\in{\mathbb R}^{\frac{n_2}{2}\times n_2}$ and $\boldsymbol{G}_{n_2/2}\in{\mathbb R}^{\frac{n_2}{2}\times n_2}$ are defined the same way as $\boldsymbol{H}_{n_1/2}$ and $\boldsymbol{G}_{n_1/2}$.
\end{definition}
By decomposing $\boldsymbol{W}_{n_1}$ and $\boldsymbol{W}_{n_2}$ into block structures, the matrix $\boldsymbol{B}$ can be represented as a block matrix:
\begin{equation}\small\label{eq:block_B}
\begin{aligned}
    \boldsymbol{B} &= \boldsymbol{W}_{n_1} \boldsymbol{A} \boldsymbol{W}_{n_2}^{T} = \begin{bmatrix}
        \boldsymbol{H}_{n_{1}/2} \\
        \boldsymbol{G}_{n_1/2}
    \end{bmatrix} \boldsymbol{A} 
    \begin{bmatrix}
        \boldsymbol{H}_{n_2/2} \\
        \boldsymbol{G}_{n_2/2}
    \end{bmatrix}^{T} \\
    &= \begin{bmatrix}
        \boldsymbol{H}_{n_1/2}\boldsymbol{A}\boldsymbol{H}_{n_2/2}^{T} & \boldsymbol{H}_{n_1/2}\boldsymbol{A}\boldsymbol{G}_{n_2/2}^{T} \\
\boldsymbol{G}_{n_1/2}\boldsymbol{A}\boldsymbol{H}_{n_2/2}^{T} & \boldsymbol{G}_{n_1/2}\boldsymbol{A}\boldsymbol{G}_{n_2/2}^{T}
    \end{bmatrix} \triangleq \begin{bmatrix}
        \boldsymbol{B}_1 & \boldsymbol{B}_2 \\
        \boldsymbol{B}_3 & \boldsymbol{B}_4
    \end{bmatrix}.
\end{aligned}
\end{equation}
The matrices $\boldsymbol{B}_1$, $\boldsymbol{B}_2$, $\boldsymbol{B}_3$, $\boldsymbol{B}_4\in \mathbb{R}^{\frac{n_1}{2} \times \frac{n_2}{2}}$ correspond to the approximation (downsampling), horizontal, vertical, and diagonal wavelet coefficients. The HWT \eqref{eq:block_B} decomposes a matrix $\boldsymbol{A}$ into four frequency components, including the approximate (low-frequency) component $\boldsymbol{B}_1$ and the horizontal, vertical, diagonal (high-frequency) components $\boldsymbol{B}_2$, $\boldsymbol{B}_3$, $\boldsymbol{B}_4$ \cite{wavelets}.\par
Since $\boldsymbol{W}_{n_1}$ and $\boldsymbol{W}_{n_2}$ are orthogonal projection matrices, the HWT is invertible. The inverse HWT (IHWT) is calculated by
\begin{equation}\small\begin{aligned}\boldsymbol{A}=\boldsymbol{W}_{n_1}^{T}\boldsymbol{B}\boldsymbol{W}_{n_2} 
        =\left[ \begin{array}{c}
            \boldsymbol{H}_{n_1/2}\\
            \boldsymbol{G}_{n_1/2}\\
        \end{array} \right]^T \left[ \begin{matrix}
            \boldsymbol{B}_1		&\boldsymbol{B}_2\\
            \boldsymbol{B}_3		&\boldsymbol{B}_4\\
        \end{matrix} \right]\left[ \begin{array}{c}
            \boldsymbol{H}_{n_2/2}\\
            \boldsymbol{G}_{n_2/2}\\
        \end{array} \right]. \\
    \end{aligned}
\end{equation}
Each element of the IHWT $\boldsymbol{A}$ can be validated through
\begin{equation}\small
\begin{aligned}
&\boldsymbol{A}(2i-1,2j-1) = (\boldsymbol{B}_1(i,j) - \boldsymbol{B}_2(i,j) - \\
&\hspace{4cm} \boldsymbol{B}_3(i,j) + \boldsymbol{B}_4(i,j)) / 2, \\[1mm]
&\boldsymbol{A}(2i,2j-1) = (\boldsymbol{B}_1(i,j) - \boldsymbol{B}_2(i,j) + \\
&\hspace{4cm} \boldsymbol{B}_3(i,j) - \boldsymbol{B}_4(i,j)) / 2, \\[1mm]
&\boldsymbol{A}(2i-1,2j) = (\boldsymbol{B}_1(i,j) + \boldsymbol{B}_2(i,j) - \\
&\hspace{4cm} \boldsymbol{B}_3(i,j) - \boldsymbol{B}_4(i,j)) / 2, \\[1mm]
&\boldsymbol{A}(2i,2j) = (\boldsymbol{B}_1(i,j) + \boldsymbol{B}_2(i,j) + \\
&\hspace{4cm} \boldsymbol{B}_3(i,j) + \boldsymbol{B}_4(i,j)) / 2.
\end{aligned}
\end{equation}\par
Given a third-order tensor $\tensor{A} \in \mathbb{R}^{n_1 \times n_2 \times n_3}$, we apply the slice-wise HWT to decompose its spatial information into different frequencies.
\begin{definition}[Frontal slice-wise HWT\cite{wavelets}]\label{def_HWT_tensor}
    Given a third-order tensor $\tensor{A}\in\mathbb{R}^{n_1\times n_2\times n_3}$, the 2-D frontal slice-wise HWT is defined as $
\tensor{B}=\tensor{A}\times_{1}\boldsymbol{W}_{n_1}\times_{2}\boldsymbol{W}_{n_2}\in\mathbb{R}^{n_1\times n_2\times n_3}$. Equivalently, this can be reformulated as a frontal slice-wise expression $\tensor{B}(:,:,k) = \boldsymbol{W}_{n_1}\tensor{A}(:,:,k) \boldsymbol{W}_{n_2}^T.$ Thus, the HWT here can be interpreted as applying the 2-D HWT to each frontal slice of the tensor.
\end{definition}
Similar to Eq. \eqref{eq:block_B}, the tensor $\tensor{B}\in\mathbb{R}^{n_1\times n_2\times n_3}$ in Definition \ref{def_HWT_tensor} can be represented as a block tensor comprising four wavelet coefficient tensors $\tensor{B}_{1},\tensor{B}_{2},\tensor{B}_{3},\tensor{B}_{4}\in \mathbb{R}^{\frac{n_1}{2} \times \frac{n_2}{2} \times n_3}$:
\begin{equation}\small\label{eq:wavelet_coefficient_block}
    \begin{aligned}
        \tensor{B}=&\tensor{A}\times_{1}\boldsymbol{W}_{n_1}\times_{2}\boldsymbol{W}_{n_2} 
        \triangleq\left[ \begin{matrix}
            \tensor{B}_1&		\tensor{B}_2\\
            \tensor{B}_3&		\tensor{B}_4\\
        \end{matrix} \right] .
    \end{aligned}
\end{equation}
For simplicity, the 2-D frontal slice-wise HWT for a tensor ${\cal A}$ is denoted by $
[\tensor{B}_{1}, \tensor{B}_{2}; \tensor{B}_{3}, \tensor{B}_{4}] = \mathrm{HWT}(\tensor{A})$ throughout the manuscript,
and the 2-D IHWT for a tensor is denoted by $
\tensor{A}= \mathrm{IHWT}
([\tensor{B}_{1}, \tensor{B}_{2}; \tensor{B}_{3}, \tensor{B}_{4}])$.
\subsection{Haar Wavelet Transform-Induced CF-INR\label{sec_2}}
\begin{figure*}[t]
\centering
\includegraphics[width=.83\linewidth]{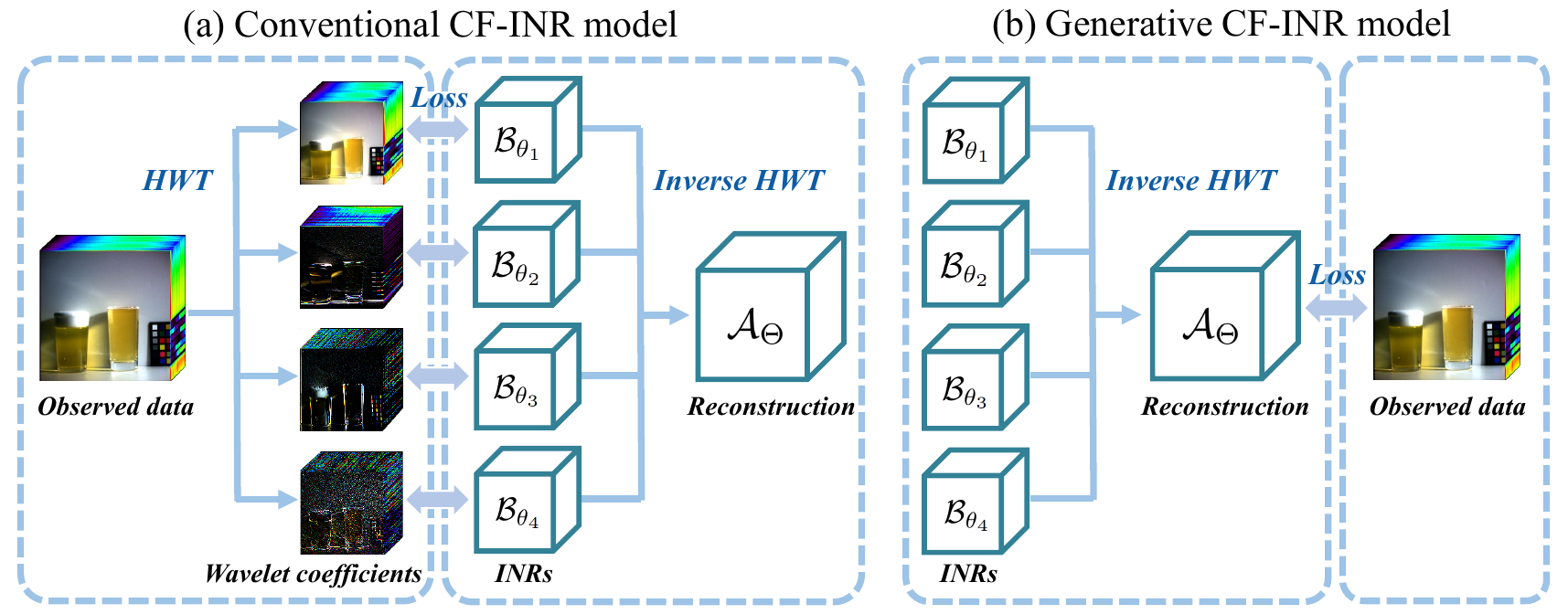}
\caption{(a) The conventional CF-INR optimization model \eqref{direct} for continuous data representation. (b) The proposed generative CF-INR optimization model \eqref{indirect} for continuous data representation, which solely uses the inverse HWT. \vspace{-0.37cm}}
\label{fig: two_different_models}
\end{figure*}
The INR uses a neural network to map coordinates to the corresponding values for continuous data representation. Given a tensor data ${\cal A}\in{\mathbb R}^{n_1\times n_2\times n_3}$ (e.g., an RGB image), the optimization model for learning an INR $f_\theta(\cdot):{\mathbb R}^{3}\rightarrow{\mathbb R}$ (with $\theta$ being the learnable parameters) is formulated as
\begin{equation}\small
\min_\theta\sum_{(i,j,k)}({\cal A}(i,j,k)-f_\Theta(i,j,k))^2.
\end{equation}
Typically, the INR $f_\theta(\cdot)$ is parameterized by an MLP that takes the coordinate $(i,j,k)$ as input and outputs the value. However, the spectral bias of MLP makes it challenging to accurately approximate the observed data ${\cal A}$ \cite{Fourier_feature}. Hence, the structure design of the network $f_\theta(\cdot)$ becomes important for INR-related applications. Although classical INR methods have made significant progress on this topic \cite{activation_function1,Fourier_feature,Wire,LRTFR}, they directly represent data in the original space where different frequency components are mixed. This makes it challenging to take a further step to accurately characterize each frequency component separately.\par

The main motivation of the proposed method is to use the HWT to effectively decouple the tensor data ${\tensor A}$ into different frequency components (including one low-frequency component $\tensor{B}_{1}$ (approximation component) and three high-frequency components $\tensor{B}_{2}$, $\tensor{B}_{3}$, and $\tensor{B}_{4}$ (horizontal, vertical, and diagonal components)) and then use distinct INRs to model these frequency components separately. The conventional way to achieve such a paradigm is to conduct the forward HWT for data $\cal A$, learn the wavelet coefficients using INRs, and then perform the inverse HWT. Specifically, one can consider the following optimization model for training four INRs:
\begin{equation}\small\label{direct}
\begin{split}
\min_{\{\theta_s\}_{s=1}^4}&\sum_{s=1,2,3,4}\sum_{(i,j,k)}({\cal B}_s(i,j,k)-f_{\theta_s}(i,j,k))^2,\\&\quad{\rm where}\;\;[\tensor{B}_{1}, \tensor{B}_{2}; \tensor{B}_{3}, \tensor{B}_{4}] = \mathrm{HWT}(\tensor{A}).
\end{split}
\end{equation}
This optimization model firstly decomposes the tensor $\cal A$ into different frequency components using the HWT and then uses four INRs $f_{\theta_1}(\cdot),f_{\theta_2}(\cdot),f_{\theta_3}(\cdot),f_{\theta_4}(\cdot):{\mathbb R}^3\rightarrow{\mathbb R}$ to fit the four wavelet coefficients respectively. The desired reconstruction can be obtained by performing the IHWT on the INR outputs. Such frequency decoupling allows the INRs to model different frequencies separately to enable higher precision.\par 
While it is valid to use the model \eqref{direct} for approximating a noise-less data $\cal A$, it would however lead to instability when approximating a noisy data $\cal A$ using the direct model \eqref{direct}. This is because conducting the HWT for a noisy data $\cal A$ would possibly disturb the semantic and structural information of the underlying clean data, making the INR approximation in the wavelet space less effective. For another example, when the observed tensor contains missing pixels, the HWT would blend the observed and missing values, affecting the effectiveness of continuous data representation. Hence, it is well-motivated to consider a direct generative model by omitting the forward HWT process and solely using the IHWT. Specifically, we firstly utilize four INRs to generate four tensors $\{{\cal B}_{\theta_s}\}_{s=1}^4$ (i.e., directly generating four wavelet coefficients using INRs) and then deploy the IHWT to obtain the desired output tensor:
\begin{equation}\small\label{indirect}
\begin{split}
\min_{\{\theta_s\}_{s=1}^4}&\left\lVert\mathrm{IHWT}([{\tensor{B}}_{\theta_1}, {\tensor{B}}_{\theta_2}; {\tensor{B}}_{\theta_3},{\tensor{B}}_{\theta_4}])-{\cal A}\right\rVert_F^2,\\&
\quad{\rm where}\;\;{\tensor{B}}_{\theta_s}(i,j,k)=f_{\theta_s}(i,j,k).
\end{split}
\end{equation}
Here, ${\tensor{B}}_{\theta_s}\in \mathbb{R}^{\frac{n_1}{2} \times \frac{n_2}{2} \times n_3}$ ($s=1,2,3,4$) can be interpreted as the generated wavelet coefficient tensor parameterized by the INR weights $\theta_s$. The model \eqref{indirect} firstly generates the four wavelet coefficient tensors $\{{\tensor{B}}_{\theta_s}\}_{s=1}^{4}$ using the INRs $\{f_{\theta_s}(\cdot)\}_{s=1}^4$ and then combines them using the inverse HWT (i.e., from the wavelet space to the original space). Subsequently, the INR weights ${\{\theta_s\}_{s=1}^4}$ are updated by minimizing the loss between the generated tensor and the observed data $\tensor{A}$. The generative optimization model \eqref{indirect} has two distinctions as compared to the conventional model \eqref{direct} (as intuitively shown in Fig. \ref{fig: two_different_models}). First, the generative model \eqref{indirect} omits the forward HWT and solely uses the IHWT. Second, the generative model \eqref{indirect} calculates the loss function in the original image space, while the conventional model \eqref{direct} calculates the loss function in the wavelet space. As a result, the generative model \eqref{indirect} allows the incorporation of a wide variety of degradation operators in the loss function (e.g., the mask operator for inpainting), allowing the model to tackle different image recovery tasks. Hence we mainly utilize the generative CF-INR model \eqref{indirect} due to its compatibility with general image recovery tasks. Numerical comparisons of the two models \eqref{direct} \& \eqref{indirect} are provided in Sec. \ref{compare_model}.\par 
The key advantage of the proposed CF-INR model \eqref{indirect} is its frequency decoupling characteristic, which enables the INRs to learn different frequency components separately, thus enjoying higher accuracy for continuous data representation. We provide numerical comparisons between the proposed CF-INR model \eqref{indirect} and some traditional INR methods \cite{activation_function1,Wire,Fourier_feature} for continuous data representation in Sec. \ref{sec: fitting} (An example is shown in Fig. \ref{fig:big_figure}). 
\subsection{Efficient Tensor Decomposition Paradigm for CF-INR\label{sec_3}}
\begin{figure}[t]
\centering
\includegraphics[width=.42\textwidth]{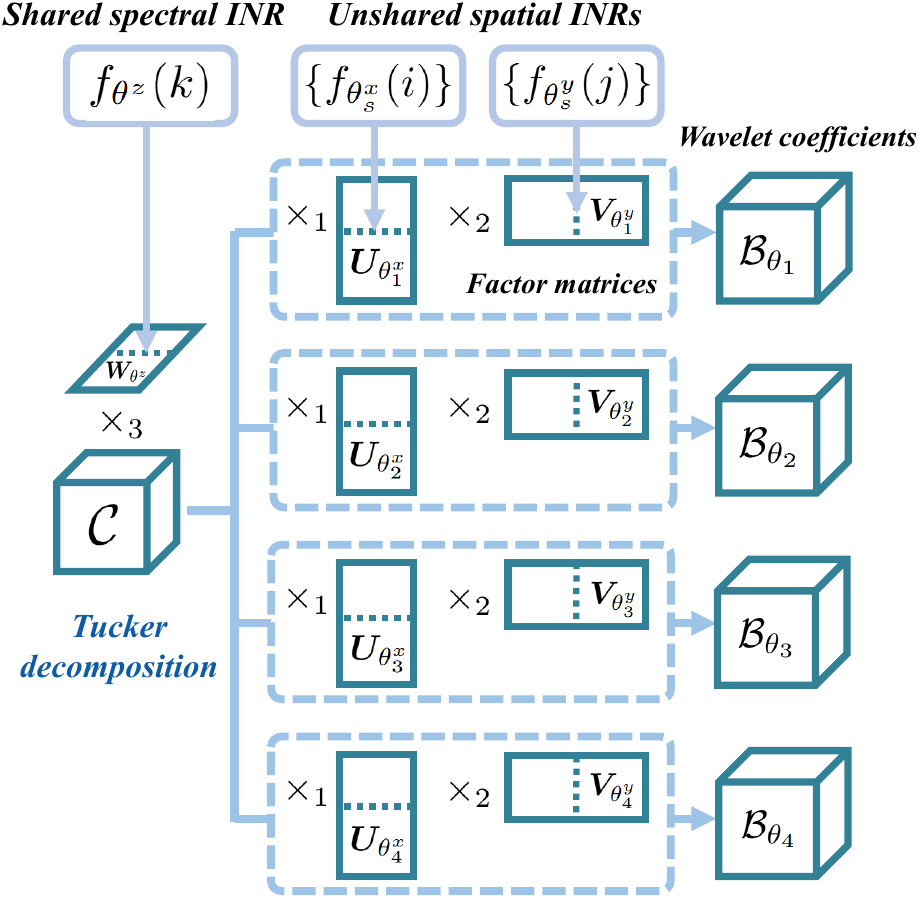}
\caption{Illustration of the proposed cross-frequency tensor decomposition with spectral coupling and spatial decoupling paradigms for CF-INR, which enjoys better computational efficiency and exploits both the intra- and inter-relationships among different wavelet coefficients.\vspace{-0.37cm}}
\label{fig: Diagram of CF-INR}
\end{figure}

Although four implicit MLPs $\{f_{\theta_s}(\cdot)\}_{s=1}^4$ (e.g., by using the MLP structure in \cite{Wire,activation_function1,FINER}) can be deployed to represent the four wavelet coefficients in the CF-INR model \eqref{indirect}, this direct implementation leads to two limitations. First, the increased number of MLP networks leads to increased computational costs under the CF-INR framework. Second, directly employing four MLPs to characterize the four wavelet coefficients neglects the intrinsic inter-relationships among different wavelet coefficients, since these MLPs are independently deployed for each wavelet coefficient. To address these issues, we propose an efficient cross-frequency tensor decomposition paradigm to represent wavelet coefficients under the CF-INR framework (see Fig. \ref{fig: Diagram of CF-INR} for a quick view), which holds better computational efficiency and exploits both the intra- and inter-relationships among different wavelet coefficient tensors to more accurately represent cross-frequency components.\par
\subsubsection{Cross-Frequency Tensor Decomposition for CF-INR} For each wavelet coefficient ${\cal B}_{\theta_s}\in{\mathbb R}^{\frac{n_1}{2}\times\frac{n_2}{2}\times n_3}$ ($s=1,2,3,4$), we employ a tensor Tucker decomposition (see Lemma \ref{Tucker}) \cite{Tucker,LRTFR} to decompose  ${\cal B}_{\theta_s}$ into a core tensor ${\cal C}_s\in{\mathbb R}^{r_1\times r_2\times r_3}$ (where $R\triangleq(r_1,r_2,r_3)$ denotes the Tucker rank) and three factor matrices $\boldsymbol{U}_{\theta_s^x}\in{\mathbb R}^{\frac{n_1}{2}\times r_1},\boldsymbol{V}_{\theta_s^y}\in{\mathbb R}^{\frac{n_2}{2}\times r_2},\boldsymbol{W}_{\theta_s^z}\in{\mathbb R}^{n_3\times r_3}$, and then use three univariate lightweight INRs $f_{\theta_s^x}:{\mathbb R}\rightarrow{\mathbb R}^{r_1},f_{\theta_s^y}:{\mathbb R}\rightarrow{\mathbb R}^{r_2},f_{\theta_s^z}:{\mathbb R}\rightarrow{\mathbb R}^{r_3}$ to parameterize these factor matrices:
\begin{equation}\small\label{LRTFR_all}
\begin{split}
&{\cal B}_{\theta_s} = {\cal C}_s\times_3{\boldsymbol{W}}_{\theta_s^z}\times_2{\boldsymbol{V}}_{\theta_s^y}\times_1{\boldsymbol{U}}_{\theta_s^x},\;s=1,2,3,4,\\
&{\boldsymbol{U}}_{\theta_s^x}(i,:)=f_{\theta_s^x}(i),i=1,2,\cdots,\frac{n_1}{2},\\
&{\boldsymbol{V}}_{\theta_s^y}(j,:)=f_{\theta_s^y}(j),j=1,2,\cdots,\frac{n_2}{2},\\
&{\boldsymbol{W}}_{\theta_s^z}(k,:)=f_{\theta_s^z}(k),k=1,2,\cdots,n_3.\\
\end{split}
\end{equation}
Here ${\boldsymbol{U}}_{\theta_s^x},{\boldsymbol{V}}_{\theta_s^y},{\boldsymbol{W}}_{\theta_s^z}$ denote the factor matrices parameterized by the INRs $f_{\theta_s^x}(\cdot)$, $f_{\theta_s^y}(\cdot)$, and $f_{\theta_s^z}(\cdot)$. The ${\cal B}_{\theta_s}$ is the generated wavelet coefficient tensor obtained by the tensor decomposed INRs. The learnable parameters of such a representation are ${\theta_s}\triangleq\{{\cal C}_s,\theta_s^x,\theta_s^y,\theta_s^z\}$. Each univariate INR (for example $f_{\theta_s^x}(\cdot)$) is an MLP in the form of 
\begin{equation}\small\label{MLP}
f_{\theta_{s}^x}(x) = \boldsymbol{H}_d(\sigma_s(\boldsymbol{H}_{d-1} \cdots \sigma_s(\boldsymbol{H}_{1}x))) : {\mathbb R} \rightarrow \mathbb{R}^{r_1},
\end{equation}
where $\theta_{s}^{x}\triangleq\{\boldsymbol{H}_q\}_{q=1}^d$ are learnable weights and $\sigma_s(\cdot)\triangleq\sin(\omega_s\cdot)$ denotes the sinusoidal activation function with a frequency parameter $\omega_s$ \cite{activation_function1}. Although multiple MLPs are used for representing $\{{\cal B}_{\theta_s}\}_{s=1}^4$, a smaller MLP depth (for example $d=2$) usually suffices to obtain satisfactory performances in our framework. Here, the cross-frequency parameters for the four wavelet coefficients (denoted by $\{\omega_s\}_{s=1}^4$ or $\omega_1$-$\omega_4$) determine the frequency preference of the CF-INR model. It is better to make distinguishments for the four frequency parameters $\{\omega_s\}_{s=1}^4$ to better characterize multi-frequency nature across different wavelet coefficients $\{{\cal B}_{\theta_s}\}_{s=1}^4$. We discuss the selection of cross-frequency parameters and corresponding self-evolving strategies in Sec. \ref{sec_5}. 
\par 
Equivalently, the model \eqref{LRTFR_all} can be re-formulated as a more concise element-wise representation:
\begin{equation}\small\label{LRTFR}
\begin{split}
{\cal B}_{\theta_s}(i,j,k) = {\cal C}_s\times_3f_{\theta_s^z}(k)\times_2f_{\theta_s^y}(j)\times_1f_{\theta_s^x}(i).
\end{split}
\end{equation}
Hence, the tensor decomposed CF-INR \eqref{LRTFR} essentially forms a coordinate-to-value mapping and can be viewed as a continuous representation of data similar to conventional INR methods \cite{activation_function1,Fourier_feature}.
The tensor decomposed CF-INR incorporates the low-rankness (see Lemma \ref{th: Relationship CF-Rank}) into the model (i.e., the rank $(r_1,r_2,r_3)$ is usually set as smaller values compared to the data size $(n_1,n_2,n_3)$). Natural images are born with low-dimensional characteristics (i.e., the property that an image can be represented by the linear combination of a limited number of bases, or the so-called low-rankness property). Hence the low-rank tensor decomposed INR is helpful to characterize such low-dimensional features of images, especially helpful for data recovery tasks (such as denoising and inpainting) where low-rankness plays as an essential prior \cite{LRTDTV,lu2019low}. 
\subsubsection{Spectrally Coupled and Spatially Decoupled Paradigm}\label{SCSD} To further enable the cross-frequency tensor decomposition \eqref{LRTFR_all} to characterize inter-relationships among different wavelet coefficients $\{{\cal B}_{\theta_s}\}_{s=1}^4$, our architecture employs shared core tensor ${\cal C}$ and spectral factor matrix ${\boldsymbol W}$ for different wavelet coefficients, while maintaining distinct spatial factor matrices ${\boldsymbol U},{\boldsymbol V}$ unshared (see Fig. \ref{fig: Diagram of CF-INR}). This sharing mechanism leads to the following spectrally coupled and spatially decoupled cross-frequency tensor representations:
\begin{equation}\small\label{LRTFR_share}
\begin{split}
&{\cal B}_{\theta_s} = \underbrace{{\cal C}\times_3{\boldsymbol{W}}_{\theta^z}}_{\rm Shared}\times_2\underbrace{{\boldsymbol{V}}_{\theta_s^y}\times_1{\boldsymbol{U}}_{\theta_s^x}}_{\rm Unshared},\;s=1,2,3,4,\\
&{\boldsymbol{U}}_{\theta_s^x}(i,:)=f_{\theta_s^x}(i),i=1,2,\cdots,\frac{n_2}{2},\\
&{\boldsymbol{V}}_{\theta_s^y}(j,:)=f_{\theta_s^y}(j),j=1,2,\cdots,\frac{n_2}{2},\\
&{\boldsymbol{W}}_{\theta^z}(k,:)=f_{\theta^z}(k),k=1,2,\cdots,n_3.\\
\end{split}
\end{equation}
Especially, we have used the same core tensor ${\cal C}\in{\mathbb R}^{r_1\times r_2\times r_3}$ and the spectral INR $f_{\theta^z}(\cdot)$ for different wavelet coefficients $\{{\cal B}_{\theta_s}\}_{s=1}^4$, while used different spatial INRs $\{f_{\theta_s^x}(\cdot)\}_{s=1}^4$ and $\{f_{\theta_s^y}(\cdot)\}_{s=1}^4$ for different wavelet coefficients. This design choice stems from the inherent properties of the HWT. As formalized in Definition \ref{def_HWT_tensor}, the HWT operates through frontal slice-wise decomposition, redistributing spatial information into distinct frequency components while maintaining spectral coherence across wavelet coefficients. Hence we keep the spatial INRs unshared while sharing the spectral INR\footnote{We can theoretically validate such sharing strategy by Theorem \ref{th: Smoothness for the
Wavelet Coefficient}, where the spectral smoothness bound of different wavelet coefficients is the same, while the spatial smoothness bound of different wavelet coefficients varies.}. The spectrally coupled and spatially decoupled tensor decomposed CF-INR in theory characterizes both the intra- and inter-relationships of different wavelet coefficients (i.e., the similarity among different wavelet coefficients and the uniqueness of each wavelet coefficient). Such optimized architecture improves the performance of CF-INR for data recovery by leveraging the intrinsic knowledge sharing among different wavelet coefficients (see Table \ref{table: Architecture Variant} for examples).
\subsubsection{Efficiency Analysis for CF-INR} The proposed cross-frequency tensor decomposed CF-INR holds better computational efficiency as compared with conventional MLP-based INRs \cite{activation_function1,Wire}. To represent a tensor of size $n_1\times n_2\times n_3$, an MLP-based INR needs to query $n_1n_2n_3$ coordinates in the form of $(x,y,z)$ to generate the tensor, which results in a total number of $n_1n_2n_3$ forward passes. The computational complexity of the MLP-based INR is $O(m^2dn_1n_2n_3)$ for representing the tensor, where $m$ denotes the network width and $d$ denotes the depth. As compared, CF-INR with the tensor decomposition structure has much lower computational complexity. Specifically, to represent each wavelet coefficient tensor of size $\frac{n_1}{2}\times \frac{n_2}{2}\times{n_3}$, the univariate INRs only need to query $\frac{n_1}{2}+\frac{n_2}{2}+{n_3}$ coordinates under the tensor decomposition paradigm \eqref{LRTFR_share}. Note that the complexity of the Tucker decomposition is $O(rn_1n_2n_3)$ and the complexity of the IHWT is $O(n_1n_2n_3)$, hence the total complexity of CF-INR for representing the tensor is $O(4m^2d(\frac{n_1}{2}+\frac{n_2}{2}+{n_3})+(r+1)n_1n_2n_3)$, which is much lower than that of the MLP-based INR, i.e., $O(m^2dn_1n_2n_3)$ (since $(r+1)$ is usually much smaller than $m^2d$). Consequently, CF-INR holds better computational efficiency for continuous data representation (see Table \ref{table: FLOPs}).\par 
\begin{table}[tbp]
\scriptsize
\caption{The computational complexity, FLOPs, and running time (5000 iterations) of different INR methods for image regression with an image of size $n_1=n_2=256,n_3=31$.}
\label{table: FLOPs}
\centering
\setlength{\tabcolsep}{0.3pt}
\begin{tabular}{cccc}
\hline
Method     & Computational complexity  & FLOPs & Running time     \\ \hline
PE\cite{Fourier_feature}        & $O(m^2dn_1n_2n_3)$      & 14109.6M         & 95.9s         \\
WIRE\cite{Wire}      & $O(m^2dn_1n_2n_3)$         & 6832.5M       & 402.5s       \\
SIREN\cite{activation_function1}     & $O(m^2dn_1n_2n_3)$       & 13438.6M        & 117.2s       \\
CF-INR   & $O(4m^2d(\frac{n_1}{2}+\frac{n_2}{2}+{n_3})+(r+1)n_1n_2n_3)$     & 103.2M           & 36.6s         \\ \hline
\end{tabular}\vspace{-0.3cm}
\end{table}
The cross-frequency tensor decomposition \eqref{LRTFR_share} enables more meticulous characterizations of multi-frequency components. Especially, by tailoring different cross-frequency
ranks ($R_1$-$R_4$) and cross-frequency parameters (i.e., the sinusoidal activation function parameters $\omega_1$-$\omega_4$)\footnote{The cross-frequency parameters include (i) the shared spectral frequency parameter $\omega^z$ for the sinusoidal activation function of the shared spectral INR $f_{\theta^z}(\cdot)$, and (ii) the unshared spatial frequency parameters $\{\omega_s\}_{s=1}^4$ for the unshared spatial INRs $\{f_{\theta_s^x}(\cdot)\}_{s=1}^4$ and $\{f_{\theta_s^y}(\cdot)\}_{s=1}^4$ (with $f_{\theta_s^x}(\cdot)$ and $f_{\theta_s^y}(\cdot)$ deploying the same frequency parameter $\omega_s$).} for different wavelet coefficients ${\cal B}_{\theta_1}$-${\cal B}_{\theta_4}$, CF-INR holds
intrinsic superiority for heterogeneous cross-frequency feature characterization. Next, we provide theoretical guidance on how to effectively (and automatically) select these cross-frequency parameters.
\subsection{Self-Evolving Cross-Frequency Low-Rankness}\label{sec_4_rank}
\subsubsection{Cross-Frequency Tensor Rank} 
\begin{figure}[t]
\setlength{\tabcolsep}{0pt}
\begin{tabular}{cc}
\includegraphics[width=0.245\textwidth]{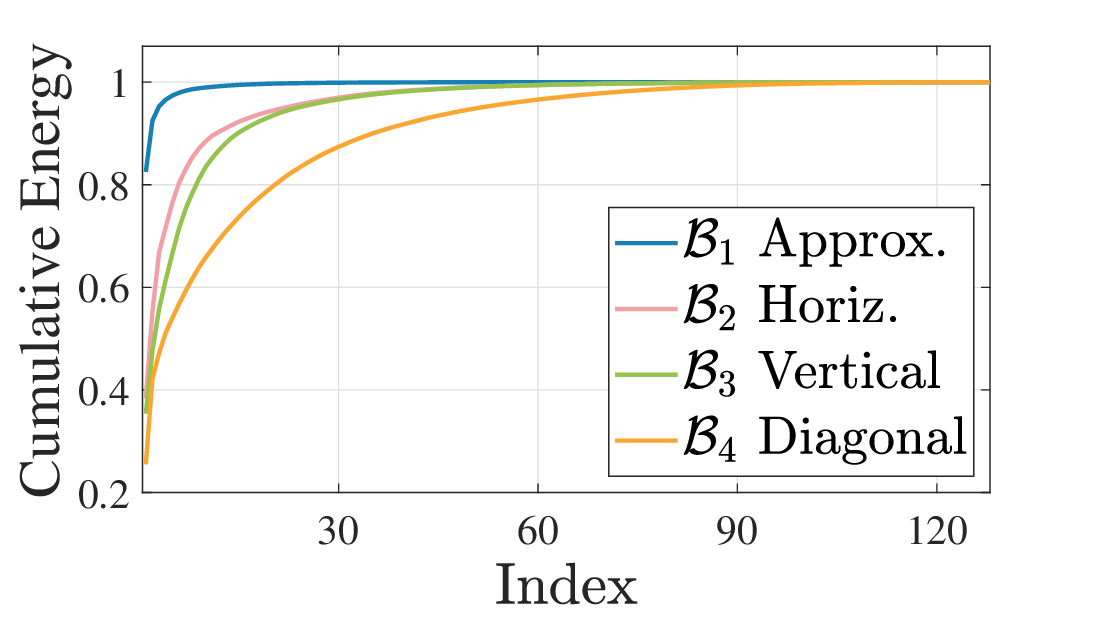}&
\includegraphics[width=0.245\textwidth]{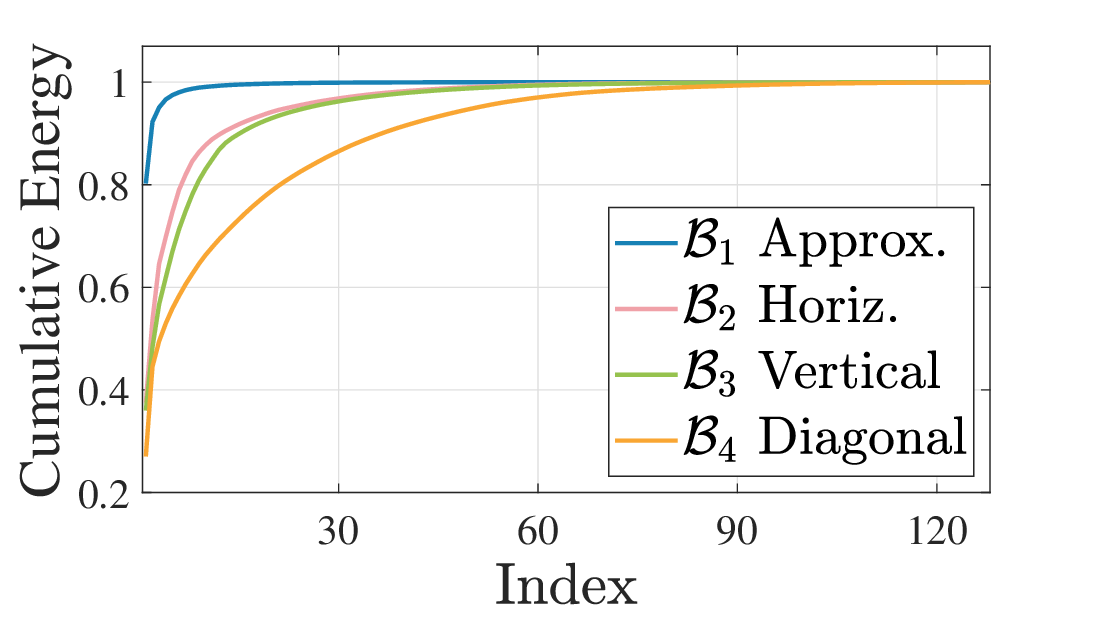}\\
(a) Dimension $x$ &(b) Dimension $y$\\
\end{tabular}
\caption{Cumulative energy curves (defined by $\mathrm{CE}(k)={\sum_{i=1}^k \sigma_i}/{\sum_{i=1}^n \sigma_i}$, where $\sigma_i$ denotes the $i$-th singular value) of the unfolded wavelet coefficient matrices along the $x$ and $y$ spatial dimensions of an image.\vspace{-0.5cm}
}
    \label{fig: CE}
\end{figure}

Intuitively, different frequency components $\{{\cal B}_s\}_{s=1}^4$ often exhibit varying low-rankness (see Fig. \ref{fig: CE}), and the approximation coefficient tends to exhibit stronger low-rank characteristics. To quantitatively evaluate the low-rankness of different frequency components, we introduce the following concept of the {\bf cross-frequency tensor rank} by using HWT.
\begin{definition}
For a tensor $\tensor{A} \in \mathbb{R}^{n_1 \times n_2 \times n_3}$, its {\bf cross-frequency tensor rank} (CF-Rank, denoted by ${\rm rank_{CF}}(\tensor{A})$) is a matrix defined as the Tucker rank of its wavelet coefficients: 
\begin{equation}\small
\mathrm{rank}_{\rm CF}(\tensor{A}) = 
\begin{bmatrix}
\mathrm{rank}_T({\cal B}_1)\\
\mathrm{rank}_T({\cal B}_2)\\
\mathrm{rank}_T({\cal B}_3)\\
\mathrm{rank}_T({\cal B}_4)
\end{bmatrix}
\in{\mathbb R}^{4\times 3},
\end{equation}
where $[\tensor{B}_{1}, \tensor{B}_{2}; \tensor{B}_{3}, \tensor{B}_{4}] = \mathrm{HWT}(\tensor{A})$ and $\mathrm{rank}_T({\cal B}_s)\in{\mathbb R}^{1\times 3}$ denotes the Tucker rank of ${\cal B}_s\;(s=1,2,3,4)$.
\end{definition}
\begin{lemma}[Relationship between CF-Rank and Tucker rank]
\label{th: Relationship CF-Rank}
Let $\mathcal{A}\in \mathbb{R}^{n_1\times n_2\times n_3}$. Denote $\mathrm{rank}_{\rm CF}(\mathcal{A})\in\mathbb{R}^{4\times 3}$ and $\mathrm{rank}_{T}(\mathcal{A})\in\mathbb{R}^{1\times 3}$ the CF-Rank and Tucker rank of the tensor $\mathcal{A}$. Then, for $s=1,2,3,4$, $n=1,2,3$, we have $\mathrm{rank}_{\rm CF}({\cal A})(s,n) \leq \mathrm{rank}_{T}({\cal A})(n)$.
\end{lemma}
Lemma \ref{th: Relationship CF-Rank} shows that if the observed data ${\cal A}$ is low-rank, its wavelet coefficients also exhibit low-rankness. This result supports the rationale for adopting the tensor decomposed CF-INR \eqref{LRTFR}. Meanwhile, determining the CF-Rank is a crucial step for setting appropriate rank parameters in the CF-INR model \eqref{LRTFR_share}, since different wavelet coefficients usually exhibit diverse low-rank characteristics. Here, we establish a theoretical convex envelope of the CF-Rank that can serve as an effective quantitative measure of the CF-Rank for automatic rank parameters updating during optimization.
\begin{lemma}[The convex envelope of CF-Rank]
\label{th: Convex envelope CF}
Let $\tensor{A} \in \mathbb{R}^{n_1 \times n_2 \times n_3}$ and $[\tensor{B}_1,\tensor{B}_2;\tensor{B}_3,\tensor{B}_4]=\mathrm{HWT}(\tensor{A})$. Suppose the largest singular value of each unfolding matrix satisfies
$\sigma_{\max}({\bm{B}_{s}}_{(i)})\le 1$ ($s=1,2,3,4,\;i=1,2,3$). Then the {\bf cross-frequency nuclear norm} defined by $
\|\tensor{A}\|_{\mathrm{CFNN}}\triangleq\sum_{s=1}^{4}\sum_{i=1}^{3}\|{\bm{B}_{s}}_{(i)}\|_*$ is the convex envelope of the maximum CF-Rank $\max_{s,n}\mathrm{rank}_{\rm CF}(\tensor{A}){(s,n)}$ over the set ${\cal D}=\{\tensor{A}:\sigma_{\max}({\bm{B}_{s}}_{(i)})\le 1$ for $s=1,2,3,4,\;i=1,2,3\}$.
\end{lemma}
\subsubsection{Self-Evolving Strategy for Cross-Frequency Rank}
By using the cross-frequency nuclear norm, we aim to automatically update the cross-frequency ranks $r_{x_1}$-$r_{x_4}$ and $r_{y_1}$-$r_{y_4}$, where $r_{x_s}$ and $r_{y_s}$ denote the two spatial ranks of the $s$-th wavelet coefficient in the tensor decomposition \eqref{LRTFR_share}. For the spectral rank $r_z$, we keep it shared across different wavelet coefficients. For simplicity we denote $R_{s}\triangleq(r_{x_s},r_{y_s},r_z)$.\par 
The motivation of the cross-frequency rank self-evolution is to use the intermediate outputs of CF-INR during optimization to guide the adjustments for $r_{x_1}$-$r_{x_4}$ and $r_{y_1}$-$r_{y_4}$ progressively using the cross-frequency nuclear norm. Let $\{{\cal B}_{\theta_s}\in{\mathbb R}^{\frac{n_1}{2}\times \frac{n_2}{2}\times n_3}\}_{s=1}^4$ be the intermediate output wavelet coefficients learned by CF-INR at a certain iteration. To ensure that the unfolding matrices satisfy $\sigma_{\max}\le 1$ in Lemma \ref{th: Convex envelope CF}, we first normalize them by dividing by their Frobenius norm, e.g., $
{\hat{\bm{B}}}_{{\theta_s}_{(1)}}={{\bm B}_{\theta_s}}_{(1)}/{\|{{\bm B}_{\theta_s}}_{(1)}\|_F}$. Then inspired by the convex envelope relationship established in Lemma \ref{th: Convex envelope CF}, we make the following key hypothesis: the nuclear norm of the normalized matrices should share consistent relative relationships with cross-frequency ranks, that is we set
    \begin{equation}\small\label{update_rank}
    \begin{aligned}
r_{x_s}/r_{x_{s'}}&\leftarrow\|{\hat{\bm{B}}}_{{\theta_s}_{(1)}}\|_*/\|{\hat{\bm{B}}}_{{\theta_{s'}}_{(1)}}\|_*\\
r_{y_s}/r_{y_{s'}}&\leftarrow\|{\hat{\bm{B}}}_{{\theta_s}_{(2)}}\|_*/\|{\hat{\bm{B}}}_{{\theta_{s'}}_{(2)}}\|_*,
    \end{aligned}
    \end{equation}
for $s,s'=1,2,3,4$. This can be equivalently expressed as $r_{x_1}:r_{x_2}:r_{x_3}:r_{x_4}\leftarrow
\|{\hat{\bm{B}}}_{{\theta_1}_{(1)}}\|_*:\|{\hat{\bm{B}}}_{{\theta_2}_{(1)}}\|_*:\|{\hat{\bm{B}}}_{{\theta_3}_{(1)}}\|_*:\|{\hat{\bm{B}}}_{{\theta_4}_{(1)}}\|_*$ and likewise for the $y$ dimension. The rationale behind this hypothesis is that, as iterations proceed, the learned wavelet coefficients $\{{\cal B}_{\theta_s}\}_{s=1}^4$ progressively converge toward more accurate representations, thereby enabling the corresponding cross-frequency nuclear norm to more precisely reveal the desired cross-frequency ranks of wavelet coefficients. In practice, we may further introduce a power parameter $k$ and update the ranks according to $\|{\hat{\bm{B}}}_{{\theta_s}_{(1)}}\|_*^{1/k}/\|{\hat{\bm{B}}}_{{\theta_{s'}}_{(1)}}\|_*^{1/k}$ to ensure balancing. In our implementations $k=3$ works well for all cases. 
Consequently, we automatically update the rank parameters $r_{x_1}$-$r_{x_4}$ and $r_{y_1}$-$r_{y_4}$ based on the hypothesis \eqref{update_rank} and the constraints $\sum_{s=1}^4r_{x_s}=\lambda_x$ and $\sum_{s=1}^4r_{y_s}=\lambda_y$, where $\lambda_x$ and $\lambda_y$ are summing hyperparameters (i.e., we reduce the four rank parameters $r_{x_1}$-$r_{x_4}$ into a single hyperparameter $\lambda_x$ indicating their sum, and our method is robust to $\lambda_x$ and $\lambda_y$; see Fig. \ref{fig: hyperparameter}). This leads to the following self-evolving update rule for cross-frequency ranks $r_{x_s}$ and $r_{y_s}$ ($s=1,2,3,4$):
\begin{equation}\small\label{auto_rank}
r_{x_s}\leftarrow\frac{\lambda_x\|{\hat{\bm{B}}}_{{\theta_s}_{(1)}}\|_*^{\frac{1}{k}}}{\sum_{s'=1}^4\|{\hat{\bm{B}}}_{{\theta_{s'}}_{(1)}}\|_*^{\frac{1}{k}}}\;,\;
r_{y_s}\leftarrow\frac{\lambda_y\|{\hat{\bm{B}}}_{{\theta_s}_{(2)}}\|_*^{\frac{1}{k}}}{\sum_{s'=1}^4\|{\hat{\bm{B}}}_{{\theta_{s'}}_{(2)}}\|_*^{\frac{1}{k}}}.
\end{equation}
We update the cross-frequency ranks $\{R_s=(r_{x_s},r_{y_s},r_{z})\}_{s=1}^4$ once for every 500 iterations using \eqref{auto_rank}. \par
Finally, we illustrate how to efficiently alter the model configuration of CF-INR using the updated cross-frequency ranks $R_1$-$R_4$. Since we have used a shared core tensor ${\cal C}\in{\mathbb R}^{r_1\times r_2\times r_3}$ in \eqref{LRTFR_share}, it is unlikely to directly use this shared core tensor ${\cal C}$ for different wavelet coefficients $\{{\cal B}_{\theta_s}\}_{s=1}^4$ with different ranks $R_1$-$R_4$. To address this problem, we propose a masking strategy that preserves the top-$R_s$ elements of the core tensor and masks the others with zeros when generating the $s$-th wavelet coefficient ${\cal B}_{\theta_s}$, which guarantees the rank-$R_s$ property of ${\cal B}_{\theta_s}$ while allowing the wavelet coefficients to still share a single core tensor. This strategy cleverly avoids the conflict of multi-scale ranks and makes the rank self-evolution algorithm subtly compatible with the proposed spectrally coupled tensor decomposition paradigm.
\subsection{Self-Evolving Cross-Frequency Smoothness}\label{sec_5_Lap}
\subsubsection{Cross-Frequency Laplacian Smoothness}
\begin{figure}[t]
    \centering
    \scriptsize
    \captionsetup[subfloat]{labelformat=empty}
    \subfloat{ 
    \raisebox{0.28cm}{\includegraphics[width=0.15\linewidth]{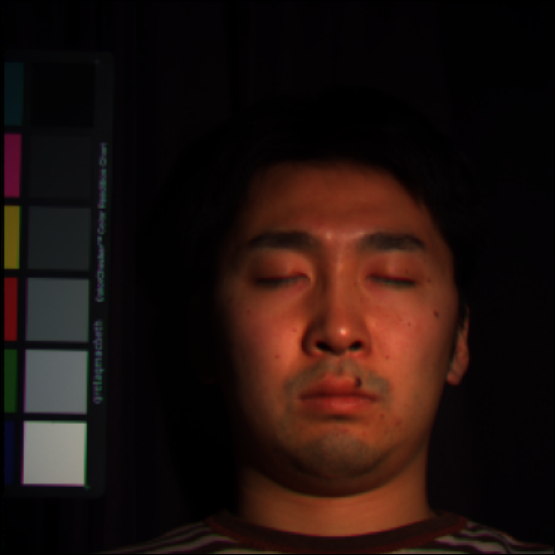} 
    }
    \label{fig: rgb_image_1}
    } 
    \hspace{-0.2cm}
    \subfloat{ \includegraphics[width=0.2\linewidth]{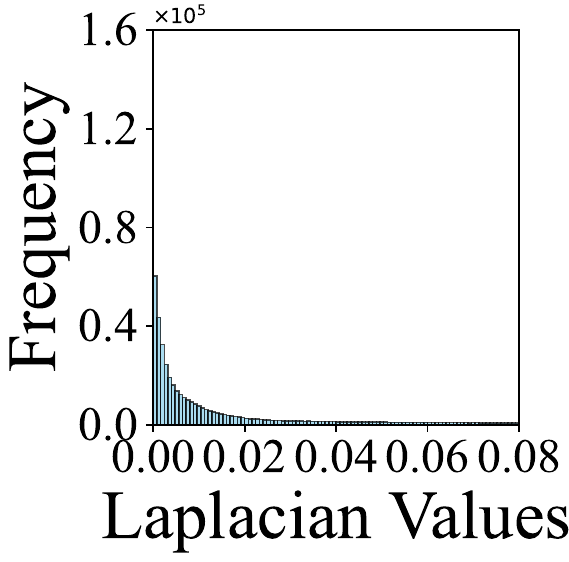} \label{fig:hist_LL_1}}  \hspace{-0.25cm}
    \subfloat{ \includegraphics[width=0.2\linewidth]{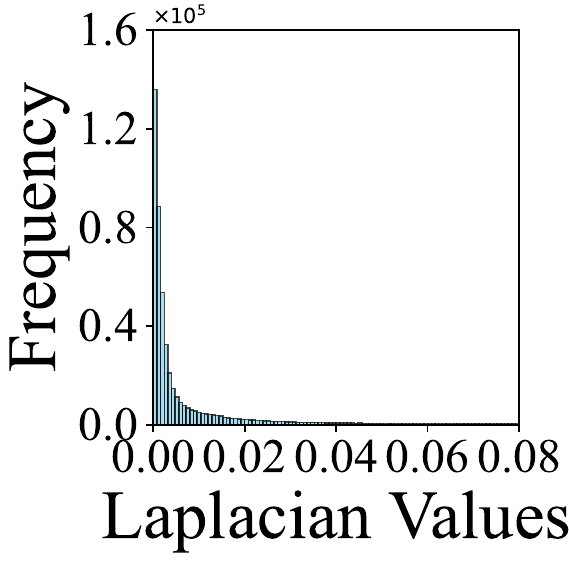} \label{fig:hist_LH_1}}  \hspace{-0.25cm}
    \subfloat{ \includegraphics[width=0.2\linewidth]{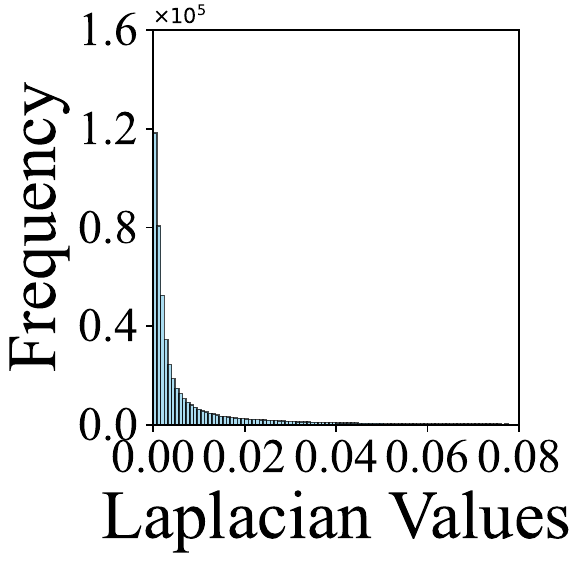} \label{fig:hist_HL_1}}  \hspace{-0.25cm}
    \subfloat{ \includegraphics[width=0.2\linewidth]{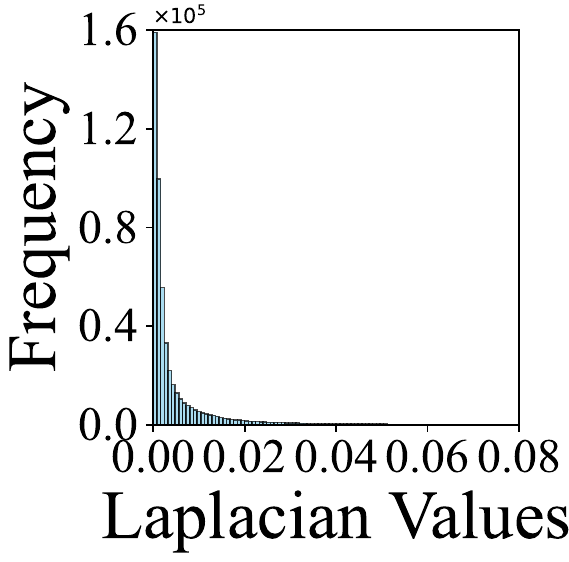} \label{fig:hist_HH_1}}
    \vskip -0.37cm
    \subfloat{ 
    \raisebox{0.28cm}{
        \includegraphics[width=0.15\linewidth]{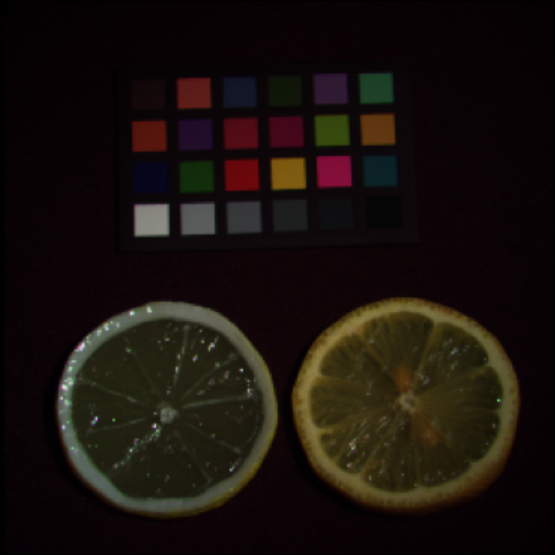} 
    }
    \label{fig: rgb_image_2}
    }   \hspace{-0.2cm}
    \subfloat{ \includegraphics[width=0.2\linewidth]{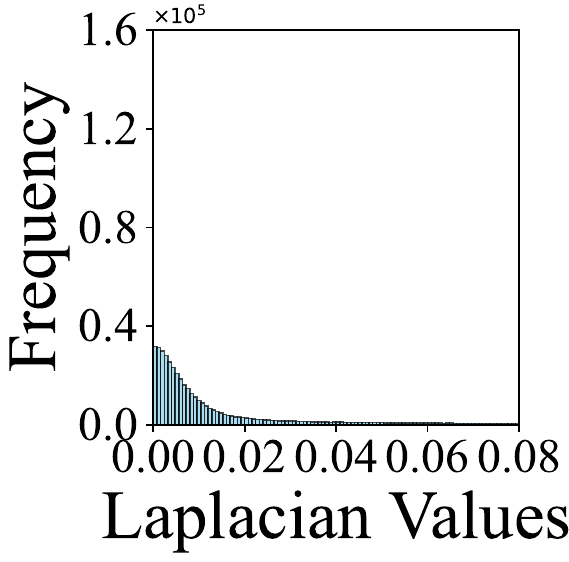} \label{fig:hist_LL_2}}  \hspace{-0.25cm}
    \subfloat{ \includegraphics[width=0.2\linewidth]{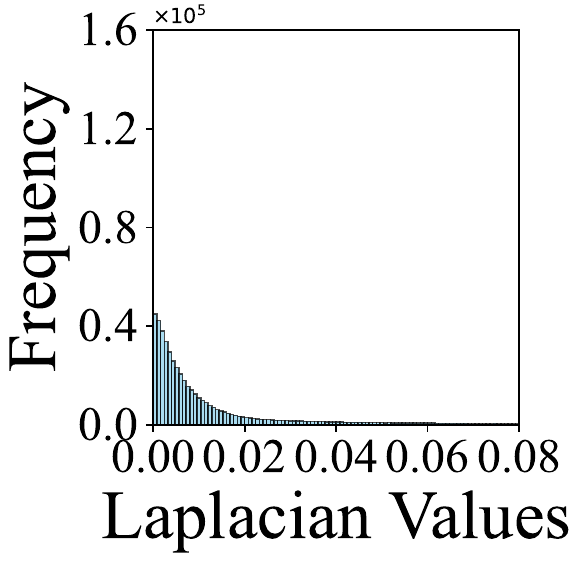} \label{fig:hist_LH_2}}  \hspace{-0.25cm}
    \subfloat{ \includegraphics[width=0.2\linewidth]{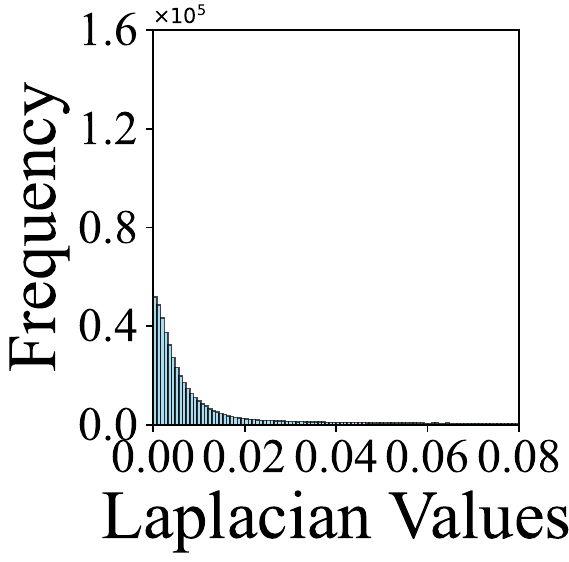} \label{fig:hist_HL_2}}  \hspace{-0.25cm}
    \subfloat{ \includegraphics[width=0.2\linewidth]{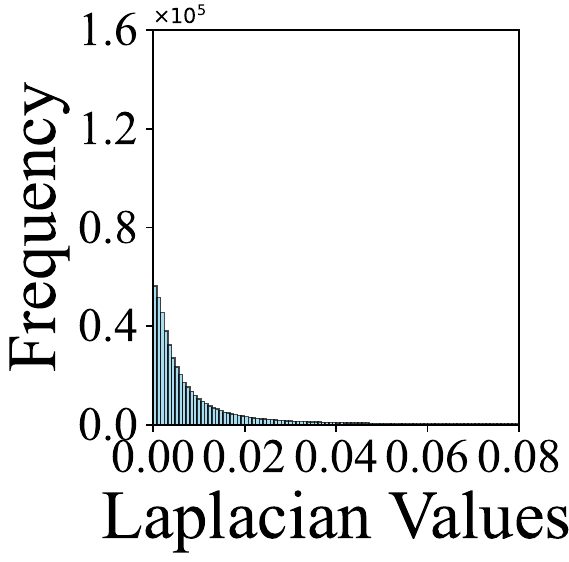} \label{fig:hist_HH_2}}
    \vskip -0.37cm
    \subfloat{ 
    \raisebox{0.28cm}{
        \includegraphics[width=0.15\linewidth]{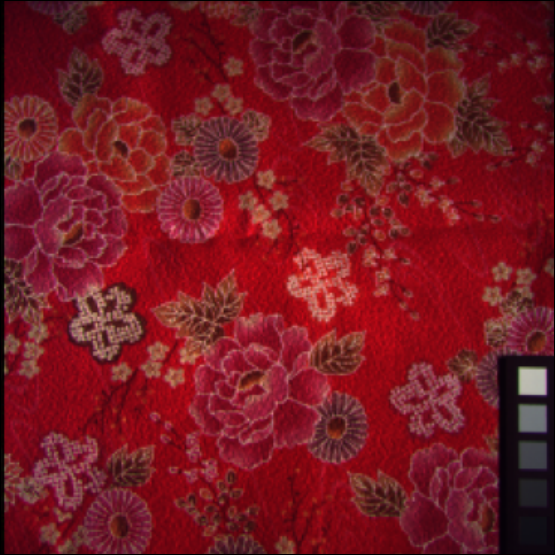} 
    }
    \label{fig: rgb_image_3}
    }   \hspace{-0.2cm}
    \subfloat[\hspace{1.7em}Approx. ${\cal B}_1$]{ \includegraphics[width=0.2\linewidth]{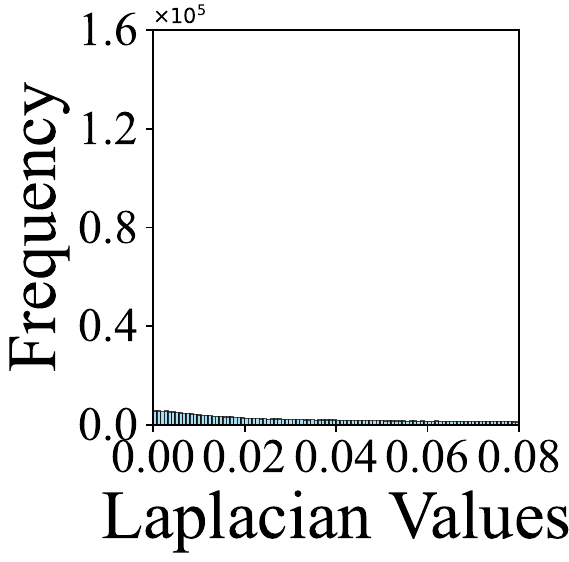} \label{fig:hist_LL_3}}  \hspace{-0.25cm}
    \subfloat[\hspace{1.7em}Horiz. ${\cal B}_2$]{ \includegraphics[width=0.2\linewidth]{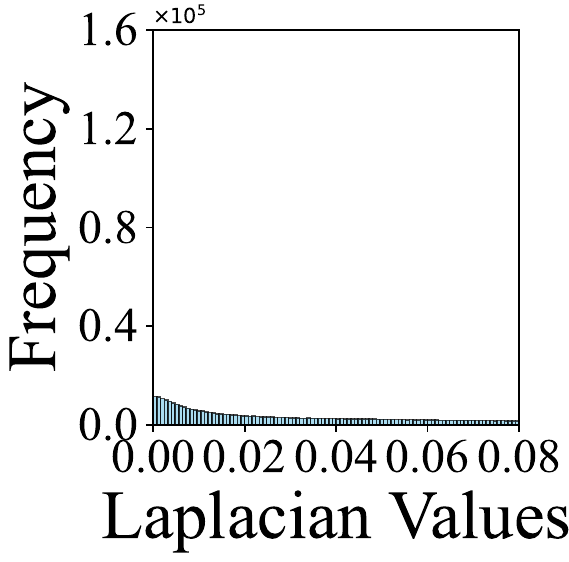} \label{fig:hist_LH_3}}  \hspace{-0.25cm}
    \subfloat[\hspace{1.7em}Vertical ${\cal B}_3$]{ \includegraphics[width=0.2\linewidth]{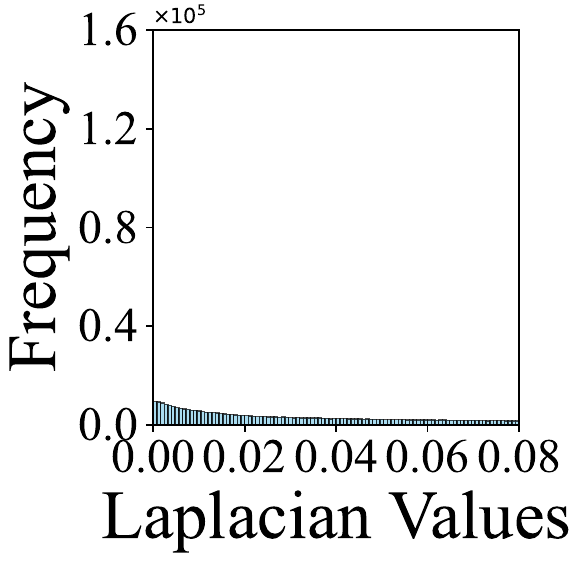} \label{fig:hist_HL_3}}  \hspace{-0.25cm}
    \subfloat[\hspace{1.7em}Diagonal ${\cal B}_4$]{ \includegraphics[width=0.2\linewidth]{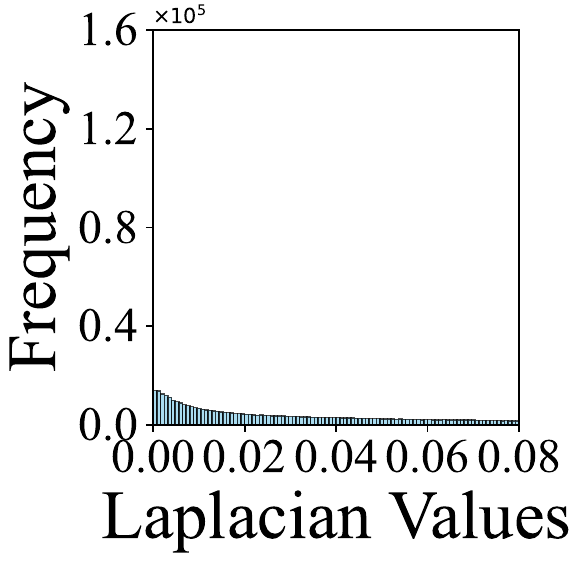} \label{fig:hist_HH_3}}
    \caption{Histograms of the Laplacian values derived from the wavelet coefficients of different images \textit{Face}, \textit{Cups}, and \textit{Cloth}.\vspace{-0.37cm}}
    \label{fig:Laplacian_for_Wavelet_Coefficients}
\end{figure}

Natural images usually enjoy a certain degree of local smoothness. Here, we are interested in how the wavelet coefficients $\{{\cal B}_s\}_{s=1}^4$ inherit such smoothness from the original image. We first deduce the following smoothness bound of wavelet coefficients of a tensor, which demonstrates that different wavelet coefficients inherit different degrees of smoothness, reflecting the necessity for delicate characterizations for cross-frequency smoothness.
\begin{theorem}[Smoothness bound for wavelet coefficients of a tensor]\label{th: Smoothness for the
Wavelet Coefficient}
Let $\tensor{A} \in \mathbb{R}^{n_1 \times n_2 \times n_3}$ and $L_1=\max_{i,j,k}|\tensor{A} (i+1,j,k)-\tensor{A} (i,j,k)|,L_2=\max_{i,j,k}|\tensor{A} (i,j+1,k)-\tensor{A} (i,j,k)|,L_3=\max_{i,j,k}|\tensor{A}(i,j,k+1)-\tensor{A} (i,j,k)|$. Let $
[\tensor{B}_{1}, \tensor{B}_{2}; \tensor{B}_{3}, \tensor{B}_{4}] = \mathrm{HWT}(\tensor{A})$. Then we have the following smoothness bounds for the wavelet coefficients:
\begin{equation}\small\label{eq:Bound for the
Wavelet Coefficient}
\begin{aligned}
&{\rm (Dimension\;}x{\rm)}\\
&\;\;\max_{i,j,k}|\tensor{B}_1 (i+1,j,k)-\tensor{B}_1 (i,j,k)|\leq 4L_1,\\
&\;\;\max_{i,j,k}|\tensor{B}_2 (i+1,j,k)-\tensor{B}_2 (i,j,k)|\leq 2\min(2L_{1},L_{2}),\\
&\;\;\max_{i,j,k}|\tensor{B}_3 (i+1,j,k)-\tensor{B}_3 (i,j,k)|\leq 2L_1,\\
&\;\;\max_{i,j,k}|\tensor{B}_4 (i+1,j,k)-\tensor{B}_4 (i,j,k)|\leq 2\min(L_{1},L_{2}),\\
\end{aligned}
\end{equation}
\begin{equation*}\small
\begin{aligned}
&{\rm (Dimension\;}y{\rm)}\\
&\;\;\max_{i,j,k}|\tensor{B}_1 (i,j+1,k)-\tensor{B}_1 (i,j,k)|\leq 4L_2,\\
&\;\;\max_{i,j,k}|\tensor{B}_2 (i,j+1,k)-\tensor{B}_2 (i,j,k)|\leq 2L_2,\\
&\;\;\max_{i,j,k}|\tensor{B}_3 (i,j+1,k)-\tensor{B}_3 (i,j,k)|\leq 2\min(L_{1},2L_{2}),\\
&\;\;\max_{i,j,k}|\tensor{B}_4 (i,j+1,k)-\tensor{B}_4 (i,j,k)|\leq 2\min(L_{1},L_{2}),\\
&{\rm (Dimension\;}z{\rm)}\\
&\;\;\max_{i,j,k}|\tensor{B}_s (i,j,k+1)-\tensor{B}_s (i,j,k)|\leq 2L_3,\;s=1,2,3,4.\\
\end{aligned}
\end{equation*}
All the above upper bounds are optimal, e.g., for any $\epsilon>0$ there exists a tensor $\tensor{A}$ that admits the assumptions such that $\max_{i,j,k}|\tensor{B}_1 (i+1,j,k)-\tensor{B}_1 (i,j,k)|>4L_1-\epsilon$.
\end{theorem}
Theorem \ref{th: Smoothness for the Wavelet Coefficient} states that the spectral smoothness bound (dimension $z$) of different wavelet coefficients is the same, whereas the spatial smoothness bound (dimensions $x$ and $y$) varies across different wavelet coefficients. Hence, it is well-motivated to utilize the spatially decoupled and spectrally coupled CF-INR (see Sec. \ref{SCSD}) to characterize such smoothness features.
Furthermore, we give a more concise formulation of the spatial smoothness of wavelet coefficients by using the discrete Laplacian, which evaluates the spatial smoothness of a tensor by integrating its spatial local differences.
\begin{definition}[Discrete Laplacian of a tensor\cite{Laplacian}\label{de: Frontal Slice-Wise Discrete Laplacian}]
The spatial discrete Laplacian of a tensor $\tensor{B} \in \mathbb{R}^{n_1 \times n_2 \times n_3}$ is defined as $
     L_{\tensor{B}}(i,j,k)\triangleq |4\tensor{B}(i,j,k)-\tensor{B}(i+1,j,k)-\tensor{B}(i-1,j,k)-\tensor{B}(i,j+1,k)-\tensor{B}(i,j-1,k)|$.
\end{definition}
\begin{corollary}[Laplacian bounds for wavelet coefficients of a tensor\label{th: discrete Laplacian for Wavelet}]

Let the assumptions in Theorem \ref{th: Smoothness for the
Wavelet Coefficient} hold. Then we have the following Laplacian bounds for wavelet coefficients:
\begin{equation}\small\label{bound_1}
    \begin{aligned}
    &\max_{i,j,k}L_{\tensor{B}_{1}}(i,j,k)\leq 8(L_{1}+L_{2}),\\
    &\max_{i,j,k}L_{\tensor{B}_{2}}(i,j,k)\leq 4(L_{2}+ \min(2L_{1},L_{2})),\\
    &\max_{i,j,k}L_{\tensor{B}_{3}}(i,j,k)\leq 4(L_{1}+ \min(L_{1},2L_{2})),\\
    &\max_{i,j,k}L_{\tensor{B}_{4}}(i,j,k)\leq 8 \min(L_{1},L_{2}),\\
    \end{aligned}
\end{equation}
and the upper bounds are optimal.
\end{corollary}
Corollary \ref{th: discrete Laplacian for Wavelet} gives the spatial Laplacian bounds for different wavelet coefficients of a tensor. By comparing the upper bounds $
8\min(L_{1}+L_{2})\geq 4(L_{2}+\min(2L_{1},L_{2}))\&4(L_{1}+\min(L_{1},2L_{2}))\geq8(L_{1},L_{2})$, we can observe the heterogeneous smoothness bounds across different frequency components, consistent with empirical findings (see Fig. \ref{fig:Laplacian_for_Wavelet_Coefficients}, where the Laplacian histograms of different wavelet coefficients vary). Furthermore, different images $\cal A$ exhibit different smoothness bounds $L_1,L_2,L_2$, hence the wavelet coefficients of different images exhibit externally different Laplacian bounds, also consistent with Fig. \ref{fig:Laplacian_for_Wavelet_Coefficients}, where the Laplacian histograms of different images vary. 
The heterogeneity of Laplacian smoothness across frequency components and across images underscores the necessity of using separate smooth characterizations for different wavelet coefficients and datasets.

\subsubsection{Cross-Frequency Laplacian Bound of CF-INR}
Next, we deduce the Laplacian bound of wavelet coefficients represented by CF-INR, which establishes the relationships between the cross-frequency Laplacian bound and cross-frequency parameters $\omega_1$-$\omega_4$ of CF-INR. 
\begin{theorem}[Cross-frequency Laplacian bound of CF-INR\label{th: Laplacian Values of Wavelet Coefficients of CF-INR}] Consider the spatially decoupled and spectrally coupled CF-INR model \eqref{LRTFR_share}, i.e., ${\cal B}_{\theta_s} = {\cal C}\times_3{\boldsymbol{W}}_{\theta^z}\times_2{\boldsymbol{V}}_{\theta_s^y}\times_1{\boldsymbol{U}}_{\theta_s^x}\in \mathbb{R}^{\frac{n_1}{2} \times \frac{n_2}{2} \times n_3}$ $(s=1,2,3,4)$. Suppose the spatial INRs $f_{\theta_s^x}(\cdot),f_{\theta_s^y}(\cdot)$ have depth $d$ with the activation function $\sin(\omega_s\cdot)$ $(s=1,2,3,4)$, and the spectral INR $f_{\theta^z}(\cdot)$ has depth $d$ with the activation function $\sin(\omega^z\cdot)$. Then we have the following Laplacian bounds for wavelet coefficients $\{{\cal B}_{\theta_s}\}_{s=1}^4$ represented by such CF-INR:
    \begin{equation}\small\label{eq: 2-D Frontal Slice-wise discrete Laplacian}
\max_{i,j,k}L_{{\cal B}_{\theta_s}}(i,j,k)\leq C\omega_{s}^{2d-2}, s=1,2,3,4,
    \end{equation}
where $C=4\eta^{3d+1}(\omega^z)^{d-1}n_3\max\{\frac{n_1}{2},\frac{n_2}{2}\}$ is a constant independent of $s$. Here $\eta=\max\{\lVert{\cal C}\rVert_{\ell_1},\max_i\lVert{\boldsymbol{H}_i}\rVert_{\ell_1}\}$ denotes the maximum $\ell_1$-norm of network parameters. The upper bounds \eqref{eq: 2-D Frontal Slice-wise discrete Laplacian} are optimal, i.e., for any $\epsilon>0$ there exist a group of network parameters $\{\theta_s\}_{s=1}^4$ that admit the assumptions such that $\max_{i,j,k}L_{{\cal B}_{\theta_s}}(i,j,k)> C\omega_{s}^{2d-2}-\epsilon$.
\end{theorem}
Theorem \ref{th: Laplacian Values of Wavelet Coefficients of CF-INR} establishes the theoretical polynomial relationship between the Laplacian smoothness bound of wavelet coefficients and the spatial frequency parameters $\omega_1$-$\omega_4$ of CF-INR, which provides understanding on how the frequency parameters $\omega_1$-$\omega_4$ influence the smoothness of the generated wavelet coefficients. The smoothness of generated wavelet coefficients is implicitly encoded through the structure of CF-INR with cross-frequency parameters $\omega_1$-$\omega_4$. For each wavelet coefficient, we can readily control its spatial smoothness by adjusting $\{\omega_s\}_{s=1}^4$, i.e., the smaller $\omega_s$ is, the smoother wavelet coefficient ${\cal B}_{\theta_s}$ will be generated.
By comparing the Laplacian bounds in Corollary \ref{th: discrete Laplacian for Wavelet} and Theorem \ref{th: Laplacian Values of Wavelet Coefficients of CF-INR}, we can tune the parameters $\omega_1$-$\omega_4$ to enable theoretically accurate characterizations for the smoothness of generated wavelet coefficients. 
Specifically, it could be well-accepted to set $\omega_4\geq\omega_3\&\omega_2\geq\omega_1$ by mimicking the Laplacian bounds sequence derived in Corollary \ref{th: discrete Laplacian for Wavelet}, which encodes the desired smoothness characteristics. However, such manual selections would be inefficient in practice. Next, we introduce a self-evolving strategy, which automatically adjusts cross-frequency parameters $\omega_1$-$\omega_4$ toward the desired characterizations in a progressive manner.
\subsubsection{Self-Evolving Strategy for Cross-Frequency Parameters\label{sec_5}}
Finding an appropriate combination of cross-frequency parameters $\omega_1$-$\omega_4$ for each dataset is challenging. Here we design a strategy that automatically adjusts $\omega_1$-$\omega_4$ for each dataset based on the theoretical analysis in Sec. \ref{sec_5_Lap}. 
The main motivation is to use the intermediate outputs of CF-INR $\{{\cal B}_{\theta_s}\}_{s=1}^4$ during the iterations to guide the adjustments for $\{\omega_s\}_{s=1}^4$ progressively. We first calculate the average Laplacian values
\begin{equation}\small\label{mean}
\overline{L}_{{\cal B}_{\theta_s}}\triangleq \frac{1}{N}\sum_{i,j,k}L_{{\cal B}_{\theta_s}}(i,j,k),\;s=1,2,3,4,
\end{equation}
where $N={n_1n_2n_3}/{4}$ denotes the total number of elements. In Theorem \ref{th: Laplacian Values of Wavelet Coefficients of CF-INR} we deduce the following Laplacian bounds for wavelet coefficients represented by CF-INR:
    \begin{equation}\small
\max_{i,j,k}L_{{\cal B}_{\theta_s}}(i,j,k)\leq C\omega_{s}^{2d-2},\;s=1,2,3,4,
    \end{equation}
where $\omega_1$-$\omega_4$ are cross-frequency parameters and $d$ denotes the network depth. Therefore, we make the following key hypothesis: the desired Laplacian bounds in \eqref{eq: 2-D Frontal Slice-wise discrete Laplacian} should share consistent relative relationships with \eqref{mean}, i.e.,
    \begin{equation}\small\label{update}
C\omega_{s}^{2d-2}/C\omega_{s'}^{2d-2}\leftarrow\overline{L}_{{\cal B}_{\theta_{s}}^t}/\overline{L}_{{\cal B}_{\theta_{s'}}^t},\;s,s'=1,2,3,4.
    \end{equation}
This can be equivalently expressed as $\omega_1:\omega_2:\omega_3:\omega_4\leftarrow(\overline{L}_{{\cal B}_{\theta_1}^t})^{\frac{1}{2d-2}}:(\overline{L}_{{\cal B}_{\theta_2}^t})^{\frac{1}{2d-2}}:(\overline{L}_{{\cal B}_{\theta_3}^t})^{\frac{1}{2d-2}}:(\overline{L}_{{\cal B}_{\theta_4}^t})^{\frac{1}{2d-2}}$ since $C$ is a constant. The rationale of this hypothesis is that with the iteration proceeds, the learned wavelet coefficients $\{{\cal B}_{\theta_s}\}_{s=1}^4$ progressively converge to more accurate representations and thus the corresponding average Laplacian \eqref{mean} more accurately reveals the desired smoothness of wavelet coefficients. Therefore, dynamically updating $\omega_1$-$\omega_4$ using \eqref{update} intends to shift the frequency preference towards more precise configurations. Especially, we automatically update $\omega_1$-$\omega_4$ based on the hypothesis \eqref{update} and the constraint $\sum_{s=1}^4\omega_s=\mu$, where $\mu$ is a single summing hyperparameter (i.e., we reduce the four hyperparameters $\omega_1$-$\omega_4$ into a single hyperparameter $\mu$ indicating their sum, and our method is robust to $\mu$; see Fig. \ref{fig: hyperparameter}(c)). This leads to the following update rule:
\begin{equation}\small\label{auto_fre}
\omega_s\leftarrow\frac{\mu\left(\overline{L}_{{\cal B}_{\theta_s}}\right)^{\frac{1}{2d-2}}}{\sum_{s'=1}^4\left(\overline{L}_{{\cal B}_{\theta_{s'}}^t}\right)^{\frac{1}{2d-2}}},\;s=1,2,3,4.
\end{equation}\par
To ensure smooth optimization, the cross-frequency parameters $\omega_1$–$\omega_4$ are updated every 500 iterations according to \eqref{auto_fre}. As the iterations proceed, the algorithm gradually identifies a more suitable combination of $\omega_1$–$\omega_4$ for each observed data. The self-evolution cross-frequency efficiently characterizes both the structure variations within a signal by learning distinct frequencies for different wavelet coefficients, and the frequency multiplicity across signals by learning customized frequency combinations for different images (see Table \ref{table: hyperparameters combinations for different images} for example). In Sec. \ref{sec: Influences of Adaptive Hyperparameter}, we conduct a comprehensive experimental analysis for the self-evolving strategies.
\subsection{CF-INR for Data Representation and Recovery}\label{sec_6}
In this section, we establish the optimization models of CF-INR for various data representation and recovery tasks including image regression, inpainting, denoising, and cloud removal to showcase the versatility of the proposed CF-INR for various vision tasks.
\subsubsection{Image Regression}
First, we introduce how to use CF-INR to directly learn a continuous representation of a given image. Given an observed data $\tensor{A}\in\mathbb{R}^{n_{1}\times n_{2}\times n_{3}}$, the image regression model using CF-INR is formulated as
\begin{equation}\small\label{model_fit}
\begin{aligned}
&\min_{\Theta}\left\lVert\mathrm{IHWT}([{\tensor{B}}_{\theta_1}, {\tensor{B}}_{\theta_2}; {\tensor{B}}_{\theta_3},{\tensor{B}}_{\theta_4}])-{\cal A}\right\rVert_F^2,\\
&\;{\cal B}_{\theta_s} = {{\cal C}\times_3{\boldsymbol{W}}_{\theta^z}}\times_2{{\boldsymbol{V}}_{\theta_s^y}\times_1{\boldsymbol{U}}_{\theta_s^x}},\;s=1,2,3,4,\\
&\;{\boldsymbol{U}}_{\theta_s^x}(i,:)=f_{\theta_s^x}(i),i=1,2,\cdots,\frac{n_2}{2},\\
&\;{\boldsymbol{V}}_{\theta_s^y}(j,:)=f_{\theta_s^y}(j),j=1,2,\cdots,\frac{n_2}{2},\\
&\;{\boldsymbol{W}}_{\theta^z}(k,:)=f_{\theta^z}(k),k=1,2,\cdots,n_3,
\end{aligned}
\end{equation}
where $\Theta=\{{\cal C},\{\theta_s^x\}_{s=1}^4,\{\theta_s^y\}_{s=1}^4,\theta^z\}$ are learnable parameters, $\{{\cal B}_{\theta_s}\}_{s=1}^4$ are the learned wavelet coefficients, $\cal C$ is the core tensor, and $f_{\theta_s^x}(\cdot),f_{\theta_s^y}(\cdot),f_{\theta^z}(\cdot)$ are univariate INRs parameterized by SIRENs \cite{activation_function1}. Since the IHWT and the Tucker decomposition are both differentiable, the overall objective function \eqref{model_fit} is differentiable w.r.t. $\Theta$. Here we employ the Adam optimizer \cite{Adam} to optimize the model parameters according to \eqref{model_fit}. The self-evolving strategies \eqref{auto_rank} and \eqref{auto_fre} are performed every 500 iterations during the optimization to automatically identify suitable cross-frequency configurations for the observed data $\cal A$.

\subsubsection{Image Inpainting}
Subsequently, we explore the applications of CF-INR for data recovery tasks. Since the CF-INR model intrinsically encodes the low-rankness through the tensor decomposition and the smoothness priors through the functional representations of INRs, it is effective for various data recovery tasks. Here we consider the image inpainting task, which focuses on restoring missing regions in an image by leveraging its spatial and spectral correlations. Given an observed data $\tensor{A}\in\mathbb{R}^{n_{1}\times n_{2}\times n_{3}}$, the optimization model for CF-INR for image inpainting is formulated as
\begin{equation}\small\label{model_inpainting}
\begin{aligned}
&\min_{\Theta}\left\lVert{\cal P}_\Omega \left(\mathrm{IHWT}([{\tensor{B}}_{\theta_1}, {\tensor{B}}_{\theta_2}; {\tensor{B}}_{\theta_3},{\tensor{B}}_{\theta_4}])-{\cal A}\right)\right\rVert_F^2,\\
&\;{\cal B}_{\theta_s} = {{\cal C}\times_3{\boldsymbol{W}}_{\theta^z}}\times_2{{\boldsymbol{V}}_{\theta_s^y}\times_1{\boldsymbol{U}}_{\theta_s^x}},\;s=1,2,3,4,\\
\end{aligned}
\end{equation}
where the factor matrices $\boldsymbol{W}_{\theta^z}, {{\boldsymbol{V}}_{\theta_s^y},{\boldsymbol{U}}_{\theta_s^x}}$ and the learnable parameters $\Theta$ follow from \eqref{model_fit} and $\tensor{P}_\Omega(\cdot)$ denotes the projection operator that retains the elements in $\Omega$ while setting others to zeros. The recovered result is obtained by $\tensor{P}_\Omega(\tensor{A})+\tensor{P}_{\Omega^C}(\mathrm{IHWT}([{\tensor{B}}_{\theta_1}, {\tensor{B}}_{\theta_2}; {\tensor{B}}_{\theta_3},{\tensor{B}}_{\theta_4}]))$, where $\Omega^C$ denotes the complementary set of $\Omega$. Similarly, we tackle the model \eqref{model_inpainting} by using the Adam optimizer along with the self-evolving strategies \eqref{auto_rank} and \eqref{auto_fre}. Cloud removal can be viewed as a special case of image inpainting, and thus shares the same model as in \eqref{model_inpainting}.
\begin{algorithm}[t]
	\begin{spacing}{0.9}
		\renewcommand\arraystretch{1.2}
		\caption[Caption for LOF]{{The proposed self-evolving CF-INR for image denoising using model \eqref{model_denoise}}}\label{alg}
		\begin{algorithmic}[1]
			\renewcommand{\algorithmicrequire}{\textbf{Input:}}
			\Require
			Observed data $\cal A$, parameters $\gamma_1,\gamma_2,\omega^z,\mu,\lambda_x,\lambda_y,r_z$;
			\renewcommand{\algorithmicrequire}{\textbf{Initialization:}}
			\Require Randomly initialize the core tensors $\cal C$ and INR weights $\{\theta_s^x,\theta_s^y\}_{s=1}^4,\theta^z$, initialize ${\cal S},\Lambda=0$,$t=0$;
			\While{not converged}
			\State Update $\cal X$ via \eqref{X} using first-order optimality;
			\State Update CF-INR weights $\Theta$ via \eqref{Theta} using Adam;
			\State Update ${\cal S}$ via \eqref{S} using soft thresholding;
			\State Update multiplier $\Lambda$ and $\rho$ via \eqref{B_rho}; $t=t+1$;
			\If{${\rm mod}(t,500)=0$}
			\State Update cross-frequency ranks $R_1$-$R_4$ via \eqref{auto_rank};
			\State Update cross-frequency parameters $\omega_1$-$\omega_4$ via \eqref{auto_fre};
			\EndIf
			\EndWhile
			\renewcommand{\algorithmicrequire}{\textbf{Output:}}
			\Require The recovered image ${\cal A}_\Theta$;
		\end{algorithmic}
	\end{spacing}
\end{algorithm}
\subsubsection{Mixed Noise Removal}
We consider the image denoising task with mixed noise, which aims to simultaneously attenuate random and sparse noise from a given noisy image $\tensor{A}\in\mathbb{R}^{n_{1}\times n_{2}\times n_{3}}$. The optimization model of CF-INR for image denoising with mixed noise is formulated as follows:
\begin{equation}\small\label{model_denoise}
\begin{aligned}
&\min _{\Theta,{\cal S}}\|\tensor{A}-\tensor{{A}}_{\Theta}-\tensor{S}\|_F^2+\gamma_1\|\tensor{S}\|_{\ell_1}+\gamma_2\|\tensor{{A}}_{\Theta}\|_{\mathrm{TV}},\\
&\;\tensor{{A}}_{\Theta} = \mathrm{IHWT}([{\tensor{B}}_{\theta_1}, {\tensor{B}}_{\theta_2}; {\tensor{B}}_{\theta_3},{\tensor{B}}_{\theta_4}]),\\
&\;{\cal B}_{\theta_s} = {{\cal C}\times_3{\boldsymbol{W}}_{\theta^z}}\times_2{{\boldsymbol{V}}_{\theta_s^y}\times_1{\boldsymbol{U}}_{\theta_s^x}},\;s=1,2,3,4,\\
\end{aligned}
\end{equation}\begin{figure*}[t]
	\centering
	\setlength{\tabcolsep}{0pt}
	\begin{tabular}{ccccc}
		\vspace{-0.25cm}
		\includegraphics[width=0.19\textwidth]{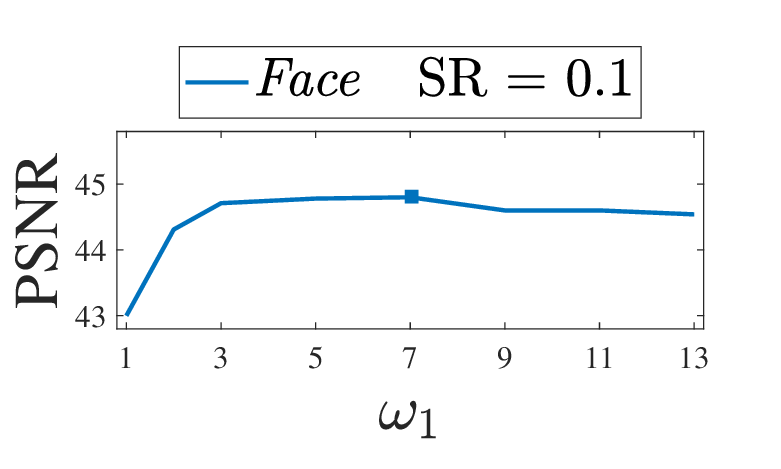}&
		\includegraphics[width=0.19\textwidth]{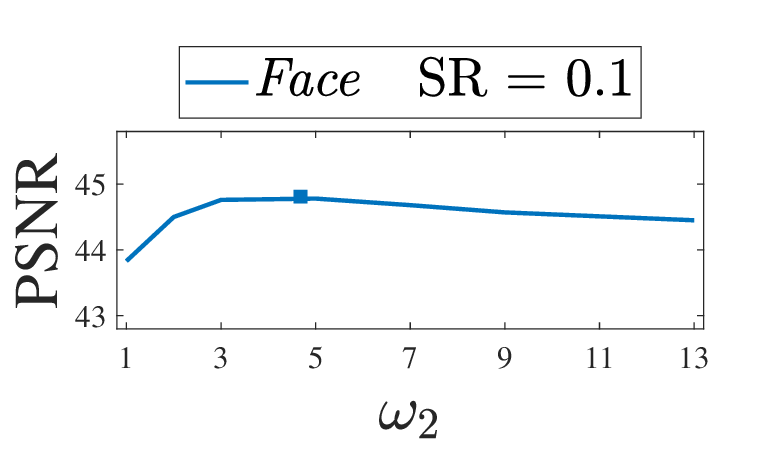}&
		\includegraphics[width=0.19\textwidth]{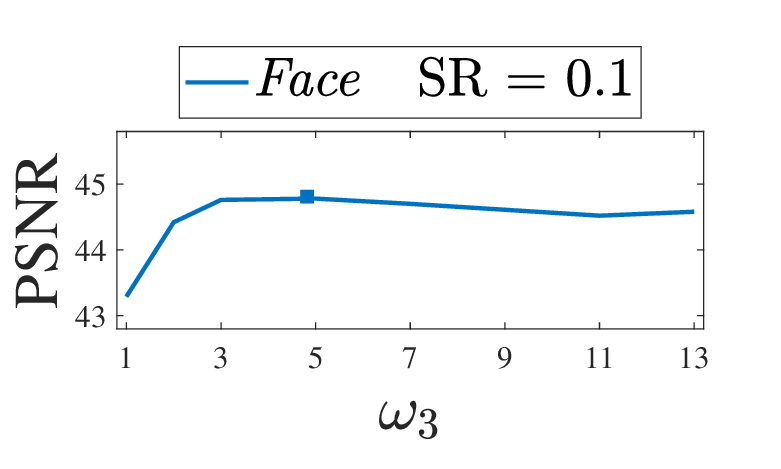}&
		\includegraphics[width=0.19\textwidth]{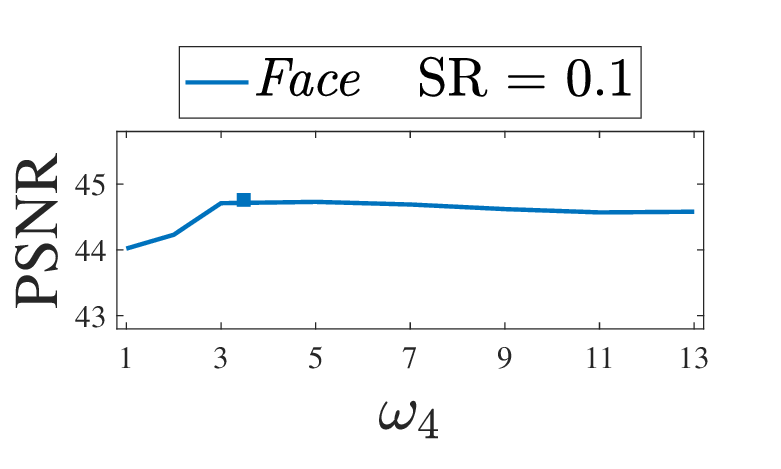}&
				\includegraphics[width=0.19\textwidth]{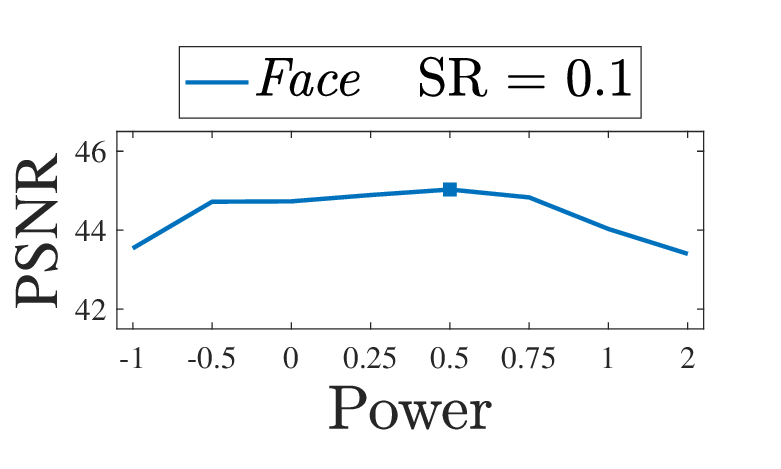}\\
		\vspace{-0.1cm}
		\includegraphics[width=0.19\textwidth]{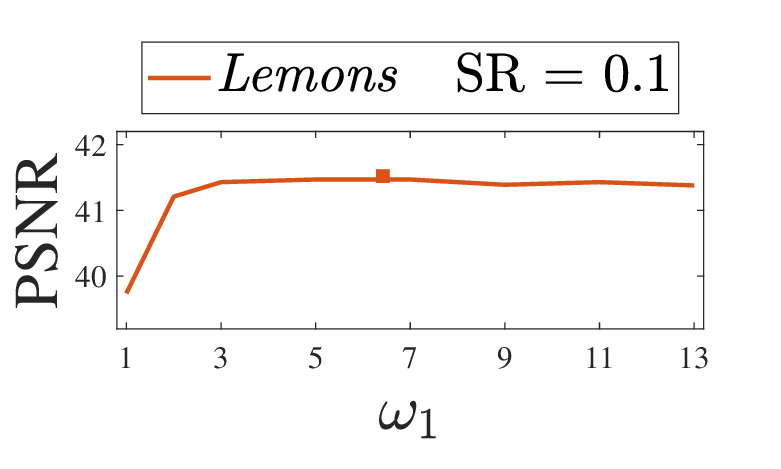}&
		\includegraphics[width=0.19\textwidth]{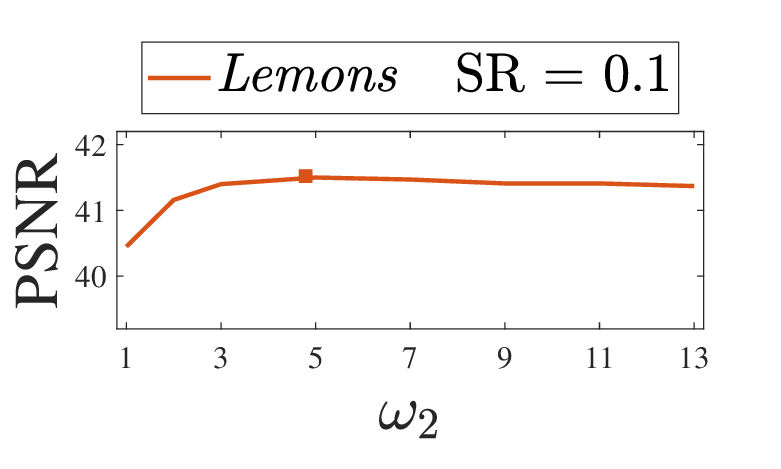}&
		\includegraphics[width=0.19\textwidth]{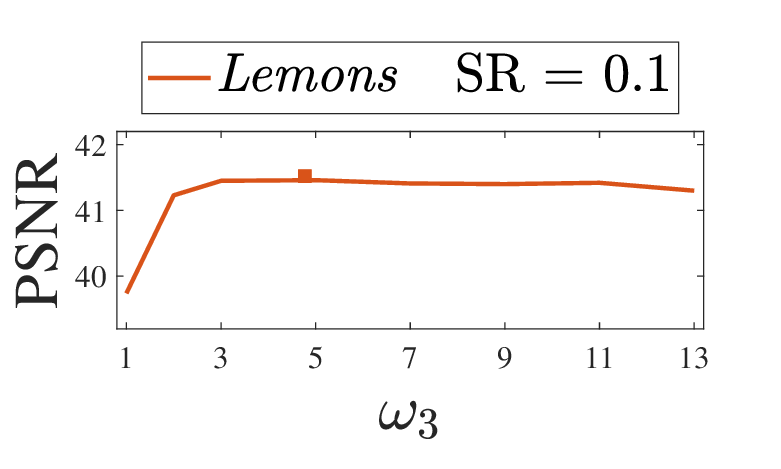}&
		\includegraphics[width=0.19\textwidth]{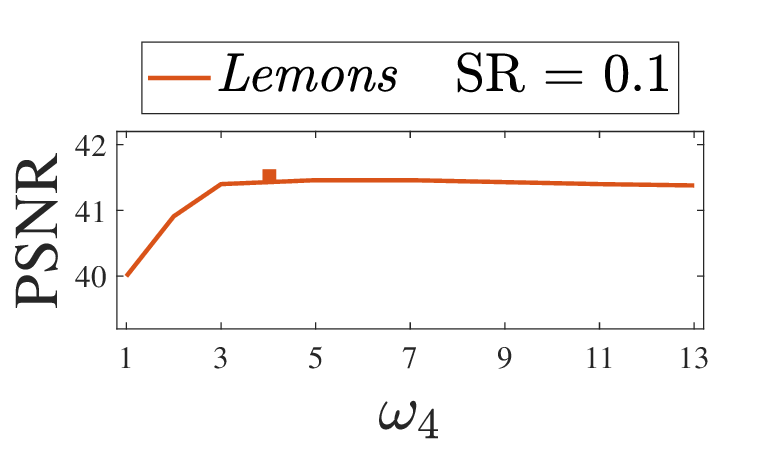}&
				\includegraphics[width=0.19\textwidth]{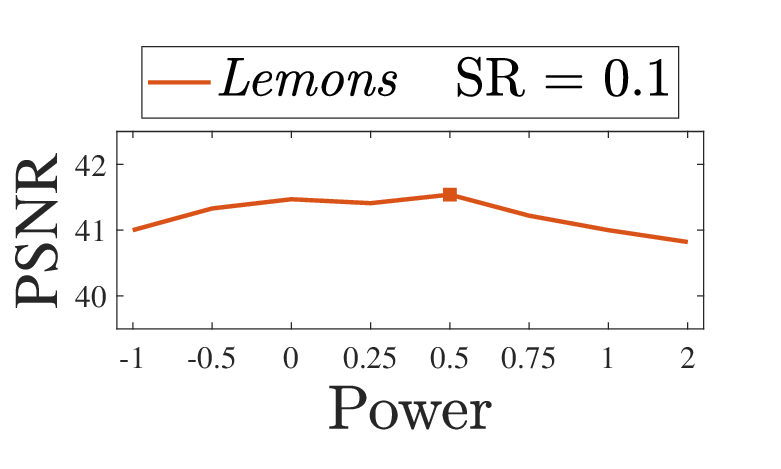}\\\vspace{-0.1cm}
		(a)&(b)&(c)&(d)&(e)\\
	\end{tabular}
	\caption{(a)-(d): Sensitivity tests for cross-frequency parameters $\omega_1$-$\omega_4$, with marked points indicating the ones automatically selected by the self-evolving strategy \eqref{auto_fre}. (e): Sensitivity tests for the power exponent parameter $\frac{1}{2d-2}$ in \eqref{auto_fre}, with marked points representing the default setting $\frac{1}{2}$ derived from theories.
}
	\vspace{-0.3cm}
	\label{fig: frequency parameters}
\end{figure*}\begin{figure*}[t]
	\centering
	\setlength{\tabcolsep}{0pt}
	\begin{tabular}{ccccc}
		\vspace{-0.25cm}
		\includegraphics[width=0.19\textwidth]{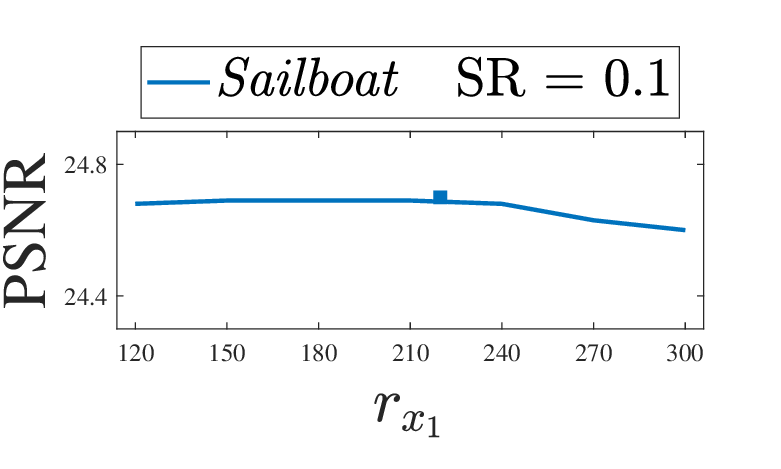}&
		\includegraphics[width=0.19\textwidth]{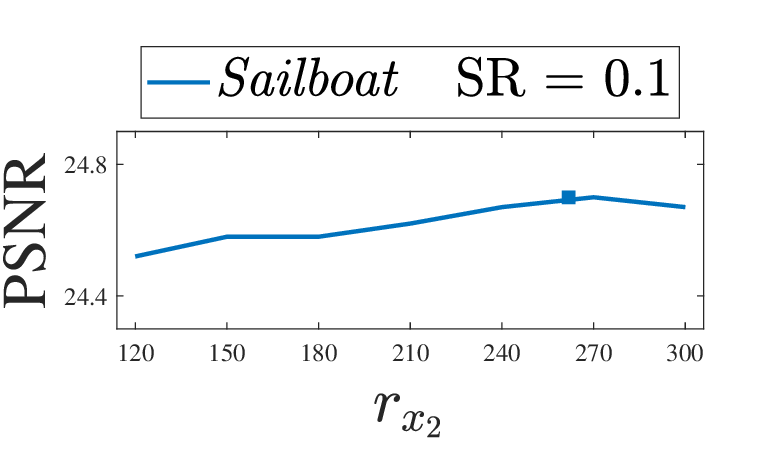}&
		\includegraphics[width=0.19\textwidth]{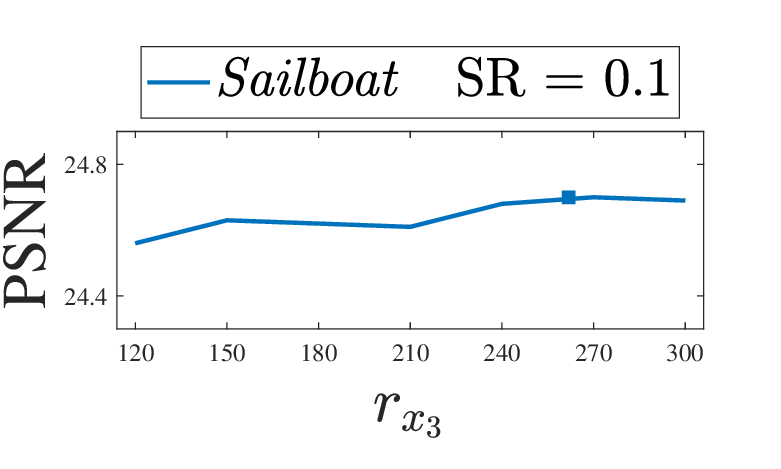}&
		\includegraphics[width=0.19\textwidth]{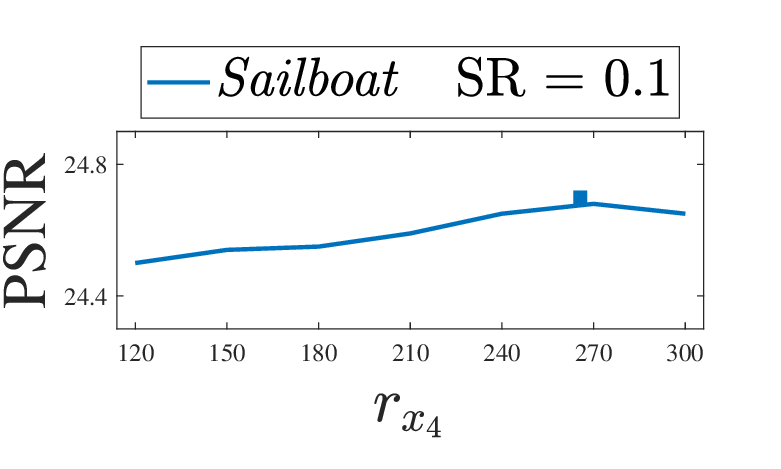}&
		\includegraphics[width=0.19\textwidth]{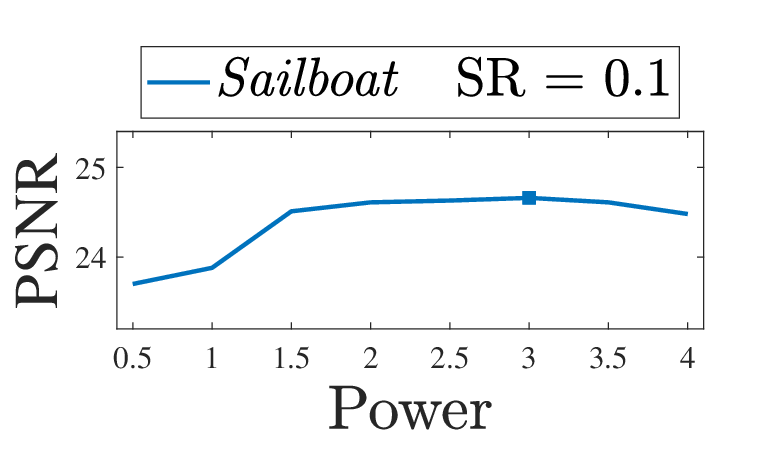}\\
		\vspace{-0.1cm}
		\includegraphics[width=0.19\textwidth]{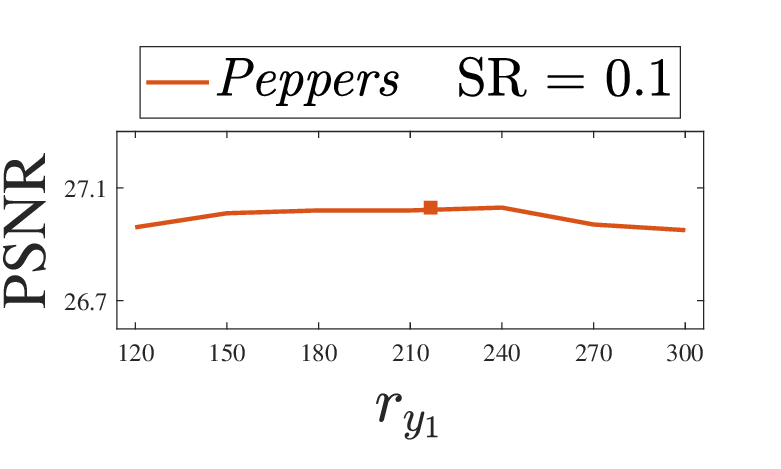}&
		\includegraphics[width=0.19\textwidth]{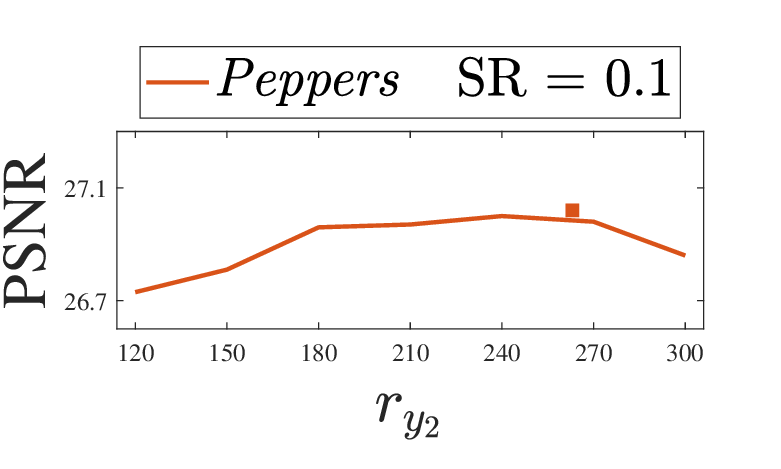}&
		\includegraphics[width=0.19\textwidth]{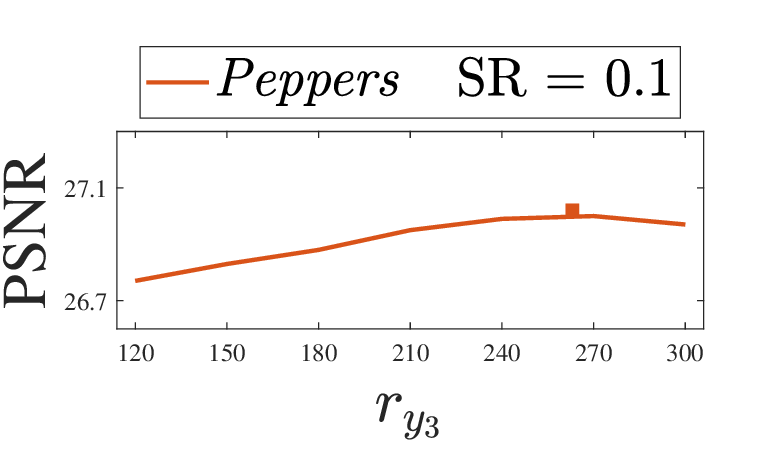}&
		\includegraphics[width=0.19\textwidth]{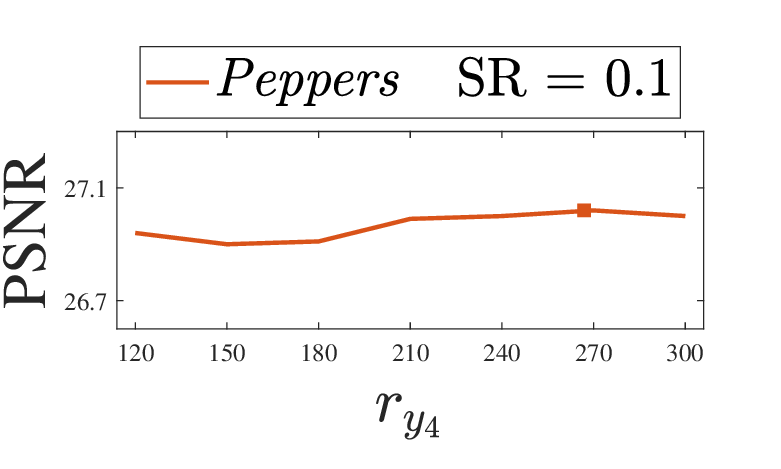}&
				\includegraphics[width=0.19\textwidth]{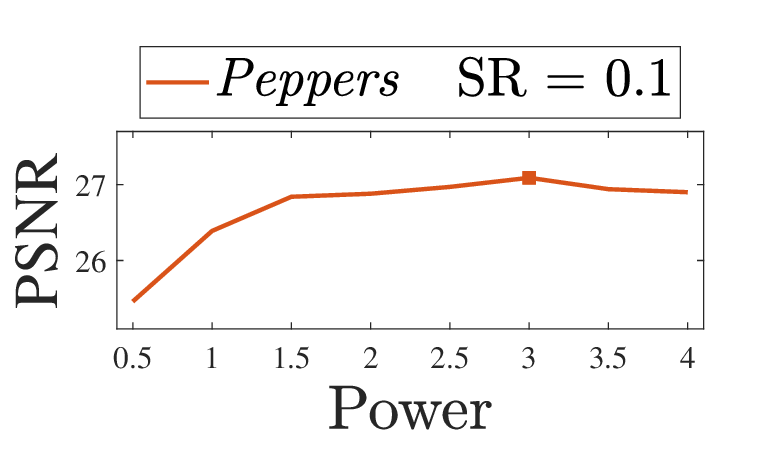}\\\vspace{-0.1cm}
		(a)&(b)&(c)&(d)&(e)\\
	\end{tabular}
	\caption{(a)-(d): Sensitivity tests for cross-frequency ranks $(r_{x_s},r_{y_s})$ $(s=1,2,3,4)$, with marked points indicating the ones automatically selected by the self-evolving strategy \eqref{auto_rank}. (e): Sensitivity tests for the power exponent parameter $k$ in \eqref{auto_rank}, with marked points representing the default setting $k=3$.
	}
	\vspace{-0.3cm}
	\label{fig: rank parameters}
\end{figure*}where the factor matrices $\boldsymbol{W}_{\theta^z}, {{\boldsymbol{V}}_{\theta_s^y},{\boldsymbol{U}}_{\theta_s^x}}$ and the learnable parameters $\Theta$ follow from \eqref{model_fit} and ${\cal S}\in\mathbb{R}^{n_{1}\times n_{2}\times n_{3}}$ denotes the sparse noise to be estimated. The $\gamma_1,\gamma_2$ are trade-off parameters. Here we employ a simple total variation regularization $\|\tensor{{A}}_{\Theta}\|_{\mathrm{TV}}\triangleq\| {\rm D}_x{\cal A}_{\Theta}\|_{\ell_1}+\|{\rm D}_y{\cal A}_{\Theta}\|_{\ell_1}$ to enhance the denoising capability, where ${\rm D}_x,{\rm D}_y$ are first-order discrete differences. To tackle the optimization in \eqref{model_denoise}, we consider the following plug-and-play ADMM \cite{TIP_PnP} paradigm by viewing the CF-INR model ${\cal A}_\Theta$ as a plug-and-play denoiser:
\begin{align}
&\min_{\cal X}\|{\cal A}-{\cal X}-{\cal S}^{t}\|_F^2+\frac{\rho^t}{2}\|{\cal X}-{\cal A}_{\Theta^{t}}+{\Lambda}^{t}\|_F^2,\label{X}\\
&\min_{\Theta}\frac{\rho^t}{2}\|{\cal X}^{t+1}-{\cal A}_\Theta+\Lambda^{t}\|_F^2+\gamma_2\|{\cal A}_\Theta\|_{\rm TV},\label{Theta}\\
&\min_{\cal S}\|{\cal A}-{\cal X}^{t+1}-{\cal S}\|_F^2+\gamma_1\|{\cal S}\|_{\ell_1},\label{S}\\
&\;{\Lambda}^{t+1}={\Lambda}^t+{\cal X}^{t+1}-{\cal A}_{\Theta^{t+1}},\;\rho^{t+1}=\kappa\rho^{t},\label{B_rho}
\end{align}
where $\cal X$ is an auxiliary variable, $\Lambda$ is the Lagrange multiplier, and $\kappa>1$ is a constant. The $\cal X$ and $\cal S$ subproblems can be solved by first-order optimality and soft thresholding. The $\Theta$ subproblem is optimized by using the Adam optimizer in each iteration of the ADMM. The overall algorithm of such a CF-INR model for image denoising is illustrated in Algorithm \ref{alg}. Under mild assumptions, we have the following fixed-point convergence guarantee for the ADMM.
\begin{lemma}
Assume that the CF-INR model is a bounded mapping, i.e., $\|{\cal A}_{\Theta^{t+1}}-({\cal X}^{t+1}+{\Lambda}^t)\|_{F}^2\leq{\frac{C}{\rho^t}}$ for a universal constant $C$. Then the iterations of the ADMM in \eqref{X}-\eqref{B_rho} admit a fixed-point convergence, i.e., there exists $({\cal X}^*,\Theta^*,{\Lambda}^*)$ such that $\|{\cal X}^t-{\cal X}^*\|_F^2\rightarrow 0$, $\|{\cal A}_{\Theta^t}-{\cal A}_{\Theta^*}\|_F^2\rightarrow 0$, and $\|{\Lambda}^t-{\Lambda}^*\|_F^2\rightarrow 0$ as $t\rightarrow \infty$.
\end{lemma}
\section{Numerical Experiments\label{sec: Experiments}}
We conduct comprehensive experiments to validate the proposed CF-INR method. Our method is implemented using PyTorch 3.12.3 on a system with an i7-12700KF CPU and an RTX 3060 GPU (12 GB of GPU memory). The results are quantitatively evaluated by peak signal-to-noise ratio (PSNR), structural similarity (SSIM), and normalized root mean square error (NRMSE). Better image quality is indicated by higher PSNR and SSIM scores and lower NRMSE. We also integrate our proposed method with INGP \cite{INGP} for NeRF representation. Please refer to supplementary file.\par We present the hyperparameter settings below.  
The sum of spatial ranks are selected in $\lambda_x\in\{2n_1,3n_1\}$ and $\lambda_y\in\{2n_2,3n_2\}$. The spectral rank $r_z$ is set as $r_z=10$ for color image inpainting, $r_z=n_3$ for inpainting of other datasets, $r_z=n_3$ for cloud removal, and $r_z=\frac{n_3}{2}$ for image denoising. The sum of spatial frequency parameters $\mu$ is selected in $\mu\in\{15,20,25,30\}$ (we show in Fig. \ref{fig: hyperparameter} that our method is quite robust to this parameter), and the shared spectral frequency parameter $\omega^z$ is selected in $\{1,2,3,4\}$. 
\begin{table}[t]
	\scriptsize
	\caption{The self-evolving strategies automatically select distinct cross-frequency parameters $\omega_1$-$\omega_4$ and cross-frequency ranks $(r_{x_s},r_{y_s})$ ($s=1,2,3,4$) for different images.}
	\label{table: hyperparameters combinations for different images}
	\centering
	{
		\setlength{\tabcolsep}{2.7pt}
		\begin{tabular}{ccccccccc}
			\hline
			Images & $\omega_1$   & $\omega_2$   & $\omega_3$   & $\omega_4$ & ($r_{x_1}$,$r_{y_1}$)   & ($r_{x_2}$,$r_{y_2}$)   & ($r_{x_3}$,$r_{y_3}$)   & ($r_{x_4}$,$r_{y_4}$)   \\ \hline
			\textit{Face}  & 7.03 & 4.68 & 4.82 & 3.48 & (101,100) & (131,131) & (133,134) & (147,147) \\
			\textit{Lemons}  & 6.42 & 4.79 & 4.77 & 4.02 & (94,94) & (132,133) & (134,133) & (152,152)\\
			\textit{Cloth} & 5.78 & 4.98 & 4.84 & 4.41 & (100,100) & (136,136) & (136,136) & (140,140)\\ \hline
	\end{tabular}}\vspace{-0.3cm}
\end{table}
\begin{table*}[tbp]
\scriptsize
\caption{
Integrating the proposed CF-INR into different INR network structures \cite{Fourier_feature,Wire,FINER,activation_function1} yields consistently improved performances for image regression, demonstrating the compatibility and effectiveness of CF-INR as a flexible plug-and-play module for INR methods. }
\label{tab:fitting}
\centering
\setlength{\tabcolsep}{2.7pt}
\renewcommand{\arraystretch}{0.95}
\begin{tabular}{lcccccccccccccccccc}
\hline
Dataset & \multicolumn{3}{c}{Color image \textit{House}}                     & \multicolumn{3}{c}{Color image \textit{Airplane}}                        & \multicolumn{3}{c}{Color image \textit{Mandrill}}                     & \multicolumn{3}{c}{MSI \textit{Flowers}}                      & \multicolumn{3}{c}{MSI \textit{Toys}}                         & \multicolumn{3}{c}{MSI \textit{Painting}}                     \\
                        Method& PSNR           & SSIM           & NRMSE          & PSNR           & SSIM           & NRMSE          & PSNR           & SSIM           & NRMSE          & PSNR           & SSIM           & NRMSE          & PSNR           & SSIM           & NRMSE          & PSNR           & SSIM           & NRMSE          \\ \hline
PE-INR\cite{Fourier_feature}                      & 32.14          & 0.897          & 0.038          & 36.38          & 0.913          & 0.021          
& 25.05          & 0.771          & 0.104          & 42.88          & 0.957          & 0.050          & 42.36          & 0.977          & 0.027          & 38.47          & 0.944          & 0.063          \\
CF-PE-INR             & \bf41.40          & \bf{0.985}    & \bf0.013          & \bf42.87          & \bf{0.987}    & \bf0.010          
& \bf36.84          & \bf0.981          & \bf0.027          & \bf53.81          & \textbf{0.997} & \bf0.014          & \bf50.19          & \textbf{0.998} & \bf0.011          & \bf50.66          & \bf0.997          & \bf0.016          \vspace{0.1cm}\\

WIRE-INR\cite{Wire}                 & 36.95          & 0.952          & 0.022          & 41.33          & 0.960          & 0.012          
& 29.64          & 0.907          & 0.061          & 44.07          & 0.974          & 0.030          & 42.78          & 0.969          & 0.026          & 42.94          & 0.975          & 0.038          \\
CF-WIRE-INR           & \bf42.69          & \bf0.982          & \bf{0.011}    & \bf45.71          & \bf0.985          & \bf{0.007}    
& \bf38.72          & {0.983}    & \bf0.021          & \bf53.59          & \bf{0.995}    & \bf0.015          & \bf51.46          & \bf{0.997}    & \bf{0.009}    & \bf50.47          & \bf0.996          & \bf0.016          \vspace{0.1cm}\\

FINER-INR\cite{FINER}                   & 39.18          & 0.968          & 0.017          & 42.47          & 0.969          & 0.010          
& 30.56          & 0.916          & 0.055          & 51.41          & 0.992          & 0.019          & 46.35          & 0.986          & 0.018          & 46.46          & 0.988          & 0.026          \\
CF-FINER-INR          & \bf{51.55}    & \textbf{0.999} & \textbf{0.004} & \textbf{53.43} & \textbf{0.998} & \textbf{0.003} 
& \textbf{48.75} & \textbf{0.999} & \textbf{0.007} & \bf{56.90}    & \bf{0.995}    & \bf{0.010}    & \textbf{54.37} & \bf{0.997}    & \textbf{0.007} & \textbf{54.88} & \textbf{0.999} & \textbf{0.009} \vspace{0.1cm}\\

SIREN-INR\cite{activation_function1}                   & 37.91          & 0.965          & 0.020          & 41.40          & 0.962          & 0.012          
& 29.56          & 0.903          & 0.062          & 47.47          & 0.983          & 0.031          & 48.29          & 0.993          & 0.014          & 45.86          & 0.987          & 0.027          \\
CF-SIREN-INR          & \textbf{51.64} & \textbf{0.999} & \textbf{0.004} & \textbf{52.80}    & \textbf{0.998} & \textbf{0.003} & \textbf{47.84}    & \textbf{0.999} & \textbf{0.008}    & \textbf{57.29} & \textbf{0.997} & \textbf{0.009} & \textbf{53.93}    & \textbf{0.997}    & \textbf{0.007} & \textbf{54.69}    & \textbf{0.998}    & \textbf{0.010}    \\ \hline
\end{tabular}\vspace{-0.3cm}
\end{table*}
\subsection{Validation for the Self-Evolving Strategies}\label{sec: Influences of Adaptive Hyperparameter}
To highlight the capability of the self-evolving strategies for multi-frequency feature characterization, we perform a series of experiments. First, we evaluate the effectiveness of cross-frequency parameters self-evolution \eqref{auto_fre}. In our settings, we employ the activation function $\sin(\omega_s \cdot)$ for the two spatial dimensions of the $s$-th wavelet coefficient, using an MLP depth of $d=2$. Therefore, the cross-frequency parameters $\omega_1$-$\omega_4$ are updated by $\omega_1:\omega_2:\omega_3:\omega_4\leftarrow(\overline{L}_{{\cal B}_{\theta_1}})^{\frac{1}{2}}:(\overline{L}_{{\cal B}_{\theta_2}})^{\frac{1}{2}}:(\overline{L}_{{\cal B}_{\theta_3}})^{\frac{1}{2}}:(\overline{L}_{{\cal B}_{\theta_4}})^{\frac{1}{2}}$. In Fig. \ref{fig: frequency parameters} (e), we compare the performance of the theoretical power exponent parameter $\frac{1}{2d-2}=\frac{1}{2}$ against other exponents when updating the frequency parameters, and the theoretical power exponent achieves the best result. Then, in Fig. \ref{fig: frequency parameters} (a)-(d), we evaluate each cross-frequency parameter by changing it and fixing the other parameters, with the marked points indicating the parameter values learned by the proposed self-evolving strategy \eqref{auto_fre}. It can be observed that the frequency parameters selected by the self-evolving strategy always fall within the optimal range, reflecting the effectiveness and robustness of the self-evolving strategy for cross-frequency parameter selections.\par
Second, we evaluate the effectiveness of the cross-frequency rank self-evolving strategies. We compare the performance of different power exponent parameters $k$ in \eqref{auto_rank} (Fig. \ref{fig: rank parameters} (e)) and the performance of different ranks $(r_{x_s},r_{y_s})$ $(s=1,2,3,4)$ (Fig. \ref{fig: rank parameters} (a)-(d), with marked points indicating the ranks learned by the self-evolving strategy \eqref{auto_rank}). It can be observed that the cross-frequency rank selected by the self-evolving strategy always fall within the optimal range, reflecting the effectiveness of the self-evolving strategy for cross-frequency rank selections, thereby confirming the superiority of our method for efficient and automatic cross-frequency feature characterizations.\par
\begin{figure}[t]
	\setlength{\tabcolsep}{0.9pt}
	\begin{tabular}{cc}
		\includegraphics[width=0.242\textwidth]{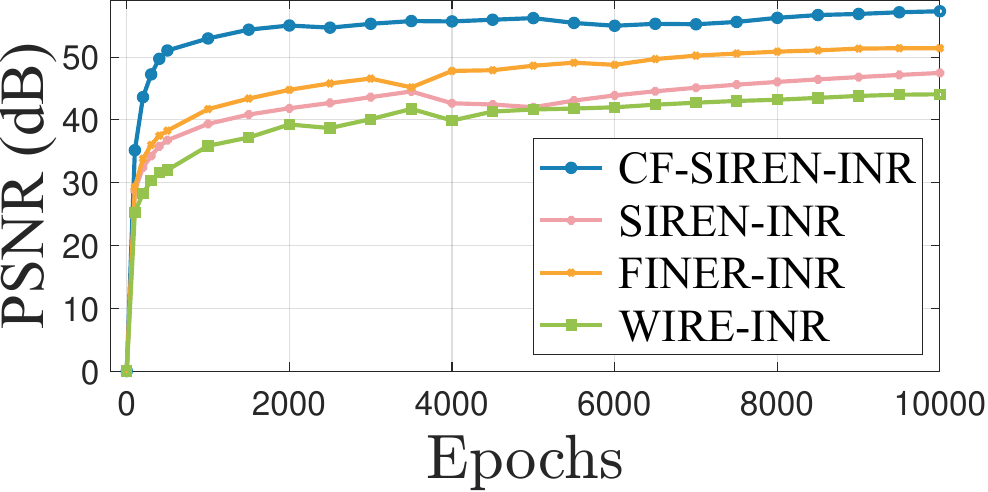}&
		\includegraphics[width=0.242\textwidth]{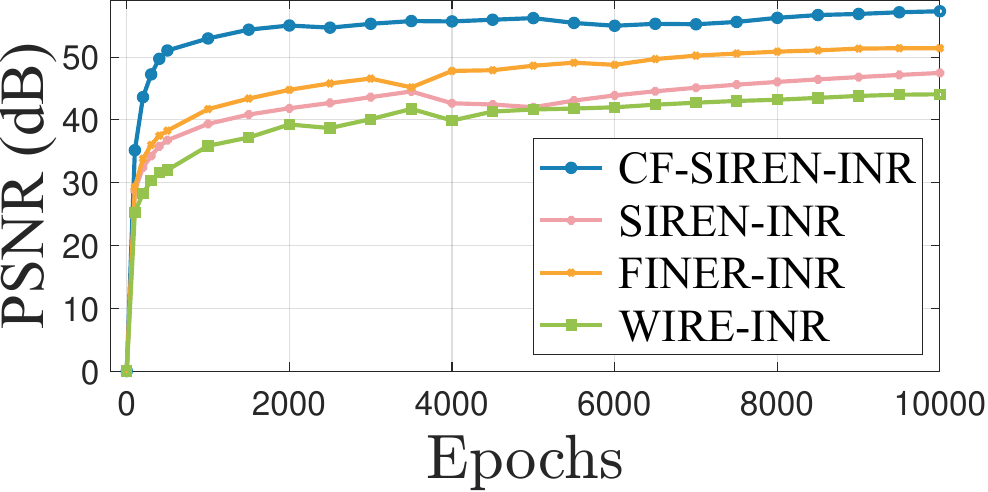}\\
		(a) Color image &(b) Multispectral image\\
	\end{tabular}
	\caption{Comparisons of PSNR curves obtained by different INR methods (WIRE \cite{Wire}, SIREN \cite{activation_function1}, FINER\cite{FINER}, and the proposed CF-SIREN-INR) for fitting the color image \textit{Airplane} and MSI \textit{Flowers}.
	}
	\label{fig: nihe_curves}\vspace{-0.5cm}
\end{figure}

Finally, Table \ref{table: hyperparameters combinations for different images} presents some detailed examples of cross-frequency parameters and ranks selected by the proposed self-evolving strategies for different images. The self-evolution automatically assigns heterogeneous cross-frequency configurations for different images, thereby enhancing the effectiveness of CF-INR for flexible multi-frequency feature characterizations based on the frequency decoupling. 

\subsection{Image Regression Results}\label{sec: fitting}
\begin{figure}[t]
	\setlength{\tabcolsep}{0.3pt}
	\scriptsize
	\centering
	\begin{tabular}{ccccc}
		\includegraphics[width=0.095\textwidth]
		{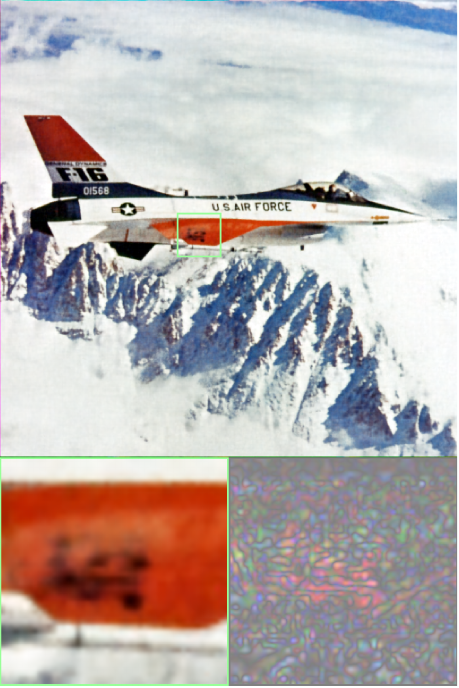}&
		\includegraphics[width=0.095\textwidth]{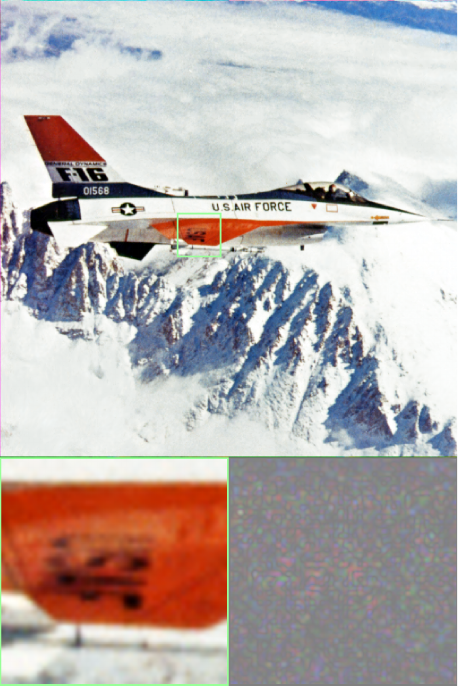}&
		\includegraphics[width=0.095\textwidth]{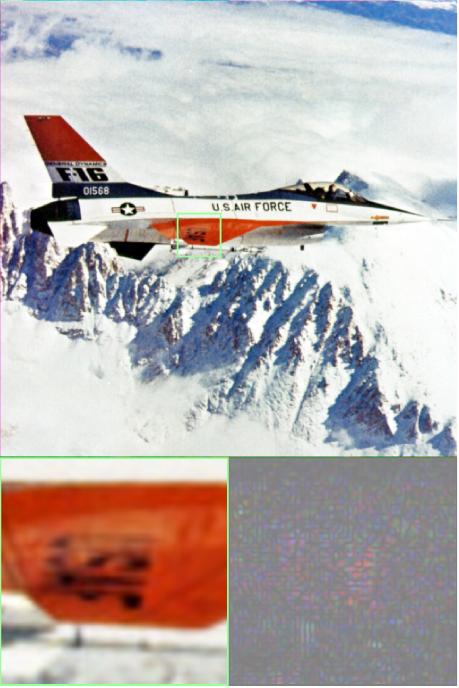}&
		\includegraphics[width=0.095\textwidth]{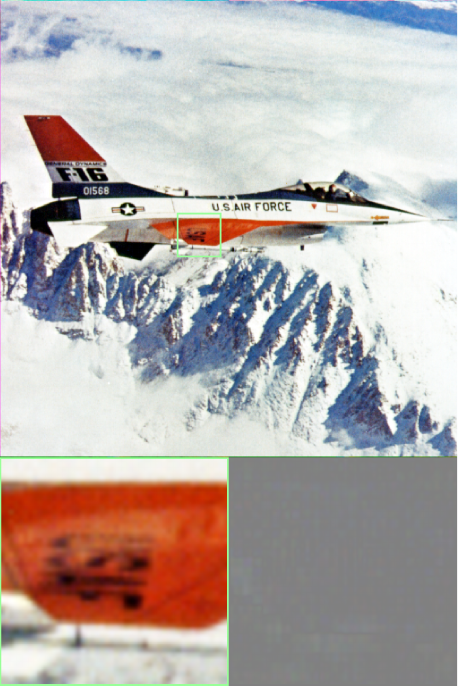}&
		\includegraphics[width=0.095\textwidth]{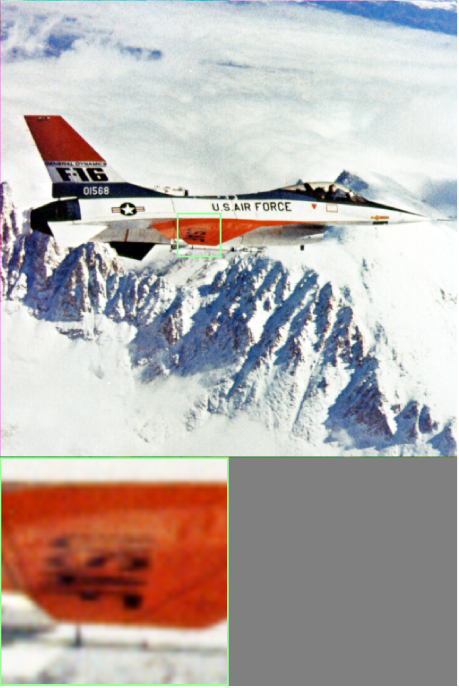}\\
		PSNR 36.38 & PSNR 41.33  & PSNR 41.40 & PSNR 52.80& PSNR Inf \\
		\includegraphics[width=0.095\textwidth]
		{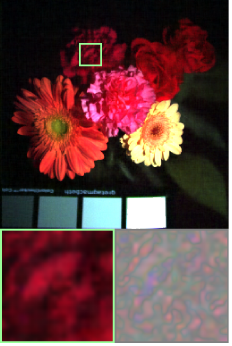}&
		\includegraphics[width=0.095\textwidth]{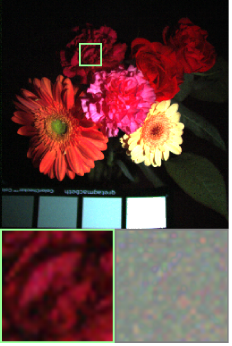}&
		\includegraphics[width=0.095\textwidth]{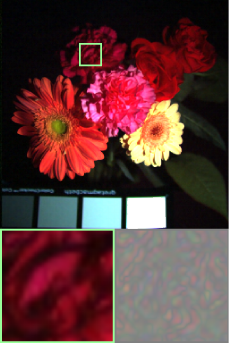}&
		\includegraphics[width=0.095\textwidth]{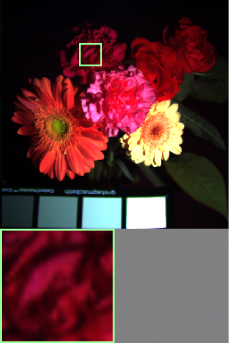}&
		\includegraphics[width=0.095\textwidth]{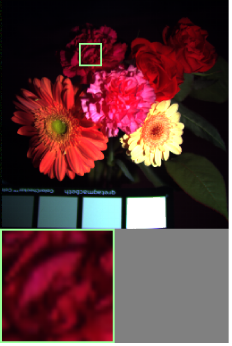}\\
		PSNR 42.88 & PSNR 44.07 & PSNR 47.47 & PSNR 57.29& PSNR Inf \\
		PE\cite{Fourier_feature}&WIRE\cite{Wire} &SIREN\cite{activation_function1} &CF-SIREN-INR&Original\\
	\end{tabular}
	\caption{Image regression using different INR methods on color image \textit{Airplane} and MSI \textit{Flowers}. The bottom left shows a zoomed-in view while the bottom right displays the corresponding error map.}
	\vspace{-0.5cm}
	\label{fig: fitting}
\end{figure}
\begin{table*}[tbp]
		\caption{
			The quantitative results by different methods for image inpainting. The best and the second best results are highlighted in \textbf{bold} and {\ul underline}, respectively.}
		\label{tab:MSI_inpainting}
		\centering
			\scriptsize
		\setlength{\tabcolsep}{2.5pt}
		\renewcommand{\arraystretch}{0.95}
		\begin{tabular}{ccccccccccccccccc}
			\hline
			\multicolumn{2}{c}{Sampling rate}                      & \multicolumn{3}{c}{0.1}                                                         & \multicolumn{3}{c}{0.15}                                                        & \multicolumn{3}{c}{0.2}                                                         & \multicolumn{3}{c}{0.25}                                                        & \multicolumn{3}{c}{0.3}                                                         \\ \hline
			Data                                  & Method         & \multicolumn{1}{c}{PSNR} & \multicolumn{1}{c}{SSIM} & \multicolumn{1}{c}{NRMSE} & \multicolumn{1}{c}{PSNR} & \multicolumn{1}{c}{SSIM} & \multicolumn{1}{c}{NRMSE} & \multicolumn{1}{c}{PSNR} & \multicolumn{1}{c}{SSIM} & \multicolumn{1}{c}{NRMSE} & \multicolumn{1}{c}{PSNR} & \multicolumn{1}{c}{SSIM} & \multicolumn{1}{c}{NRMSE} & \multicolumn{1}{c}{PSNR} & \multicolumn{1}{c}{SSIM} & \multicolumn{1}{c}{NRMSE} \\ \hline
			\multirow{8}{*}{\makecell{Color images\\ \textit{Peppers}\\ \textit{Sailboat}\\ \textit{Foster}\\$(512\times512\times3)$}}       & Observed       & 5.51                     & 0.023                    & 0.949                     & 5.76                     & 0.031                    & 0.922                     & 6.02                     & 0.038                    & 0.895                     & 6.30                     & 0.046                    & 0.866                     & 6.60                     & 0.054                    & 0.837                     \\
			& TCTV \cite{TCTV}          & 25.29                    & {\ul 0.633}              & 0.104                     & 26.71                    & {\ul 0.690}              & {\ul 0.087}               & 27.73                    & {\ul 0.723}              & 0.077                     & 28.61                    & {\ul 0.745}              & 0.070                     & 29.40                    & {\ul 0.773}              & 0.064                     \\
			& HLRTF\cite{HLRTF}          & 23.08                    & 0.424                    & 0.142                     & 24.75                    & 0.520                    & 0.115                     & 25.37                    & 0.547                    & 0.105                     & 26.12                    & 0.613                    & 0.094                     & 27.39                    & 0.659                    & 0.085                     \\
			& FTNN \cite{FTNN}          & 21.14                    & 0.337                    & 0.183                     & 23.19                    & 0.444                    & 0.142                     & 24.62                    & 0.530                    & 0.119                     & 25.85                    & 0.594                    & 0.102                     & 26.92                    & 0.650                    & 0.090                     \\
			& t-$\epsilon$-LogDet \cite{LogDet}  & 22.08                    & 0.430                    & 0.157                     & 23.80                    & 0.520                    & 0.126                     & 25.09                    & 0.575                    & 0.108                     & 26.13                    & 0.629                    & 0.095                     & 27.04                    & 0.669                    & 0.086                     \\
			& DIP \cite{DIP}           & {\ul 25.98}              & 0.574                    & {\ul 0.099}               & {\ul 26.99}              & 0.636                    & 0.089                     & {\ul 27.76}                    & 0.673                    & 0.080                     & 28.61                    & 0.721                    & 0.073                     & 29.30                    & 0.747                    & 0.067                     \\
			& LRTFR\cite{LRTFR}          & 25.25                    & 0.546                    & 0.109                     & 26.93                    & 0.629                    & 0.089                     & 27.69             & 0.685                    & {\ul 0.076}               & {\ul 28.64}              & 0.723                    & {\ul 0.069}               & {\ul 29.33}              & 0.758                    & {\ul 0.063}               \\
			& CF-INR & \textbf{26.84}           & \textbf{0.654}           & \textbf{0.091}            & \textbf{28.04}           & \textbf{0.706}           & \textbf{0.078}            & \textbf{28.89}           & \textbf{0.749}           & \textbf{0.071}            & \textbf{29.67}           & \textbf{0.777}           & \textbf{0.064}            & \textbf{30.36}           & \textbf{0.802}           & \textbf{0.059}            \\ \hline
			\multirow{8}{*}{\makecell{MSIs\\ \textit{Lemons}\\ \textit{Cups}\\ \textit{Face}\\$(256\times256\times31)$}}       & Observed       & 15.57                    & 0.190                    & 0.949                     & 15.82                    & 0.222                    & 0.922                     & 16.08                    & 0.253                    & 0.894                     & 16.36                    & 0.281                    & 0.866                     & 16.66                    & 0.307                    & 0.837                     \\
			& TCTV           & 40.93                    & {\ul 0.978}              & 0.065                     & 43.35                    & 0.985                    & 0.049                     & 45.25                    & 0.990                    & 0.040                     & 46.85                    & {\ul 0.992}              & 0.033                     & 48.27                    & {\ul 0.994}              & 0.029                     \\
			& HLRTF          & 40.76                    & 0.976                    & 0.060                     & 43.74                    & 0.985                    & 0.045                     & 45.75                    & 0.989                    & 0.035                     & 46.46                    & 0.991                    & 0.031                     & 48.50                    & {\ul 0.994}              & 0.025                     \\
			& FTNN           & 39.11                    & 0.967                    & 0.076                     & 41.65                    & 0.979                    & 0.058                     & 43.55                    & 0.985                    & 0.047                     & 45.25                    & 0.989                    & 0.039                     & 46.74                    & 0.991                    & 0.034                     \\
			& t-$\epsilon$-LogDet   & 37.56                    & 0.942                    & 0.095                     & 40.22                    & 0.965                    & 0.071                     & 41.98                    & 0.975                    & 0.058                     & 43.42                    & 0.981                    & 0.050                     & 44.74                    & 0.985                    & 0.043                     \\
			& DIP            & 41.26                    & 0.973                    & 0.059                     & 44.24                    & 0.983                    & {\ul 0.041}               & 45.89                    & 0.986                    & 0.036                     & 46.72                    & 0.986                    & 0.033                     & 47.41                    & 0.988                    & 0.030                     \\
			& LRTFR          & {\ul 41.71}              & {\ul 0.978}              & {\ul 0.057}               & {\ul 44.28}              & {\ul 0.986}              & {\ul 0.041}               & {\ul 46.45}              & {\ul 0.991}              & {\ul 0.033}               & {\ul 47.85}              & {\ul 0.992}              & {\ul 0.027}               & {\ul 49.18}              & {\ul 0.994}              & {\ul 0.023}               \\
			& CF-INR & \textbf{43.86}           & \textbf{0.989}           & \textbf{0.045}            & \textbf{46.26}           & \textbf{0.992}           & \textbf{0.034}            & \textbf{48.33}           & \textbf{0.994}           & \textbf{0.027}            & \textbf{49.83}           & \textbf{0.995}           & \textbf{0.023}            & \textbf{51.06}           & \textbf{0.996}           & \textbf{0.020}            \\ \hline
			\multirow{8}{*}{\makecell{HSIs\\ \textit{WDC mall}\\ \textit{PaviaU}\\ \textit{Xiongan}\\$(256\times256\times32)$}}      & Observed       & 15.00                    & 0.059                    & 0.949                     & 15.25                    & 0.082                    & 0.922                     & 15.51                    & 0.106                    & 0.894                     & 15.80                    & 0.130                    & 0.866                     & 16.09                    & 0.152                    & 0.837                     \\
			& TCTV           & 41.16                    & 0.960                    & 0.069                     & 43.72                    & {\ul 0.969}              & 0.052                     & 45.46                    & {\ul 0.974}              & 0.043                     & 46.76                    & {\ul 0.978}              & 0.038                     & 47.73                    & {\ul 0.981}              & 0.034                     \\
			& HLRTF          & 42.53                    & {\ul 0.961}              & {\ul 0.060}               & 44.62                    & 0.968                    & 0.047                     & 46.45                    & {\ul 0.974}              & {\ul 0.039}               & 47.26                    & {\ul 0.978}              & {\ul 0.036}               & 47.97                    & 0.980                    & {\ul 0.033}               \\
			& FTNN           & 39.87                    & 0.951                    & 0.074                     & 42.71                    & 0.964                    & 0.055                     & 44.72                    & 0.972                    & 0.045                     & 46.08                    & 0.976                    & 0.040                     & 47.51                    & 0.979                    & 0.034                     \\
			& t-$\epsilon$-LogDet   & 38.75                    & 0.941                    & 0.085                     & 41.17                    & 0.956                    & 0.066                     & 42.62                    & 0.963                    & 0.057                     & 43.69                    & 0.969                    & 0.051                     & 44.61                    & 0.973                    & 0.046                     \\
			& DIP            & {\ul 42.67}              & 0.959                    & {\ul 0.060}               & 44.69                    & 0.963                    & 0.050                     & 46.27                    & 0.967                    & 0.043                     & 46.90                    & 0.971                    & 0.040                     & 47.52                    & 0.973                    & 0.037                     \\
			& LRTFR          & 42.46                    & 0.957                    & {\ul 0.060}               & {\ul 44.73}              & 0.965                    & {\ul 0.046}               & {\ul 46.51}              & 0.971                    & {\ul 0.039}               & {\ul 47.42}              & 0.974                    & {\ul 0.036}               & {\ul 48.02}              & 0.977                    & {\ul 0.033}               \\
			& CF-INR & \textbf{43.77}           & \textbf{0.970}           & \textbf{0.051}            & \textbf{46.20}           & \textbf{0.974}           & \textbf{0.040}            & \textbf{47.71}           & \textbf{0.978}           & \textbf{0.034}            & \textbf{48.60}           & \textbf{0.981}           & \textbf{0.031}            & \textbf{49.26}           & \textbf{0.983}           & \textbf{0.028}            \\ \hline
			\multirow{8}{*}{\makecell{Videos\\ \textit{News}\\ \textit{Akiyo}\\ \textit{Container}\\$(288\times352\times3\times10)$}} & Observed       & 7.09                     & 0.024                    & 0.949                     & 7.33                     & 0.032                    & 0.922                     & 7.60                     & 0.040                    & 0.894                     & 7.88                     & 0.047                    & 0.866                     & 8.18                     & 0.056                    & 0.837                     \\
			& TCTV           & 35.74                    & {\ul 0.965}              & 0.037                     & 38.81                    & {\ul 0.979}              & 0.027                     & 41.03                    & {\ul 0.985}              & 0.021                     & 42.81                    & {\ul 0.988}              & {\ul 0.017}               & 44.28                    & {\ul 0.990}              & 0.015                     \\
			& HLRTF          & 35.14                    & 0.953                    & 0.042                     & 37.39                    & 0.969                    & 0.032                     & 39.15                    & 0.979                    & 0.027                     & 40.56                    & 0.984                    & 0.023                     & 41.97                    & 0.987                    & 0.020                     \\
			& FTNN           & 30.46                    & 0.907                    & 0.066                     & 33.14                    & 0.944                    & 0.049                     & 35.16                    & 0.960                    & 0.039                     & 36.86                    & 0.971                    & 0.032                     & 38.38                    & 0.978                    & 0.027                     \\
			& t-$\epsilon$-LogDet   & 34.89                    & 0.947                    & 0.042                     & 37.63                    & 0.966                    & 0.032                     & 39.32                    & 0.974                    & 0.027                     & 40.70                    & 0.980                    & 0.023                     & 41.76                    & 0.983                    & 0.020                     \\
			& DIP            & 36.43                    & 0.959                    & {\ul 0.035}               & {\ul 39.72}              & 0.977                    & {\ul 0.024}               & {\ul 41.36}              & 0.983                    & {\ul 0.020}               & 42.79                    & 0.986                    & 0.018                     & 43.98                    & 0.988                    & 0.015                     \\
			& LRTFR          & {\ul 35.83}              & 0.957                    & 0.037                     & 39.16                    & 0.975                    & 0.026                     & 41.18                    & 0.982                    & 0.021                     & {\ul 42.85}              & 0.986                    & {\ul 0.017}               & {\ul 44.32}              & 0.989                    & {\ul 0.014}               \\
			& CF-INR & \textbf{37.51}           & \textbf{0.971}           & \textbf{0.031}            & \textbf{40.52}           & \textbf{0.980}           & \textbf{0.022}            & \textbf{42.66}           & \textbf{0.986}           & \textbf{0.017}            & \textbf{44.40}           & \textbf{0.989}           & \textbf{0.014}            & \textbf{45.65}           & \textbf{0.991}           & \textbf{0.012}            \\ \hline
	\end{tabular}\vspace{-0.2cm}
\end{table*}
\begin{figure*}[t]
\setlength{\tabcolsep}{-1pt}
\centering
\scriptsize
\begin{tabular}{cccccccc}
\includegraphics[width=0.115\textwidth]
{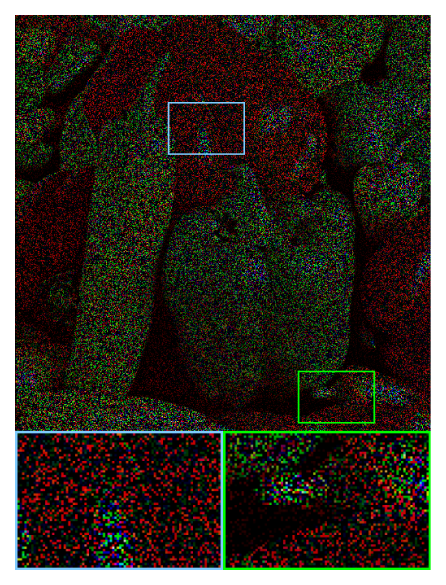}&
\includegraphics[width=0.115\textwidth]{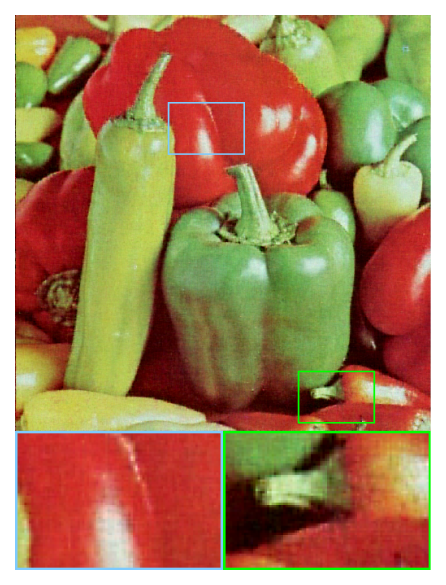}&
\includegraphics[width=0.115\textwidth]{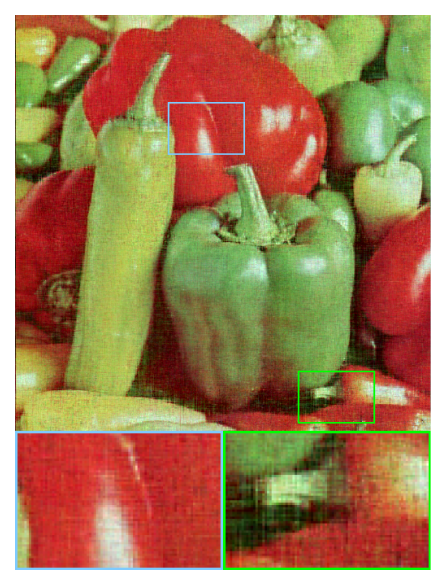}&
\includegraphics[width=0.115\textwidth]{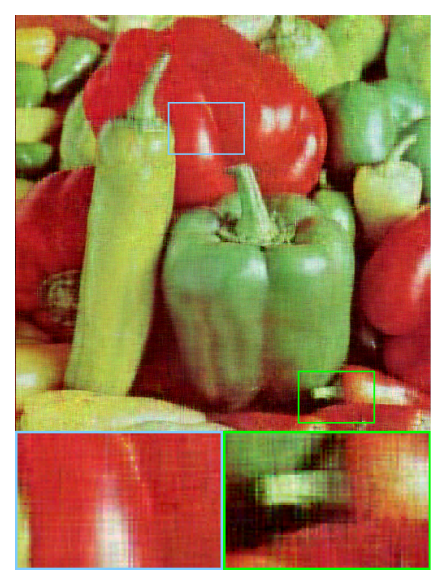}&
\includegraphics[width=0.115\textwidth]{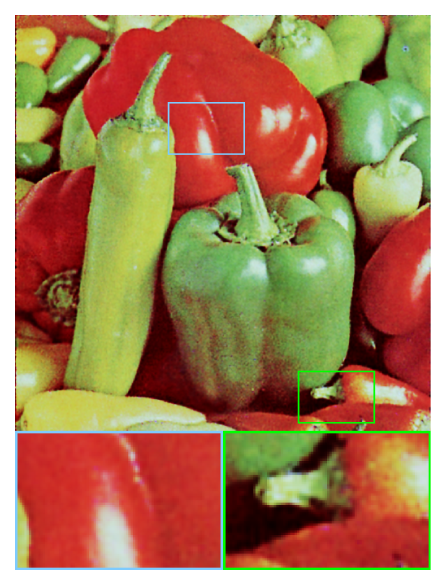}&
\includegraphics[width=0.115\textwidth]{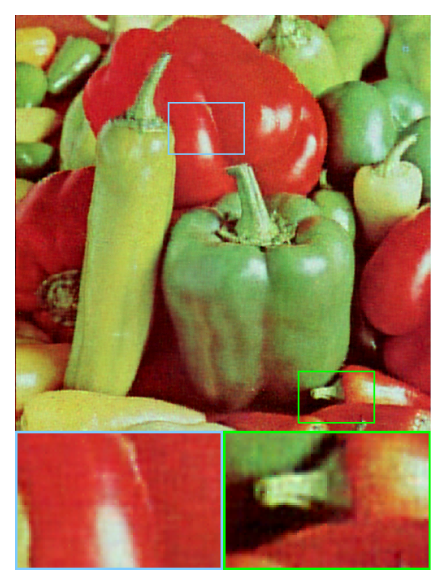}&
\includegraphics[width=0.115\textwidth]{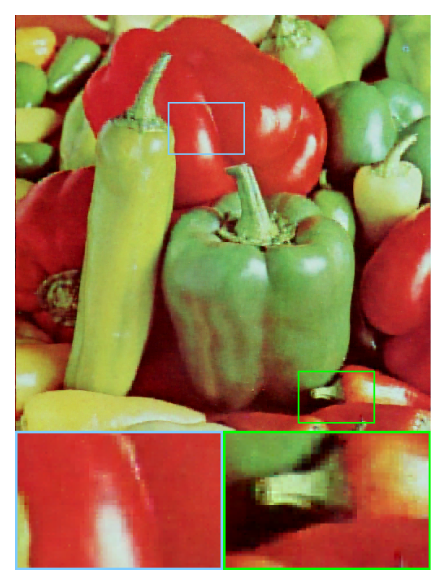}&
\includegraphics[width=0.115\textwidth]{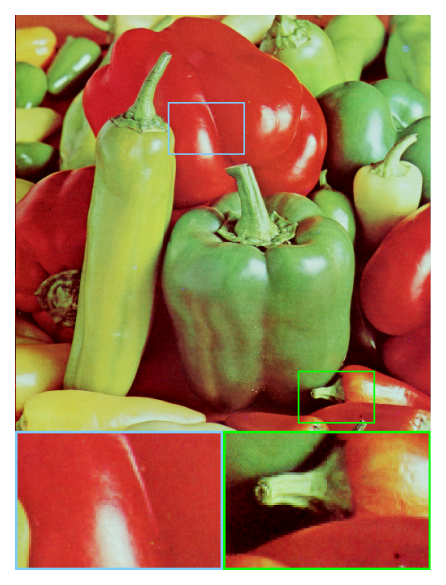}\\
PSNR 7.52 & PSNR 28.26 & PSNR 24.57 & PSNR 24.97 & PSNR 28.19 & PSNR 28.03 & PSNR 29.26
 & PSNR Inf \\
\includegraphics[width=0.115\textwidth]
{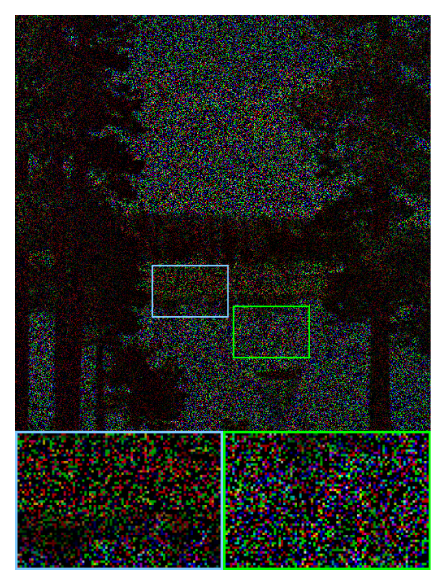}&
\includegraphics[width=0.115\textwidth]{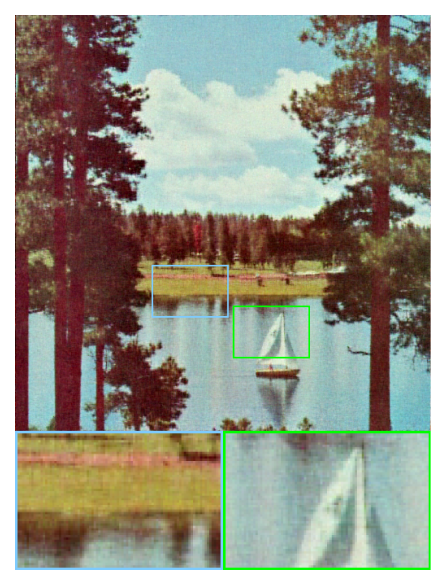}&
\includegraphics[width=0.115\textwidth]{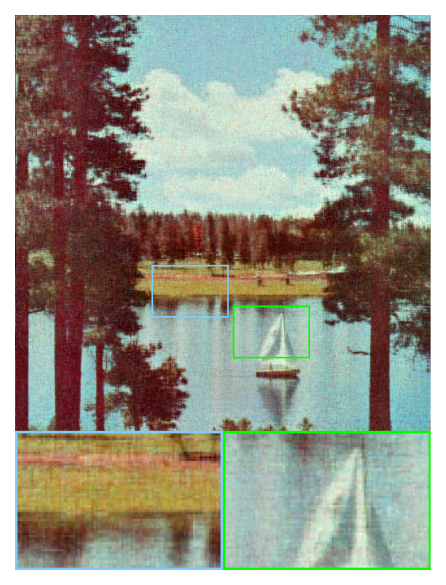}&
\includegraphics[width=0.115\textwidth]{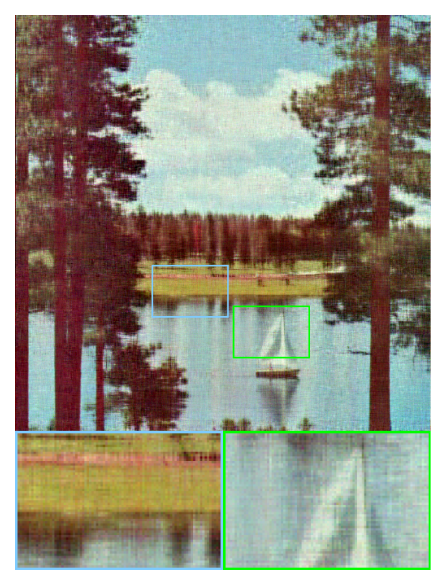}&
\includegraphics[width=0.115\textwidth]{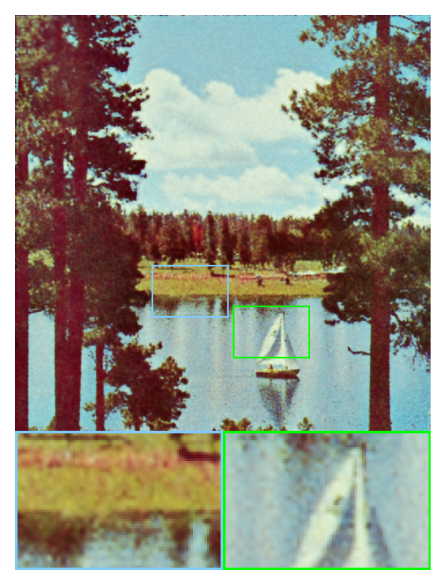}&
\includegraphics[width=0.115\textwidth]{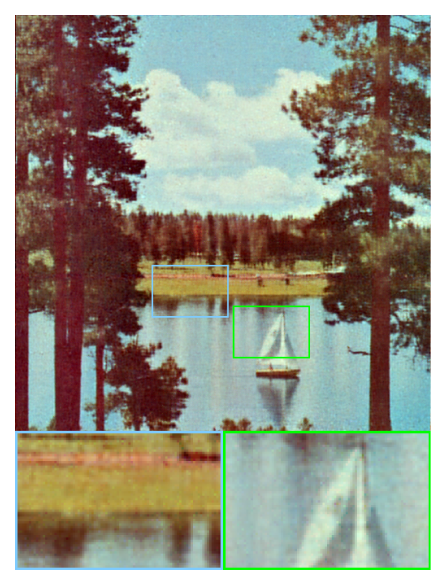}&
\includegraphics[width=0.115\textwidth]{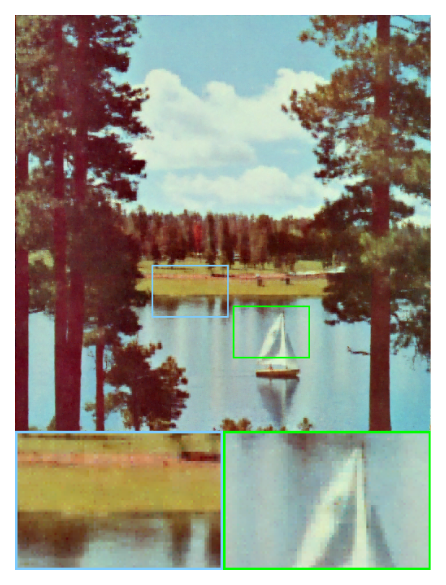}&
\includegraphics[width=0.115\textwidth]{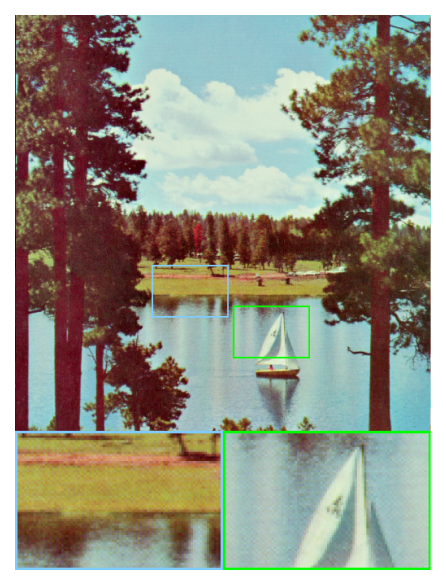}\\
PSNR 6.13 & PSNR 26.15 & PSNR 23.95 & PSNR 23.15 & PSNR 25.96 & PSNR 25.82 & PSNR 26.84
 & PSNR Inf \\

\includegraphics[width=0.115\textwidth]{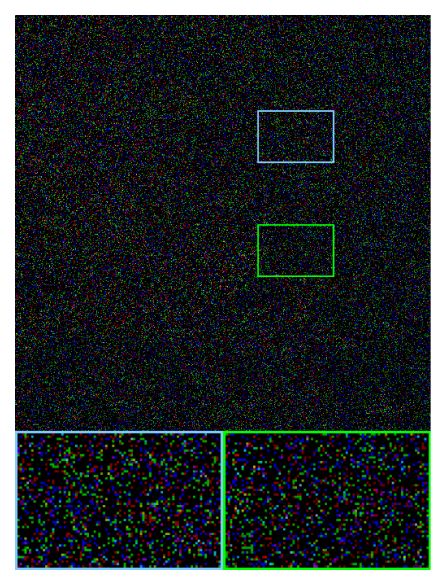}&
\includegraphics[width=0.115\textwidth]{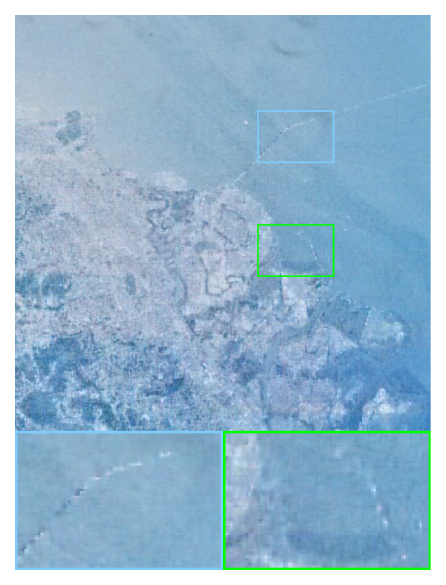}&
\includegraphics[width=0.115\textwidth]{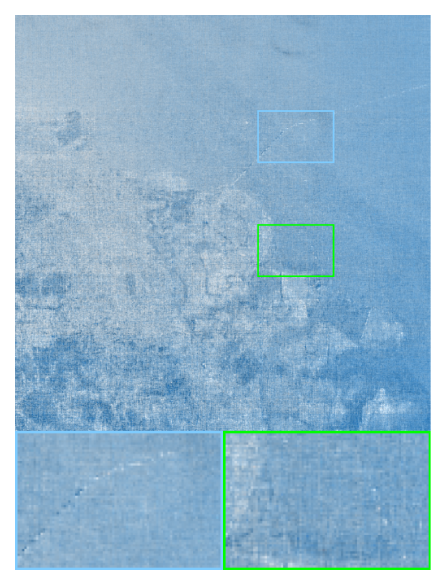}&
\includegraphics[width=0.115\textwidth]{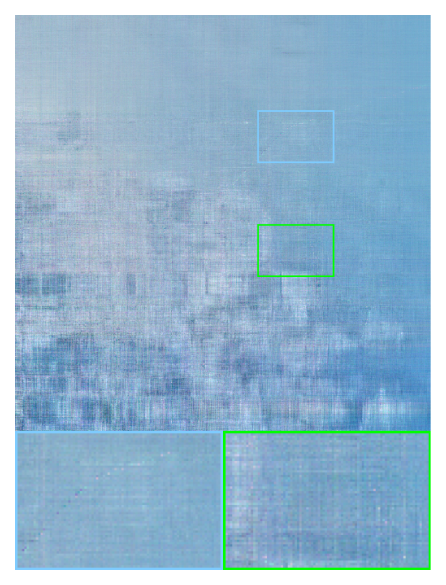}&
\includegraphics[width=0.115\textwidth]{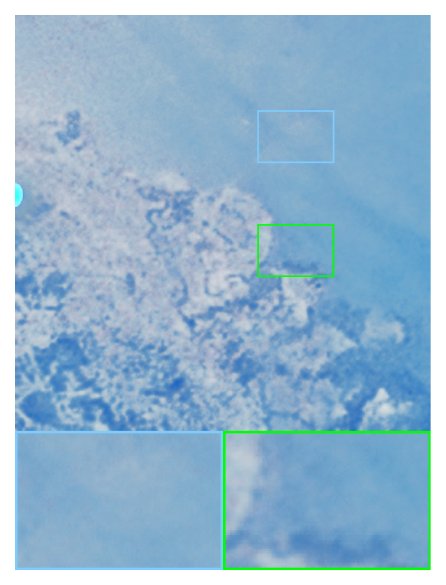}&
\includegraphics[width=0.115\textwidth]{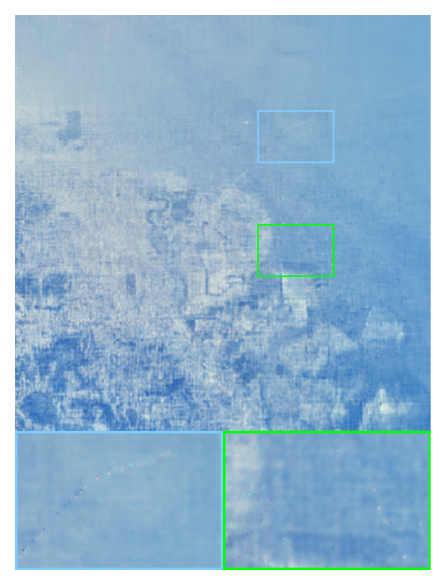}&
\includegraphics[width=0.115\textwidth]{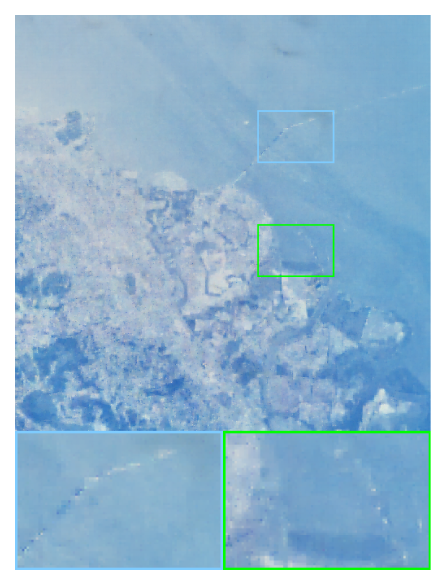}&
\includegraphics[width=0.115\textwidth]{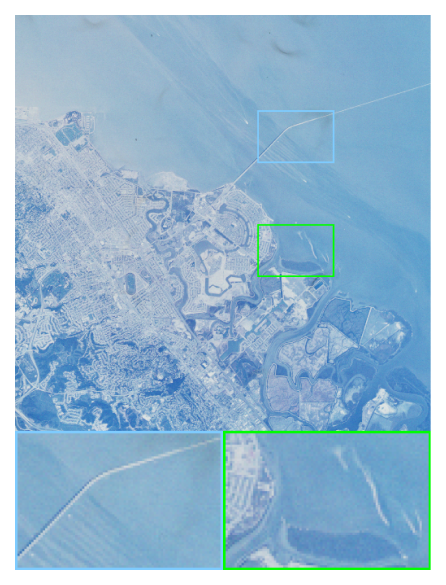}\\
PSNR 4.41 & PSNR 28.78 & PSNR 27.58  & PSNR 27.16 & PSNR 29.13 & PSNR 29.22 & PSNR 30.58
 & PSNR Inf \\
Observed&TCTV\cite{TCTV}&HLRTF\cite{HLRTF}&t-$\epsilon$-LogDet\cite{LogDet}&DIP\cite{DIP}&LRTFR\cite{LRTFR} &CF-INR&Original\\
\end{tabular}
\caption{From top to bottom: the image inpainting results on color images \textit{Peppers}, \textit{Sailboat}, and \textit{Foster} with SR of 0.2.}
\vspace{-0.2cm}
    \label{fig: inpainting RGB}
\end{figure*}

For the image regression (i.e., using INR to continuously represent images), we integrate CF-INR into several well-established INR methods, including positional encoding (PE) \cite{Fourier_feature}, WIRE \cite{Wire} based on Gabor wavelet basis function, SIREN \cite{activation_function1} based on sinusoidal function, and FINER \cite{FINER} based on variable-periodic activation function. We evaluate the effectiveness of such integrations in Table \ref{tab:fitting}. All models are trained under identical experimental conditions using the Adam optimizer. The testing datasets include three color images \textit{House}, \textit{Airplane}, and \textit{Mandrill}\footnote{[Online]. Available: \url{http://sipi.usc.edu/database/database.php}\label{color_image}}, and three multispectral images (MSIs) from the CAVE dataset \textit{Flowers}, \textit{House}, and \textit{Peppers}\footnote{[Online]. Available: \url{https://cave.cs.columbia.edu/repository/Multispectral}\label{CAVE_image}}.
It can be observed from Table \ref{tab:fitting} that the performance of each INR method (PE, WIRE, SIREN, and FINER) is significantly improved after integrating with the proposed CF-INR, indicating that CF-INR effectively improves existing INR approaches by further capturing multi-frequency characteristics. The integration also highlights the flexibility of CF-INR as a general continuous representation framework with broad applicability. In Fig. \ref{fig: nihe_curves}, we plot the PSNR curves over training epochs of different INR methods. The proposed CF-INR holds consistently better PSNR values across epochs, achieving the fastest convergence. Fig. \ref{fig: fitting} shows some visual examples, where CF-INR obtains evidently better image representation accuracy by virtue of the multi-frequency feature characterization. Considering the overall better performance of CF-SIREN-INR, we adopt the sinusoidal activation function with cross-frequency ranks and cross-frequency parameters self-evolving strategies for CF-INR throughout the remainder of the experiments.

\subsection{Image Inpainting Results}

Multi-dimensional image inpainting represents a prototypical data recovery challenge. To assess the performance of CF-INR for image inpainting, we benchmark it against three state-of-the-art model-based methods (TCTV\cite{TCTV}, t-$\epsilon$-LogDet\cite{LogDet}, and FTNN\cite{FTNN}), two deep learning-based methods (HLRTF\cite{HLRTF} and DIP\cite{DIP}), and a recently developed INR-based method (LRTFR\cite{LRTFR}). The testing datasets consist of three color images: \textit{Peppers}, \textit{Sailboat}, and \textit{Foster}\footref{color_image}, as well as three MSIs from the CAVE dataset: \textit{Lemon}, \textit{Cups}, and \textit{Face}\footref{CAVE_image}. We also include three classical hyperspectral images (HSIs): \textit{Washington DC Mall} (\textit{WDC mall})\footnote{[Online]. Available: \url{https://engineering.purdue.edu/~biehl/MultiSpec/}}, \textit{Pavia University} (\textit{PaviaU})\footnote{[Online]. Available: \url{https://lesun.weebly.com/hyperspectral-data-set.html}\label{HSI_image}}, and \textit{Xiongan}\cite{xu2024stacked}, as well as three video datasets: \textit{News}, \textit{Akiyo}, and \textit{Container}\footnote{[Online]. Available: \url{http://trace.eas.asu.edu/yuv/}}. All HSIs are preprocessed with size $256\times256\times32$. We consider scenarios with random missing values at sampling rates (SRs) of 0.1, 0.15, 0.2, 0.25, and 0.3. The quantitative and some qualitative results of multi-dimensional image inpainting are shown in Table \ref{tab:MSI_inpainting} and Fig. \ref{fig: inpainting RGB}. CF-INR consistently outperforms competing approaches in quantitative evaluations. Especially, CF-INR outperforms LRTFR, a state-of-the-art INR-based method, with PSNR gains around 1.5 dB. This performance advantage demonstrates that integrating the frequency decoupling of CF-INR yields substantial improvements by enhancing cross-frequency feature characterizations. Additionally, restored images obtained by CF-INR exhibit superior edge preservation effects (Fig. \ref{fig: inpainting RGB}), further demonstrating the advantage of CF-INR for capturing multi-frequency components.
\begin{table*}[tbp]
	\caption{
		The quantitative results by different methods for hyperspectral mixed noise removal. The best and the second best results are highlighted in \textbf{bold} and {\ul underline}, respectively.}
	\label{tab:MSI_denoising}
	\scriptsize
	\centering
	\setlength{\tabcolsep}{2.7pt}
	\renewcommand{\arraystretch}{0.9}
	\begin{tabular}{ccccccccccccccccc}
		\hline
		\multicolumn{2}{c}{Noisy}             & \multicolumn{3}{c}{Case1}                                                       & \multicolumn{3}{c}{Case2}                                                       & \multicolumn{3}{c}{Case3}                                                       & \multicolumn{3}{c}{Case4}                                                       & \multicolumn{3}{c}{Case5}                                                       \\ \hline
		Data                 & Method         & \multicolumn{1}{c}{PSNR} & \multicolumn{1}{c}{SSIM} & \multicolumn{1}{c}{NRMSE} & \multicolumn{1}{c}{PSNR} & \multicolumn{1}{c}{SSIM} & \multicolumn{1}{c}{NRMSE} & \multicolumn{1}{c}{PSNR} & \multicolumn{1}{c}{SSIM} & \multicolumn{1}{c}{NRMSE} & \multicolumn{1}{c}{PSNR} & \multicolumn{1}{c}{SSIM} & \multicolumn{1}{c}{NRMSE} & \multicolumn{1}{c}{PSNR} & \multicolumn{1}{c}{SSIM} & \multicolumn{1}{c}{NRMSE} \\ \hline
		\multirow{8}{*}{\makecell{MSIs\\ \textit{Cloth}\\ \textit{Stones}\\ \textit{Pills}\\$(256\times256\times31)$}}  & Observed       & 16.79                    & 0.255                    & 0.712                     & 14.69                    & 0.223                    & 1.058                     & 16.99                    & 0.282                    & 0.712                     & 16.48                    & 0.283                    & 0.769                     & 14.23                    & 0.209                    & 0.985                     \\
		& TCTV \cite{TCTV}          & 25.97                    & 0.647                    & 0.247                     & 25.70                    & 0.616                    & 0.258                     & 24.63                    & 0.604                    & 0.306                     & 23.34                    & 0.623                    & 0.400                     & 22.62                    & 0.635                    & 0.389                     \\
		& RCTV \cite{RCTV}          & 25.33                    & 0.681                    & 0.272                     & 26.34                    & 0.695                    & 0.245                     & {\ul 26.99}              & 0.722                    & {\ul 0.227}               & {\ul 27.07}              & {\ul 0.725}              & {\ul 0.226}               & 26.95                    & 0.720                    & 0.228                     \\
		& HLRTF \cite{HLRTF}         & 25.52                    & 0.641                    & 0.265                     & 24.95                    & 0.623                    & 0.292                     & 25.58                    & 0.655                    & 0.278                     & 25.76                    & 0.669                    & 0.282                     & 26.09                    & 0.663                    & 0.251                     \\
		& HIR-Diff \cite{HIR-Diff}      & 25.85                    & 0.746                    & 0.254                     & 20.41                    & 0.538                    & 0.519                     & 25.66                    & {\ul 0.741}              & 0.267                     & 24.72                    & 0.714                    & 0.301                     & 23.26                    & 0.665                    & 0.359                     \\
		& RCILD \cite{RCILD}         & 26.67                    & {\ul 0.774}              & 0.231                     & 23.84                    & 0.703                    & 0.394                     & 25.93                    & 0.725                    & 0.265                     & 24.08                    & 0.699                    & 0.365                     & 23.04                    & 0.670                    & 0.377                     \\
		& LRTFR \cite{LRTFR}         & {\ul 27.15}              & 0.745                    & {\ul 0.221}               & {\ul 28.06}              & {\ul 0.747}              & {\ul 0.202}               & 26.71                    & 0.715                    & 0.242                     & 26.45                    & 0.708                    & 0.249                     & {\ul 28.05}              & {\ul 0.747}              & {\ul 0.202}               \\
		& CF-INR & \textbf{29.04}           & \textbf{0.825}           & \textbf{0.174}            & \textbf{29.92}           & \textbf{0.818}           & \textbf{0.159}            & \textbf{29.73}           & \textbf{0.812}           & \textbf{0.166}            & \textbf{29.75}           & \textbf{0.822}           & \textbf{0.170}            & \textbf{29.97}           & \textbf{0.820}           & \textbf{0.159}            \\ \hline
		\multirow{8}{*}{\makecell{HSIs\\ \textit{WDC mall}\\ \textit{PaviaU}\\ \textit{PaviaC}\\$(256\times256\times32)$}} & Observed       & 15.77          & 0.164          & 1.044          & 13.40          & 0.133          & 1.525          & 15.52          & 0.172          & 1.087          & 15.05          & 0.170          & 1.164          & 13.22          & 0.125          & 1.443          \\
		& TCTV           & 24.86          & 0.561          & 0.363          & 24.60          & 0.540          & 0.378          & 23.93          & 0.535          & 0.420          & 22.48          & 0.549          & 0.562          & 22.36          & 0.564          & 0.521          \\
		& RCTV           & 26.23          & 0.663          & 0.308          & 28.17          & 0.690          & 0.246          & {\ul 28.40}    & 0.715          & {\ul 0.240}    & {\ul 28.47}    & 0.725          & {\ul 0.241}    & 28.31          & 0.703          & 0.242          \\
		& HLRTF          & 25.75          & 0.639          & 0.334          & 26.57          & 0.646          & 0.307          & 27.15          & 0.682          & 0.305          & 27.39          & 0.700          & 0.308          & 27.03          & 0.671          & 0.293          \\
		& HIR-Diff       & 27.20          & 0.764          & 0.287          & 21.98          & 0.541          & 0.602          & 25.82          & {\ul 0.768}    & 0.350          & 25.28          & {\ul 0.736}    & 0.378          & 23.43          & 0.522          & 0.472          \\
		& RCILD          & {\ul 27.79}    & {\ul 0.789}    & 0.268          & 23.36          & 0.715          & 0.539          & 25.89          & 0.758          & 0.348          & 23.83          & 0.715          & 0.501          & 22.93          & 0.698          & 0.504          \\
		& LRTFR          & 27.65          & 0.755          & {\ul 0.263}    & {\ul 29.61}    & {\ul 0.790}    & {\ul 0.209}    & 28.10          & 0.735          & 0.263          & 27.74          & 0.719          & 0.272          & {\ul 29.17}    & {\ul 0.778}    & {\ul 0.223}    \\
		& CF-INR & \textbf{28.57} & \textbf{0.799} & \textbf{0.238} & \textbf{30.80} & \textbf{0.826} & \textbf{0.183} & \textbf{31.06} & \textbf{0.837} & \textbf{0.178} & \textbf{31.02} & \textbf{0.838} & \textbf{0.184} & \textbf{30.85} & \textbf{0.831} & \textbf{0.183} \\ \hline
	\end{tabular}
    \vspace{-0.4cm}
\end{table*}
\subsection{Hyperspectral Mixed Noise Removal Results}
HSI mixed noise removal is another challenging data recovery problem in image processing\cite{LRTFR,RCTV}. We compare CF-INR with three model-based methods (TCTV\cite{TCTV}, RCTV\cite{RCTV}, and HLRTF\cite{HLRTF}), two state-of-the-art deep learning-based methods (HIR-Diff\cite{HIR-Diff} and RCILD\cite{RCILD}), and LRTFR\cite{LRTFR}. Here, HIR-Diff is a diffusion model-based method and RCILD is a supervised deep-learning method. The testing datasets include three MSIs from the CAVE dataset, namely \textit{Cloth}, \textit{Stones}, and \textit{Pills}\footref{CAVE_image}, and three HSIs \textit{Washington DC Mall} (\textit{WDC mall}), \textit{Pavia University} (\textit{PaviaU}), and \textit{Pavia Centre} (\textit{PaviaC})\footref{HSI_image}. We consider five mixed noisy cases. Case 1: The data is corrupted by Gaussian noise with standard deviation of $\sigma=0.2$ and by sparse noise\cite{LRTFR} with SR of 0.1; Case 2: The data is corrupted by Gaussian noise with standard deviation of $\sigma=0.2$, and one-third of the bands are corrupted by sparse noise with varying SRs between $[30\%, 60\%]$; Case 3: The data is corrupted by Gaussian noise with standard deviation of $\sigma=0.2$, and one-third of the bands are corrupted by stripe noise \cite{impulse_noise} with varying ratios between $[10\%, 20\%]$, with the stripe noise magnitude confined to the range \([-0.5, 0.5]\); Case 4: The data is corrupted by Gaussian noise with standard deviation of $\sigma=0.2$, and one-third of the bands are corrupted by deadline noise\cite{impulse_noise} with varying ratios between $[10\%, 20\%]$; Case 5: The data is corrupted by Gaussian noise with standard deviation of $\sigma=0.2$ and by a mixture of sparse, stripe, and deadline noise described in the preceding cases. Following \cite{LRTFR}, for cases with deadline noise we employ an additional mask operator in the loss function.\par
\begin{figure*}[t]
	\setlength{\tabcolsep}{-1pt}
	\centering
	\scriptsize
	\begin{tabular}{cccccccc}
		\includegraphics[width=0.115\textwidth]
		{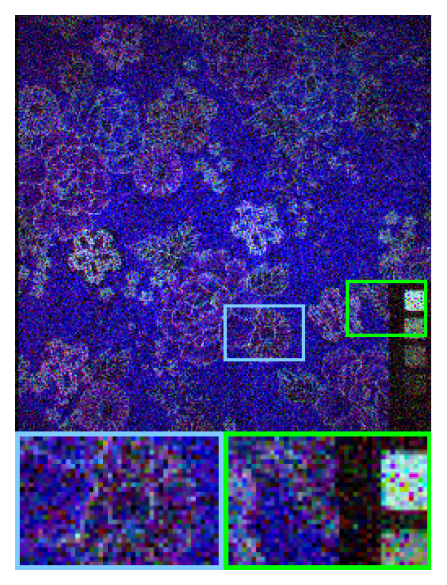}&
		\includegraphics[width=0.115\textwidth]{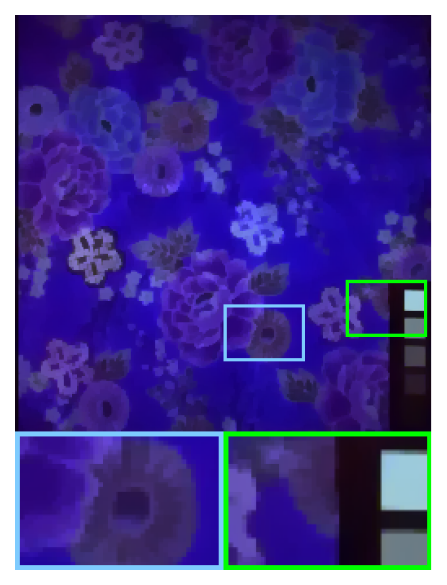}&
		\includegraphics[width=0.115\textwidth]{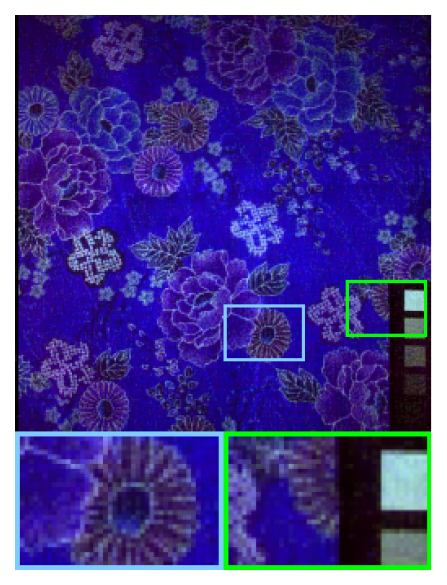}&
		\includegraphics[width=0.115\textwidth]{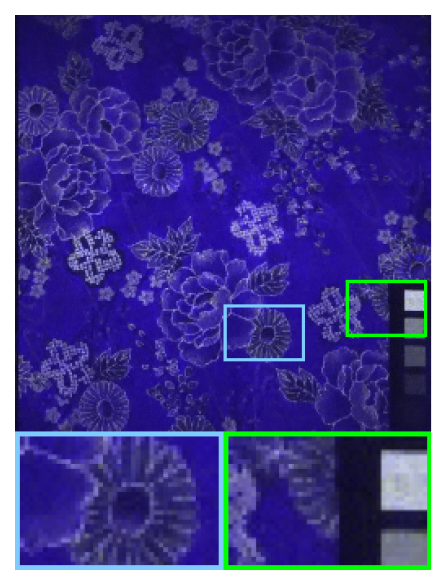}&
		\includegraphics[width=0.115\textwidth]{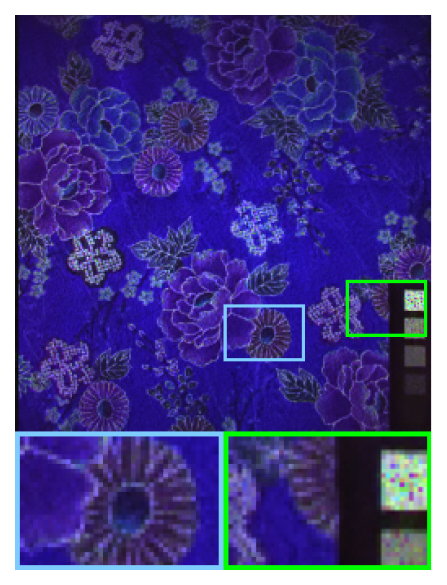}&
		\includegraphics[width=0.115\textwidth]{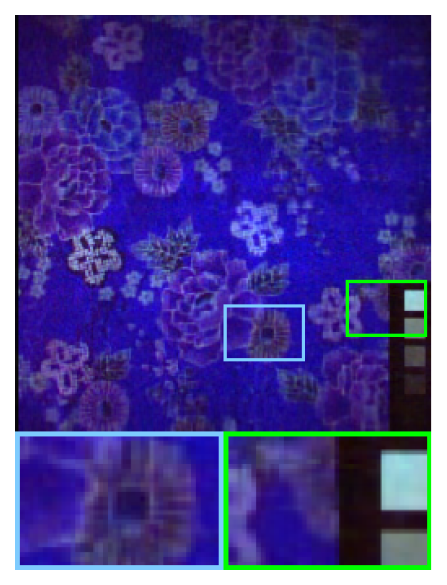}&
		\includegraphics[width=0.115\textwidth]{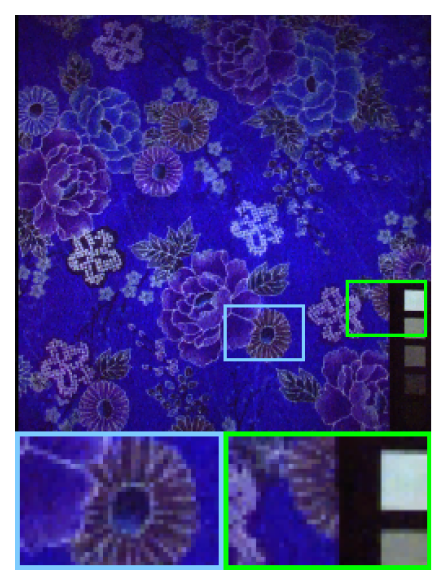}&
		\includegraphics[width=0.115\textwidth]{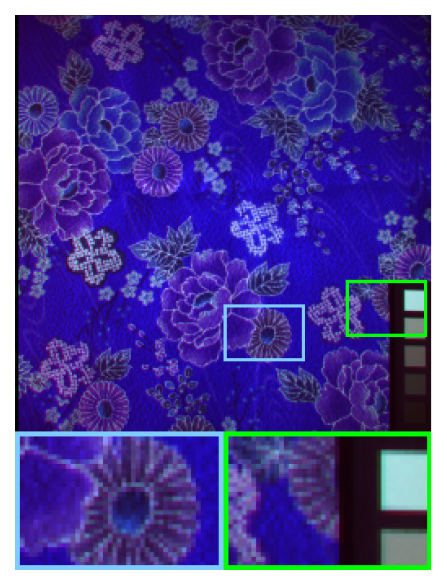}\\
		PSNR 19.45 & PSNR 25.03 & PSNR 27.77 & PSNR 27.79 & PSNR 28.84 & PSNR 26.59 & PSNR 30.25 & PSNR Inf \\

		\includegraphics[width=0.115\textwidth]
		{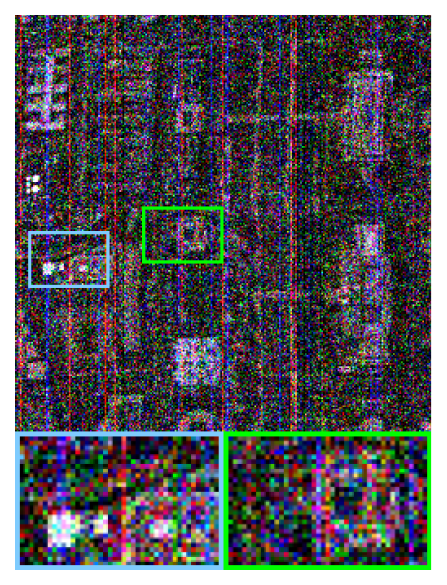}&
		\includegraphics[width=0.115\textwidth]{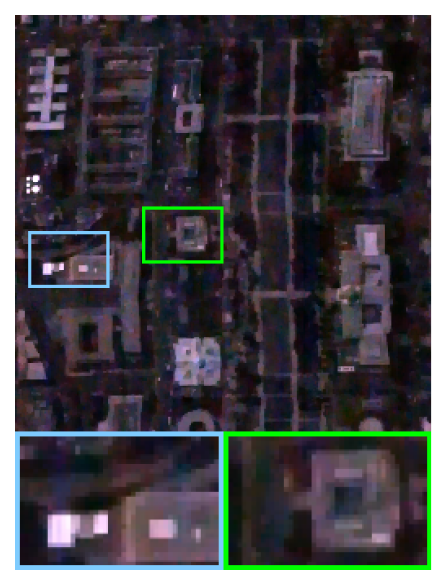}&
		\includegraphics[width=0.115\textwidth]{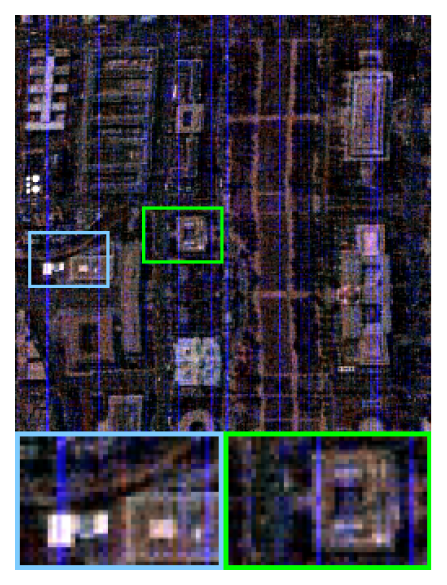}&
		\includegraphics[width=0.115\textwidth]{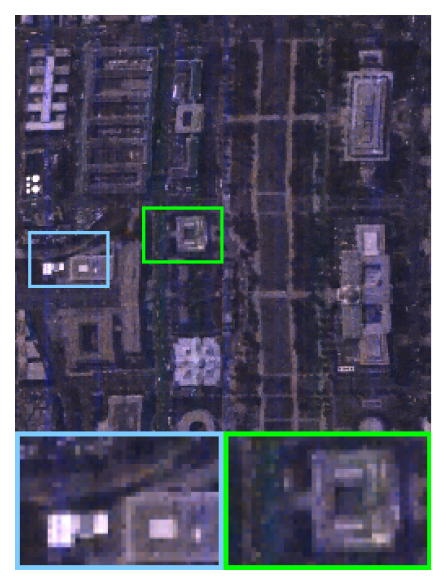}&
		\includegraphics[width=0.115\textwidth]{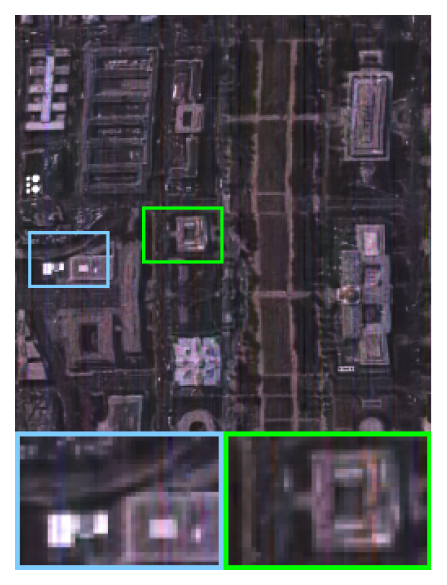}&
		\includegraphics[width=0.115\textwidth]{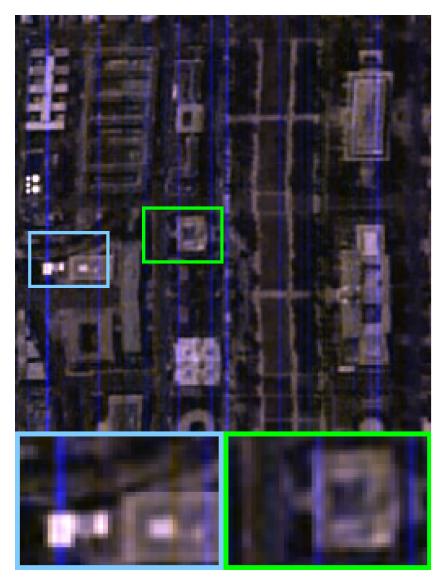}&
		\includegraphics[width=0.115\textwidth]{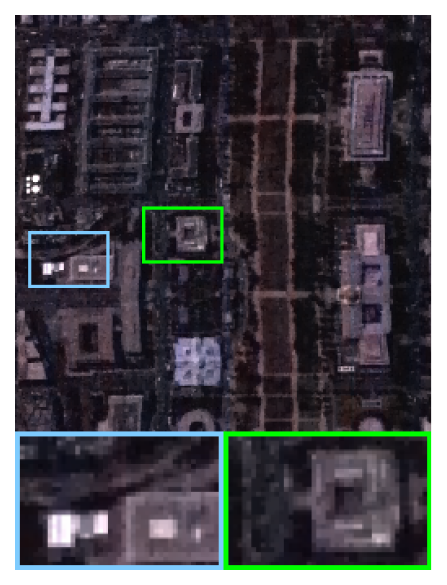}&
		\includegraphics[width=0.115\textwidth]{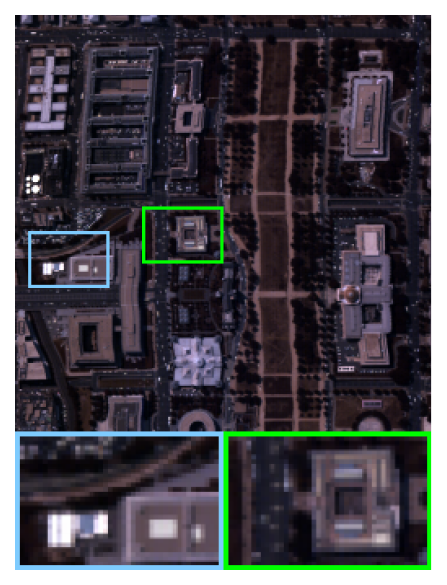}\\
		PSNR 15.76 & PSNR 28.95 & PSNR 27.87 & PSNR 25.03 & PSNR 25.11 & PSNR 28.35 & PSNR 31.37 & PSNR Inf \\

		Observed&RCTV\cite{RCTV}&HLRTF\cite{HLRTF}&HIR-Diff\cite{HIR-Diff}&RCILD\cite{RCILD}&LRTFR\cite{LRTFR} &CF-INR&Original\\
	\end{tabular}
	\caption{From top to bottom: the image denoising results on MSI \textit{Cloth} (Case 1) and HSI \textit{WDC mall} (Case 3).}\vspace{-0.4cm}
	\label{fig:MSI_denoising}
\end{figure*}

Table \ref{tab:MSI_denoising} and Fig. \ref{fig:MSI_denoising} present the quantitative and qualitative results for HSI denoising. CF-INR consistently demonstrates better quantitative performances under various noise scenarios and across multiple datasets, as shown in Table \ref{tab:MSI_denoising}. Especially, the unsupervised CF-INR is able to outperform the pretrained diffusion model-based method HIR-Diff \cite{HIR-Diff} and the supervised deep learning method RCILD \cite{RCILD}. The strong capability of CF-INR comes from its advantage for delicate cross-frequency modeling of the tensor, which improves edge and details recovery from noisy measurements. Fig. \ref{fig:MSI_denoising} illustrates that CF-INR simultaneously captures fine details of images and attenuates mixed noise better than its baseline LRTFR. Overall, the frequency decoupling of CF-INR facilitates effective multi-frequency feature utilization, hence could better extract the underlying clean image structure.

\subsection{Multi-Temporal Cloud Removal Results}
\begin{figure*}[t]
	\setlength{\tabcolsep}{0.3pt}
	\centering
	\scriptsize
	\begin{tabular}{cccccccc}
		\includegraphics[width=0.11\textwidth]
		{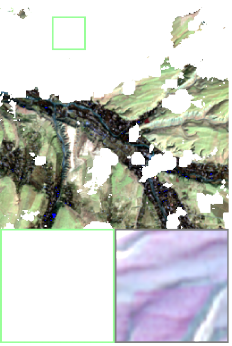}&
		\includegraphics[width=0.11\textwidth]{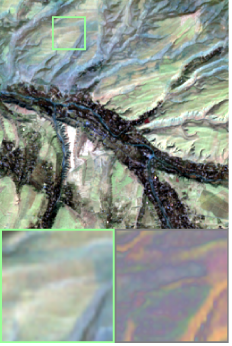}&
		\includegraphics[width=0.11\textwidth]{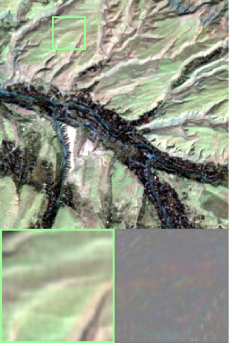}&
		\includegraphics[width=0.11\textwidth]{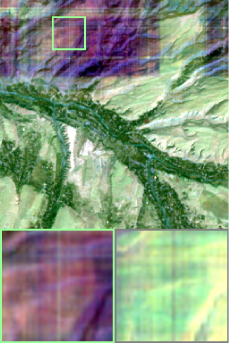}&
		\includegraphics[width=0.11\textwidth]{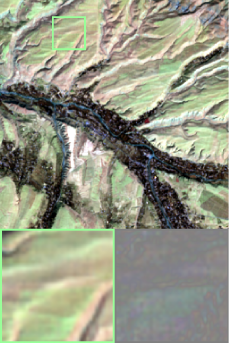}&
		\includegraphics[width=0.11\textwidth]{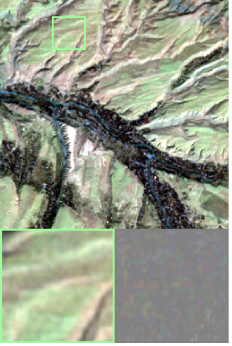}&
		\includegraphics[width=0.11\textwidth]{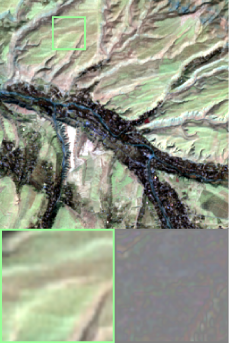}&
		\includegraphics[width=0.11\textwidth]{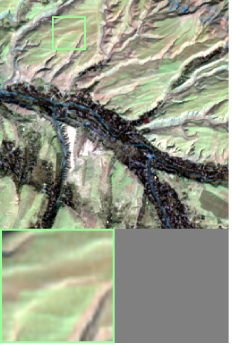}\\
		PSNR 14.54 & PSNR 35.98 & PSNR 37.78 & PSNR 33.15 & PSNR 34.63 & PSNR 37.83 & PSNR 39.03 & PSNR Inf \\
		\includegraphics[width=0.11\textwidth]
		{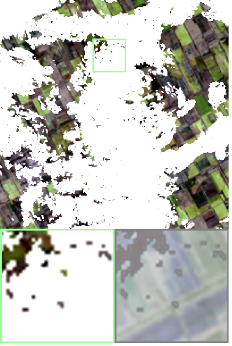}&
		\includegraphics[width=0.11\textwidth]{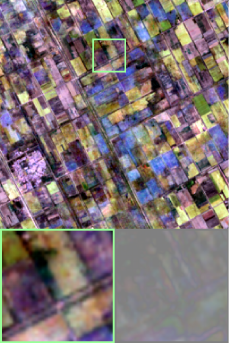}&
		\includegraphics[width=0.11\textwidth]{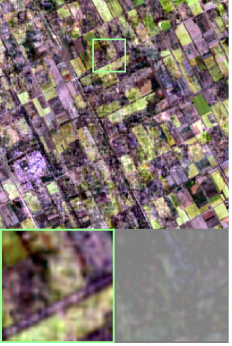}&
		\includegraphics[width=0.11\textwidth]{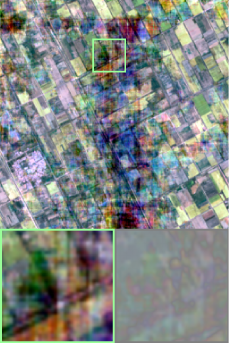}&
		\includegraphics[width=0.11\textwidth]{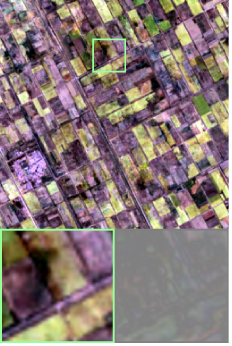}&
		\includegraphics[width=0.11\textwidth]{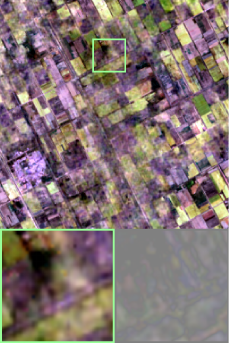}&
		\includegraphics[width=0.11\textwidth]{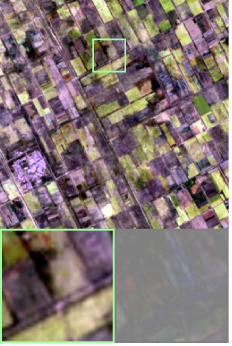}&
		\includegraphics[width=0.11\textwidth]{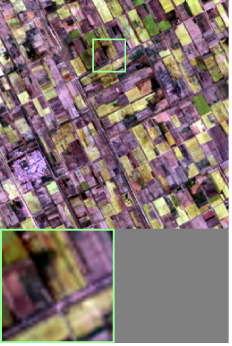}\\
		PSNR 15.12 & PSNR 31.71 & PSNR 32.24 & PSNR 28.18 & PSNR 31.11 & PSNR 32.13 & PSNR 33.84 & PSNR Inf \\
		
		Observed&TCTV\cite{TCTV}&HLRTF\cite{HLRTF}&t-$\epsilon$-LogDet\cite{LogDet}&DIP\cite{DIP}&LRTFR\cite{LRTFR} &CF-INR&Original\\
	\end{tabular}
	\caption{From top to bottom: the cloud removal results on remote sensing images \textit{Jizzakh1} and \textit{Jizzakh2} at the first and fifth nodes, respectively. The bottom left corner shows a zoomed-in view while the bottom right displays the corresponding error map.}
	\vspace{-0.4cm}
	\label{fig:cloud_removal}
\end{figure*}

Multi-temporal remote sensing cloud removal is a distinct and challenging task in multi-dimensional image processing. We conduct multi-temporal cloud removal experiments to evaluate the effectiveness of the CF-INR method. The testing datasets include two images from the \textit{Jizzakh} dataset and one from the \textit{Puyang} dataset\cite{xu2024stacked}. All datasets are preprocessed to a uniform size of $256 \times 256$ pixels, with 7 spectral bands and 9 temporal snapshots. We select nine real cloud masks with varying cloud cover levels from the remote sensing cloud detection dataset\cite{xu2024stacked}. We compare our method with TCTV\cite{TCTV}, HLRTF\cite{HLRTF}, t-$\epsilon$-LogDet\cite{LogDet}, FTNN\cite{FTNN}, DIP\cite{DIP}, and LRTFR\cite{LRTFR}. Table \ref{tab:Cloud_Removal} and Fig. \ref{fig:cloud_removal} present the quantitative and some qualitative results of multi-temporal cloud removal. CF-INR achieves the best evaluation metrics across datasets. To better illustrate the effectiveness of CF-INR, we select three time points with the largest cloud masks for result visualization (Fig. \ref{fig:cloud_removal}). Notably, TCTV and t-$\epsilon$-LogDet methods exhibit certain degrees of color distortion, while HLRTF, DIP, and LRTFR methods are capable of recovering coarse structures but fail to preserve fine textures. As compared, CF-INR produces the most visually favorable results, reflecting its capability to handle large missing areas in the cloud removal task. The superiority of CF-INR confirms its efficacy for capturing multi-frequency features underlying multi-dimensional visual data.

\subsection{Discussions\label{sec: Discussions}}
\subsubsection{Effectiveness of Generative Optimization Model}\label{compare_model}
First we evaluate the effectiveness of the proposed generative CF-INR optimization model \eqref{indirect} by comparing it with the conventional model \eqref{direct} on both image regression and denoising tasks (Table \ref{table: generative optimization models}). The results indicate that the generative optimization model \eqref{indirect} performs slightly better than the conventional model \eqref{direct} for image regression, and exhibits a significant performance advantage over the conventional model for image denoising. This is because the conventional model \eqref{direct} uses the HWT to transform the noisy data into the wavelet space for INR, which results in information distortion. As compared, the generative model \eqref{indirect} directly generates wavelet coefficients and solely uses the IHWT, which avoids the information distortion when handling noisy data. This indicates the compatibility and effectiveness of the generative CF-INR model \eqref{indirect} for image recovery tasks.

\subsubsection{Effectiveness of Cross-Frequency Tensor Decomposition} 
We conduct ablation studies to show the effectiveness of the proposed cross-frequency tensor decomposition \eqref{LRTFR_share} with spectral coupling and spatial decoupling. As shown in Table \ref{table: Architecture Variant}, removing tensor decomposition (i.e., using pure MLPs) leads to a significant drop in both performance and efficiency, highlighting the effectiveness of tensor decomposition paradigms in CF-INR. Additionally, the spatially decoupled and spectrally coupled sharing strategy \eqref{LRTFR_share} further enhances performance and computational efficiency against the classical Tucker decomposition \eqref{LRTFR_all}, by leveraging intra- and inter-correlations among different frequency components under a parameter efficient manner.
\subsubsection{Influences of Other Hyperparameters}
\begin{table}[t]
	\caption{
		The quantitative results by different methods for multi-temporal cloud removal. The best and the second best results are highlighted in \textbf{bold} and {\ul underline}, respectively.}
	\label{table: cloud removal}
	\scriptsize
	\centering
	\setlength{\tabcolsep}{0.5pt}
	\renewcommand{\arraystretch}{1.1}
	\begin{tabular}{cccccccccc}
		\hline
		Data                      & Metric & Observed & TCTV  & HLRTF       & FTNN  & t-$\epsilon$-LogDet & DIP            & LRTFR       & CF-INR\\ \hline
		\multirow{3}{*}{\textit{Jizzakh1}} & PSNR   & 14.54    & 35.98 & 37.78       & 31.37 & 33.15        & 34.63          & {\ul 37.83} & \textbf{39.03} \\
		& SSIM   & 0.676    & 0.953 & {\ul 0.969} & 0.923 & 0.935        & 0.965          & 0.962       & \textbf{0.970} \\
		& NRMSE  & 0.440    & 0.047 & {\ul 0.035} & 0.088 & 0.084        & {\ul 0.035}    & 0.037       & \textbf{0.031} \\ \hline
		\multirow{3}{*}{\textit{Jizzakh2}} & PSNR   & 15.12    & 31.71 & {\ul 32.24} & 29.19 & 28.18        & 31.11          & 32.13       & \textbf{33.84} \\
		& SSIM   & 0.674    & 0.894 & 0.904       & 0.870 & 0.859        & {\ul 0.922}    & 0.893       & \textbf{0.924} \\
		& NRMSE  & 0.441    & 0.082 & {\ul 0.072} & 0.115 & 0.133        & \textbf{0.060} & {\ul 0.072} & \textbf{0.060} \\ \hline
		\multirow{3}{*}{\textit{Puyang}}   & PSNR   & 18.04    & 29.46 & 30.10       & 27.97 & 27.15        & 28.61          & {\ul 30.47} & \textbf{32.06} \\
		& SSIM   & 0.701    & 0.865 & 0.899       & 0.850 & 0.839        & {\ul 0.900}    & 0.893       & \textbf{0.910} \\
		& NRMSE  & 0.422    & 0.133 & 0.107       & 0.148 & 0.164        & {\ul 0.101}    & 0.110       & \textbf{0.089} \\ \hline
	\end{tabular}\vspace{-0.4cm}
	\label{tab:Cloud_Removal}
\end{table}
\begin{table}[t]
	\scriptsize
	\caption{
		Comparison of two CF-INR models \eqref{direct} and \eqref{indirect} on image regression and denoising (Gaussian noise with standard deviation $\sigma=0.2$) for HSIs \textit{WDC mall} and \textit{PaviaU}.}
	\label{table: generative optimization models}
	\centering
	\setlength{\tabcolsep}{4.5pt}
	\begin{tabular}{cccccc}
		\hline
		\multirow{2}{*}{Task}        & Data  & \multicolumn{2}{c}{\textit{WDC mall}} & \multicolumn{2}{c}{\textit{PaviaU}} \\ 
		& Model & Model \eqref{direct}     & Model \eqref{indirect}         & Model \eqref{direct}       & Model \eqref{indirect}          \\ \hline
		\multirow{3}{*}{Regression}   & PSNR  & 52.62          & \textbf{52.72}     & 46.00            & \textbf{46.05}      \\
		& SSIM  & \textbf{0.998}          & \textbf{0.998}              & \textbf{0.979}            & \textbf{0.979}               \\
		& NRMSE & \textbf{0.017}          & \textbf{0.017}              & \textbf{0.047}            & \textbf{0.047}               \\ \hline
		\multirow{3}{*}{Denoising} & PSNR  & 30.47          & \textbf{33.17}     & 32.23            & \textbf{34.18}      \\  
		& SSIM  & 0.839          & \textbf{0.898}     & 0.783            & \textbf{0.840}      \\  
		& NRMSE & 0.211          & \textbf{0.155}     & 0.208            & \textbf{0.168}      \\ \hline
	\end{tabular}\vspace{-0.1cm}
\end{table}
Our method can automatically select suitable cross-frequency ranks $R_1$-$R_4$ and cross-frequency parameters $\omega_1$-$\omega_4$ for precise multi-frequency characterization, based on a set of predefined hyperparameters (e.g., the summing parameters $\mu,\lambda_x,\lambda_y$, the spectral frequency $\omega^z$ and rank $r_z$, and the Adam optimizer parameter such as weight decay). We vary each hyperparameter individually while keeping others fixed, as shown in Fig. \ref{fig: hyperparameter} and Fig. \ref{fig: regularization parameters}. The results indicate that our method is relatively robust w.r.t. these hyperparameters, especially for the summing parameters $\mu,\lambda_x,\lambda_y$, reflecting the robustness of CF-INR for cross-frequency modeling over different data samples. Furthermore, as shown in Fig. \ref{fig: regularization parameters}, the regularization techniques improve CF-INR performance when appropriate hyperparameter values are selected. This also demonstrates CF-INR's compatibility with other proven regularization techniques. In future work, it would be interesting to further integrate such a cross-frequency modeling framework with other intrinsic continuous modeling regularization techniques, such as regularization-by-denoising \cite{RED_TCI} or neural total variation \cite{NeurTV}. 

\begin{table}[t]
	\scriptsize
	\caption{Comparison of different network structures of CF-INR for image inpainting.}
	\label{table: Architecture Variant}
	\centering
	{
		\setlength{\tabcolsep}{4.5pt}
		\begin{tabular}{cccccc}
			\hline
			&CF-INR structure& PSNR  & SSIM  & NRMSE & Time   \\ \hline
			\multirow{3}{*}{\makecell{{\it Flowers}\\SR=0.2}}&Pure MLPs & 37.92 & 0.948 & 0.086 & 163.77 \\
			~&Tucker w/o sharing \eqref{LRTFR_all}& 46.36 & 0.989 & 0.035 & 120.76 \\
			~&Spectrally coupled Tucker \eqref{LRTFR_share}& \bf47.03 & \bf0.990 & \bf0.031 & 99.86  \\ \hline
			\multirow{3}{*}{\makecell{{\it Cups}\\SR=0.2}}&Pure MLPs & 40.23 & 0.978 & 0.029 & 167.24 \\
			~&Tucker w/o sharing \eqref{LRTFR_all}& 46.73 & 0.993 & 0.014 & 130.36 \\
			~&Spectrally coupled Tucker \eqref{LRTFR_share}& \bf48.50 & \bf0.994 & \bf0.012 & 114.30  \\ \hline
	\end{tabular}}\vspace{-0.4cm}
\end{table}
\section{Conclusions\label{sec: Conclusion}}
In this paper, we introduced the self-evolving CF-INR, which explicitly decomposes data into four distinct frequency components through the HWT, enabling tailored modeling for each frequency component using INR. Furthermore, we suggested the cross-frequency tensor decomposition paradigm for CF-INR, which largely improves its efficiency and modeling capability for cross-frequency correlation excavation. Theoretically, we developed novel analytical frameworks for CF-INR, including cross-frequency tensor rank and Laplacian smoothness analyses, which rigorously underpin the self-evolving strategies to optimize cross-frequency parameters and cross-frequency tensor ranks automatically and enhance model adaptability across different datasets. We conducted extensive experimental evaluations on multi-dimensional data representation and recovery tasks (image regression, inpainting, denoising, and cloud removal), which validated the broad applicability and superior performance of our method compared with state-of-the-art methods.
\begin{figure*}[t]
	\setlength{\tabcolsep}{0pt}
	\centering
	\begin{tabular}{cccccc}
		\vspace{-0.1cm}
		\includegraphics[width=0.167\textwidth]{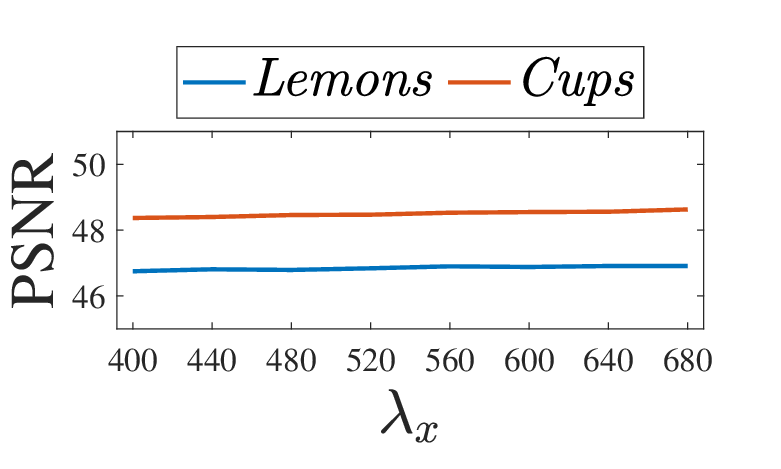}&
		\includegraphics[width=0.167\textwidth]{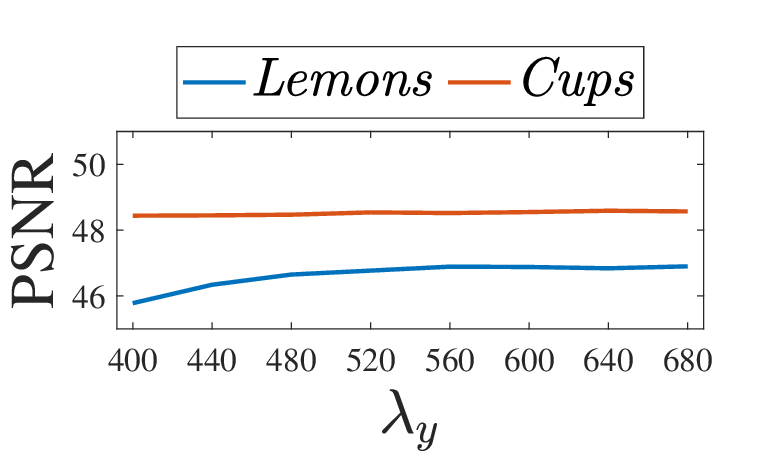}&
		\includegraphics[width=0.167\textwidth]{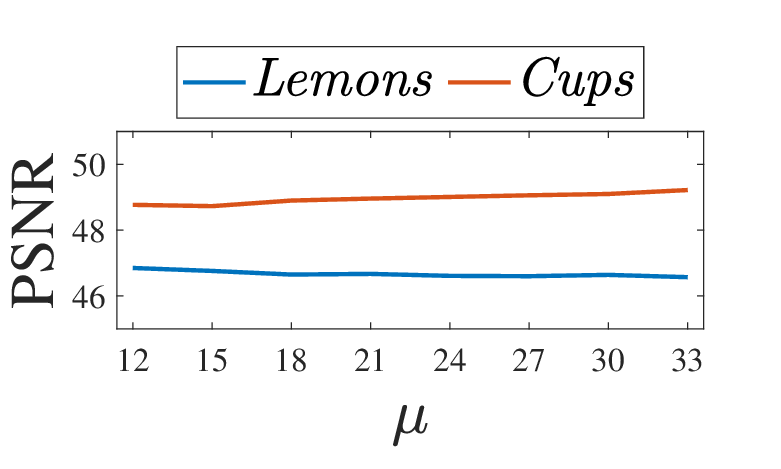}&
		\includegraphics[width=0.167\textwidth]{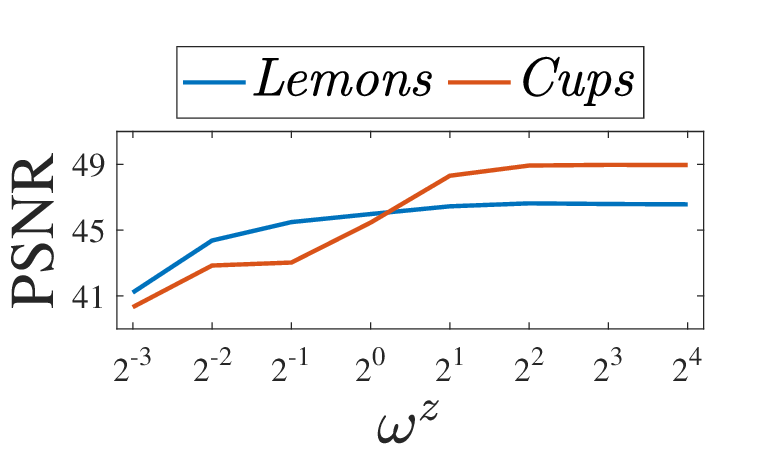}&
		\includegraphics[width=0.167\textwidth]{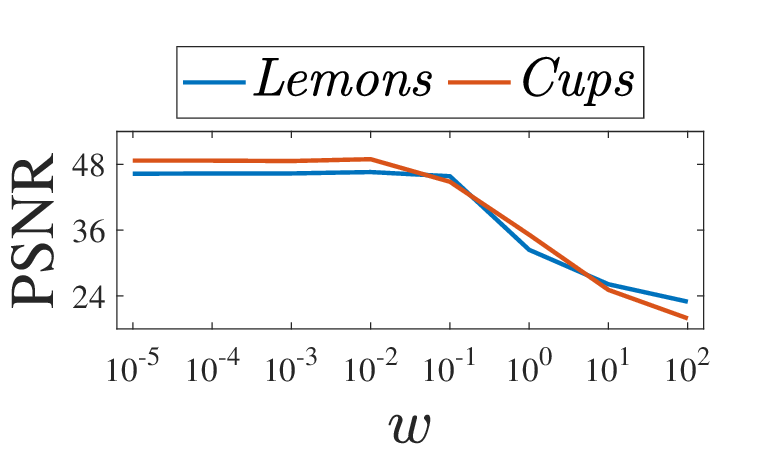}&
		\includegraphics[width=0.167\textwidth]{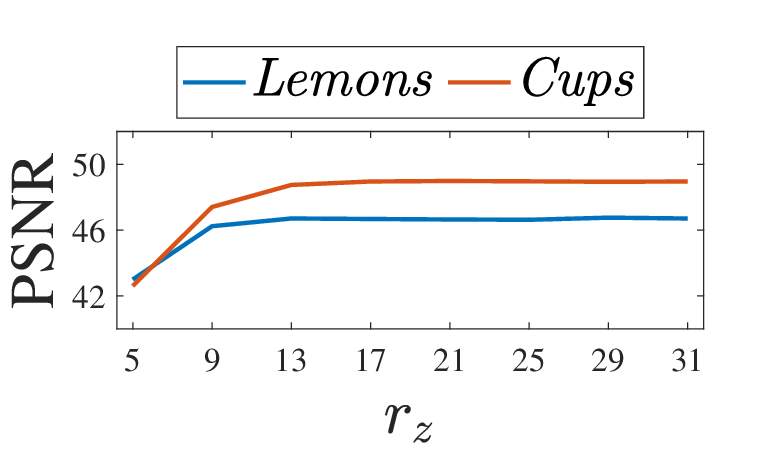}\\
		\vspace{-0.1cm}
		\includegraphics[width=0.167\textwidth]{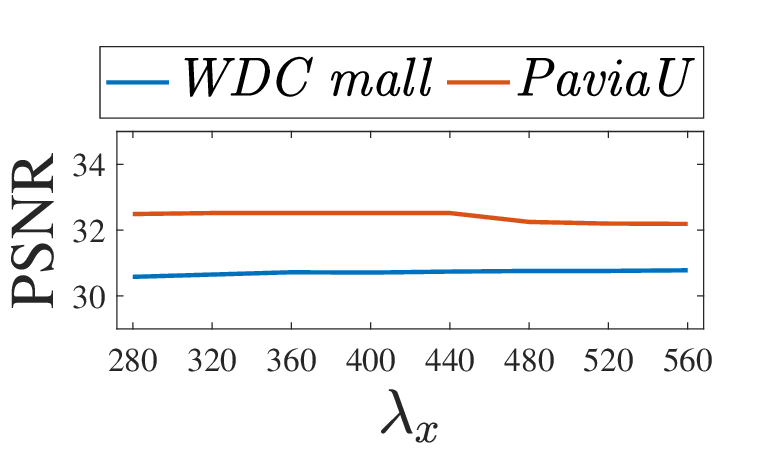}&
		\includegraphics[width=0.167\textwidth]{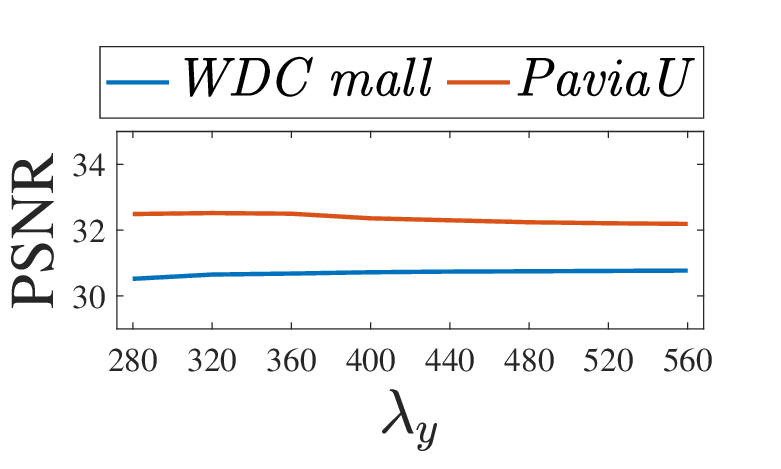}&
		\includegraphics[width=0.167\textwidth]{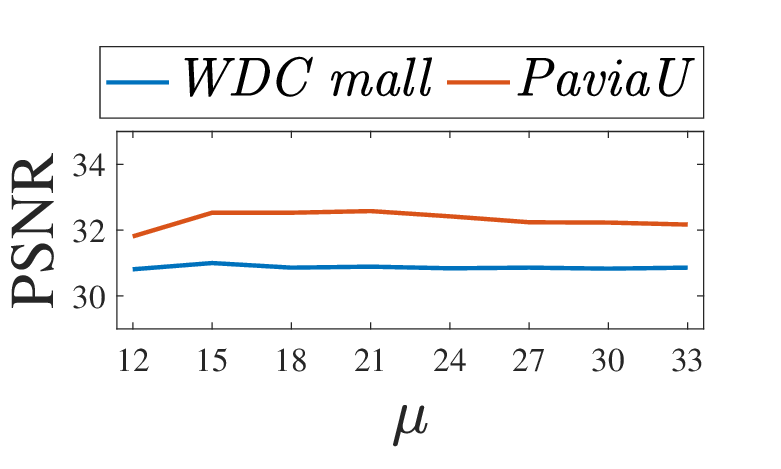}&
		\includegraphics[width=0.167\textwidth]{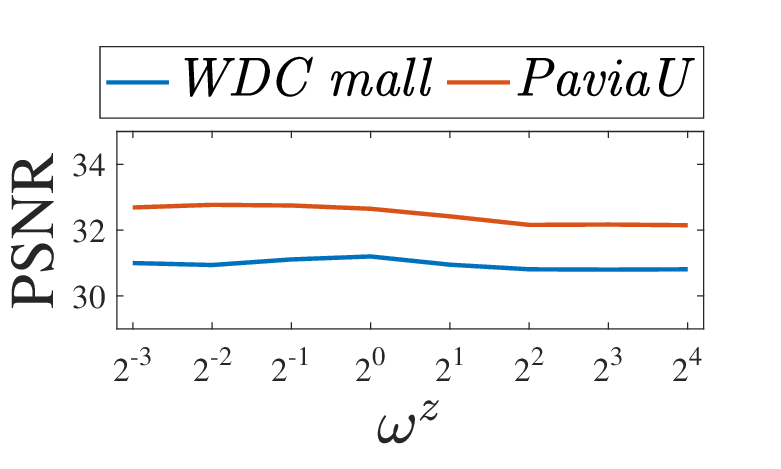}&
		\includegraphics[width=0.167\textwidth]{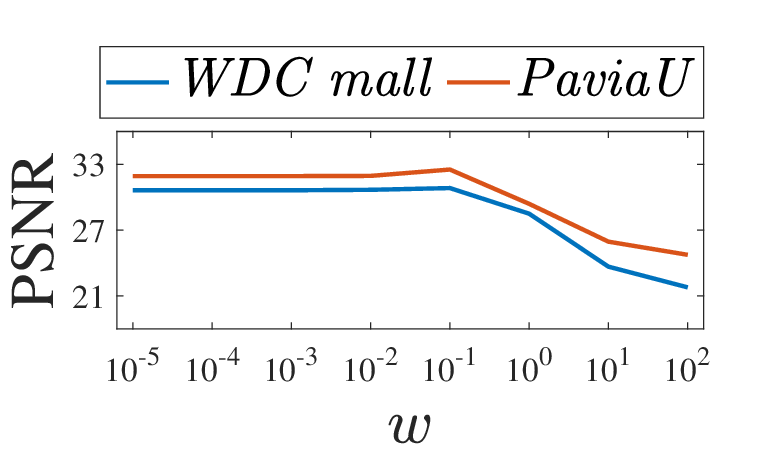}&
		\includegraphics[width=0.167\textwidth]{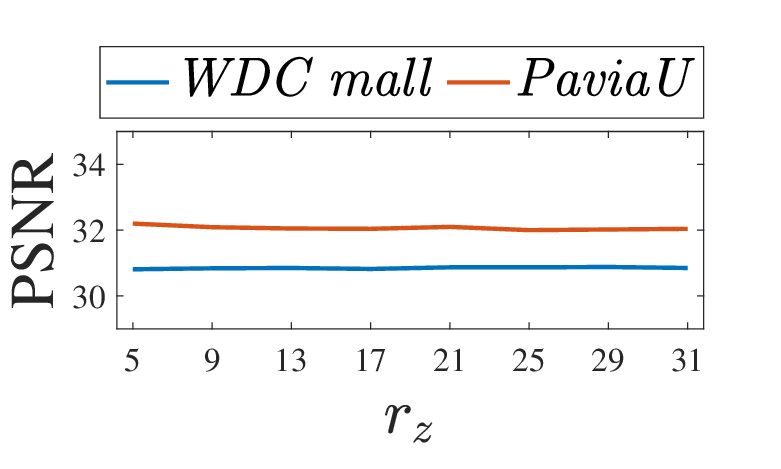}\\\vspace{-0.1cm}
		(a)&(b)&(c)&(d)&(e)&(f)\\
	\end{tabular}
	\caption{Sensitivity tests for hyperparameters except for the self-evolving cross-frequency ranks $R_1$-$R_4$ and cross-frequency parameters $\omega1$-$\omega_4$. (a)-(b): The rank summing parameters $\lambda_x$, $\lambda_y$. (c) The frequency summing parameters $\mu$. (d) The spectral frequency parameter $\omega^z$. (e) The weight decay $w$. (f) The spectral rank $r_z$. Tests were conducted for the inpainting (top) and denoising (bottom).
	}
	\vspace{-0.5cm}
	\label{fig: hyperparameter}
\end{figure*}
\begin{figure}[t]
	\centering
	\setlength{\tabcolsep}{0pt}
	\begin{tabular}{cc}
		\includegraphics[width=0.235\textwidth]{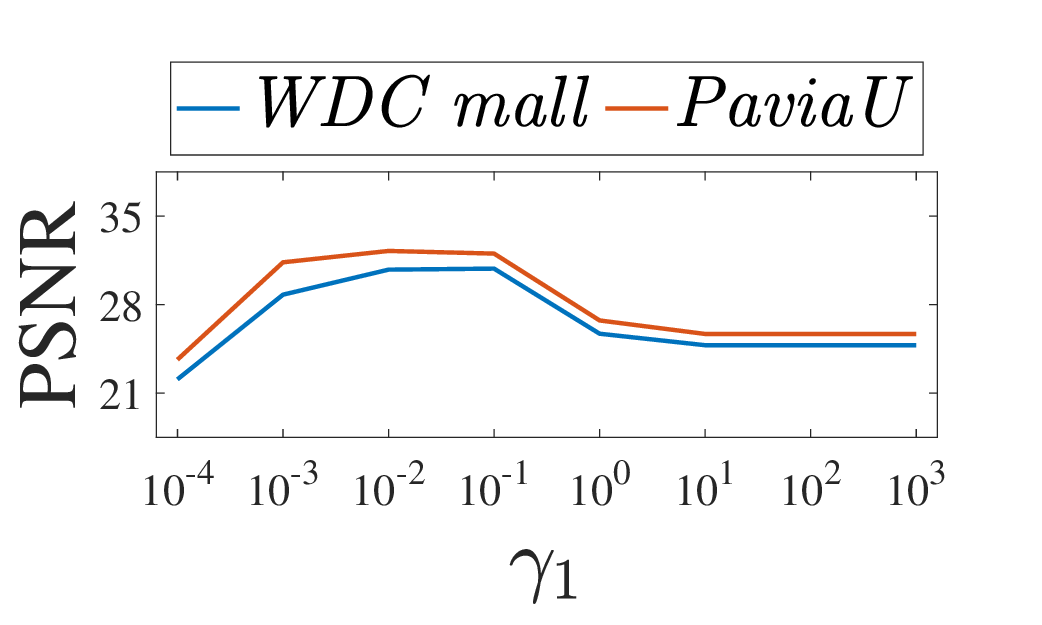}&
		\includegraphics[width=0.235\textwidth]{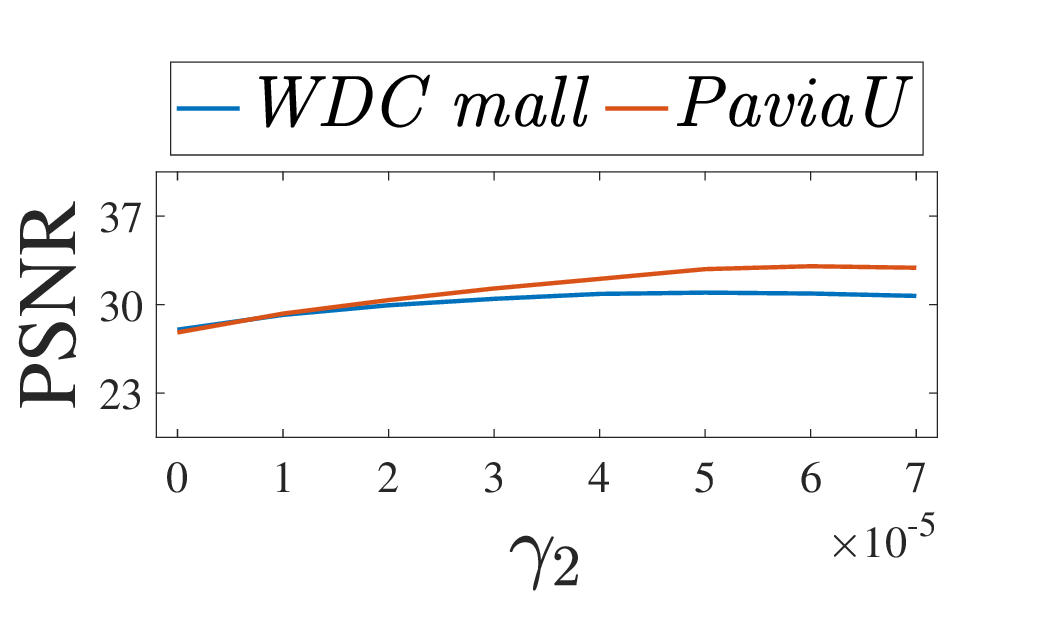}\\
	\end{tabular}\vspace{-0.2cm}
	\caption{Sensitivity tests for the regularization parameters in the denoising model \eqref{model_denoise}.
	}\vspace{-0.5cm}
	\label{fig: regularization parameters}
\end{figure}

\bibliographystyle{IEEEtran}
\bibliography{IEEEabrv}
\end{document}